\documentclass[twoside,11pt]{article}
\usepackage{jmlr2e}
\usepackage{calc}
\usepackage{longtable}
\usepackage{booktabs}
\usepackage{array}
\usepackage[T1]{fontenc}
\usepackage[utf8]{inputenc} % pdfLaTeX input encoding
\usepackage{lmodern}
\usepackage{microtype}

\usepackage{amsmath,amssymb}



\ShortHeadings{From Garbage to Gold}{Lee-St. John, Lawson, and Piechowski-Jozwiak}
\firstpageno{1}

\begin{document}

\title{From Garbage to Gold:

A Data-Architectural Theory of Predictive Robustness}

% Example author block(s) -- repeat or group as needed
\author{\name Terrence J. Lee-St. John \email tjleestjohn@gmail.com\\
       \addr Independent Researcher \& Institute for Healthier Living Abu Dhabi\\
       Abu Dhabi, UAE
\AND
       \name Jordan L. Lawson \email lawsonja31@gmail.com\\
       \addr Independent Researcher \& Brown University, Center for Computation and Visualization\\
       Providence, RI, USA
\AND
       \name Bartlomiej Piechowski-Jozwiak \email neurobart@gmail.com\\
       \addr Independent Researcher \& Institute for Healthier Living Abu Dhabi\\
       Abu Dhabi, UAE}

\editor{Submitted for Review} % leave empty if unknown at submission

\maketitle

% `content.tex` should start with \begin{abstract}...\end{abstract}
\makeatletter
% --- Compatibility shims (keep even if packages are loaded) ---
\providecommand{\real}[1]{#1} % used in p{(\linewidth - 2\tabcolsep) * \real{...}}
\providecommand{\toprule}{\hline}
\providecommand{\midrule}{\hline}
\providecommand{\bottomrule}{\hline}
\providecommand{\arraybackslash}{}
\makeatother

\begin{abstract}
In the era of big data, tabular machine learning (ML) presents a paradox: modern models have demonstrated state-of-the-art performance using high-dimensional (high-D), collinear, error-prone data, defying the ``Garbage In, Garbage Out'' mantra. To help resolve this, we synthesize principles from Information Theory, Latent Factor Models, and Psychometrics, clarifying that predictive robustness arises not solely from data cleanliness, but from the synergy between data architecture and model capacity. By partitioning predictor-space ``noise'' into ``Predictor Error'' and ``Structural Uncertainty'' (informational deficits due to stochastic generative mappings), we prove that leveraging a high-D set of error-prone predictors can asymptotically overcome both types of noise, whereas cleaning a low-D set is fundamentally bounded by Structural Uncertainty. We demonstrate why ``Informative Collinearity'' (dependencies arising from shared latent drivers) is an asset that enhances reliability and convergence efficiency, not a nuisance. Further, we explain why increased dimensionality reduces the burden of latent inference, enabling feasibility with finite samples. However, to balance the ideal with practical constraints, we present ``Proactive Data-Centric AI'' as a strategy to identify sets of predictors that enable predictive robustness efficiently. Additionally, we derive boundaries imposed by Systematic Error Regimes and show why models capable of absorbing ``rogue'' dependencies can mitigate violations of key assumptions. We also show (in the binary context) that optimally exploiting these mechanisms requires interactions. Finally, we link latent architecture to Benign Overfitting, offering a first step toward a unified understanding of robustness to both ``Outcome Error'' and predictor-space noise. Critically, however, we also provide a principled operational boundary delineating when traditional DCAI's focus on outcome variable cleaning remains distinctly powerful. By redefining data quality from item-level perfection to portfolio-level architecture, this framework provides a theoretical rationale for the development of ``Local Factories'' -- systems that learn from uncurated, live data streams characteristic of enterprise ``data swamps'' -- supporting a deployment paradigm shift from ``Model Transfer'' to ``Methodology Transfer'' to overcome the inherent limitations of static model generalizability.
\end{abstract}


\section*{Table of Contents}

\noindent \textbf{\ref{Intro}. Introduction} \dotfill \pageref{Intro} \\
\noindent \textbf{\ref{Background}. Background} \dotfill \pageref{Background} \\
\noindent \textbf{\ref{Framework Model and Notation}. Framework Model and Notation} \dotfill \pageref{Framework Model and Notation} \\
\noindent \hspace*{1.5em} \textit{\ref{Causal Consistency and Parametric Constraints}. ``Causal Consistency'' and Parametric Constraints} \dotfill \pageref{Causal Consistency and Parametric Constraints} \\
\noindent \hspace*{1.5em} \textit{\ref{Glossary}. Glossary of Terms and Concepts} \dotfill \pageref{Glossary} \\
\noindent \textbf{\ref{Core Theoretical Analysis}. Core Theoretical Analysis} \dotfill \pageref{Core Theoretical Analysis} \\
\noindent \hspace*{1.5em} \textit{\ref{Triangulating the Truth: The Detective Analogy}. Triangulating the Truth: The Detective Analogy} \dotfill \pageref{Triangulating the Truth: The Detective Analogy} \\
\noindent \textbf{\ref{A Model of Systematic Predictor Errors and Its Impact}. A Model of Systematic Predictor Errors and Its Impact} \dotfill \pageref{A Model of Systematic Predictor Errors and Its Impact} \\
\noindent \textbf{\ref{Interplay of N and m}. A Statistical Treatment of Inferential Power and Breadth} \dotfill \pageref{Interplay of N and m} \\
\noindent \hspace*{1.5em} \textit{\ref{Resolving Conflict: Prev Floor}. Resolving the Conflict: Breadth and the Prevalence Floor} \dotfill \pageref{Resolving Conflict: Prev Floor} \\
\noindent \hspace*{1.5em} \textit{\ref{Benefits of Causal Consistency}. Benefits of Causal Consistency: Efficiency and Interpolation} \dotfill \pageref{Benefits of Causal Consistency} \\
\noindent \textbf{\ref{Benign Overfitting Link: Towards a Unified Understanding of Robustness}. Benign Overfitting Link: Towards a Unified Understanding of Robustness} \dotfill \pageref{Benign Overfitting Link: Towards a Unified Understanding of Robustness} \\
\noindent \hspace*{1.5em} \textit{\ref{The Generative Origin of Low Rank: Effective Latent Complexity and Causal Consistency}. The Generative Origin of Low Rank} \dotfill \pageref{The Generative Origin of Low Rank: Effective Latent Complexity and Causal Consistency} \\
\noindent \textbf{\ref{The Objective-Capability Gap and Representation Learning}. The Objective-Capability Gap and Representation Learning} \dotfill \pageref{The Objective-Capability Gap and Representation Learning} \\
\noindent \hspace*{1.5em} \textit{\ref{The Engine and Fuel Analogy, Part 1}. The Engine and Fuel Analogy, Part 1} \dotfill \pageref{The Engine and Fuel Analogy, Part 1} \\
\noindent \textbf{\ref{Violating LI}. Violations of Local Independence (LI)} \dotfill \pageref{Violating LI} \\
\noindent \hspace*{1.5em} \textit{\ref{Absorption Hypothesis}. The ``Absorption Hypothesis''} \dotfill \pageref{Absorption Hypothesis} \\
\noindent \hspace*{1.5em} \textit{\ref{The Risk of Malignant Overfitting}. The Risk of Malignant Overfitting} \dotfill \pageref{The Risk of Malignant Overfitting} \\
\noindent \textbf{\ref{The Geometry of Outcome Error}. The Geometry of Outcome Error} \dotfill \pageref{The Geometry of Outcome Error} \\
\noindent \hspace*{1.5em} \textit{\ref{A DCAI-Aligned Resource Shortcut: ``Dirty Inputs, Clean Targets''}. A DCAI-Aligned Resource Shortcut: ``Dirty Inputs, Clean Targets''} \dotfill \pageref{A DCAI-Aligned Resource Shortcut: ``Dirty Inputs, Clean Targets''} \\
\noindent \textbf{\ref{Towards Proactive Data-Centric AI}. Proactive Data-Centric AI: Data Collection and Feature Selection} \dotfill \pageref{Towards Proactive Data-Centric AI} \\
\noindent \hspace*{1.5em} \textit{\ref{FeatureSelection}. Implications for Feature Selection} \dotfill \pageref{FeatureSelection} \\
\noindent \hspace*{1.5em} \textit{\ref{The Engine and Fuel Analogy, Part 2: P-DCAI as Strategic Refinement}. The Engine and Fuel Analogy, Part 2: P-DCAI as Strategic Refinement} \dotfill \pageref{The Engine and Fuel Analogy, Part 2: P-DCAI as Strategic Refinement} \\
\noindent \hspace*{1.5em} \textit{\ref{Methodology Transfer: A Democratized Deployment Paradigm}. Methodology Transfer: A Democratized Deployment Paradigm} \dotfill \pageref{Methodology Transfer: A Democratized Deployment Paradigm} \\
\noindent \textbf{\ref{Modeling Methodology}. Modeling Methodology} \dotfill \pageref{Modeling Methodology} \\
\noindent \textbf{\ref{Sim}. Simulation Study} \dotfill \pageref{Sim} \\
\noindent \textbf{\ref{CCAD Motivating Example}. Motivating Case Study: Cleveland Clinic Abu Dhabi} \dotfill \pageref{CCAD Motivating Example} \\
\noindent \textbf{\ref{Limitations and a Future Research Agenda}. Limitations and Future Research Agenda} \dotfill \pageref{Limitations and a Future Research Agenda} \\
\noindent \textbf{\ref{Conclusion}. Conclusion} \dotfill \pageref{Conclusion} \\
\noindent \textbf{Disclosures} \dotfill \pageref{Disclosures} \\

\noindent \textbf{Appendices} \\
\noindent \ref{phi corr}. Formulation of \(\phi(X_{i}^{(1)},X_{j}^{(2)})\) \dotfill \pageref{phi corr} \\
\noindent \ref{continuous vars}. Generalization to Continuous Variables \dotfill \pageref{continuous vars} \\
\noindent \ref{asymp Breadth v Depth}. Asymptotic Analysis: Comparing Infinite Breadth and Infinite Depth \dotfill \pageref{asymp Breadth v Depth} \\
\noindent \ref{latent sparsity}. Analysis of Latent Sparsity and Effective Rank \dotfill \pageref{latent sparsity} \\
\noindent \ref{Robustness of the Framework Across Alt Causal Structures}. Alignment with Alternative Causal Structures \dotfill \pageref{Robustness of the Framework Across Alt Causal Structures} \\
\noindent \ref{finite convergence}. Derivation of the Finite-\(m\) Convergence Rate (Efficiency) \dotfill \pageref{finite convergence} \\
\noindent \ref{equivalence for LI viol}. Proof of Equivalence for LI Violations \dotfill \pageref{equivalence for LI viol} \\
\noindent \ref{Additional Simulation Graphs: Breadth vs. Depth as AUROC and AUPRC}. Additional Simulation Graphs: Breadth vs. Depth as AUROC and AUPRC \dotfill \pageref{Additional Simulation Graphs: Breadth vs. Depth as AUROC and AUPRC} \\

\section{Introduction}
\label{Intro}

In the era of big data, an emerging paradox has surfaced in tabular machine learning (ML): highly flexible models have achieved state-of-the-art results despite using high-dimensional (high-D), collinear, error-prone data that have undergone minimal-to-no manual curation. Such results defy conventional wisdom because the long-standing, dominant pre-processing paradigm, summed up by the maxim of ``Garbage In, Garbage Out'' (GIGO), views high dimensionality, collinearity, and data errors as compounding liabilities that preclude the construction of robust models. Focusing on the predictor-variable side (the predictor-space), the conventional response has relied on a toolkit of aggressive data cleaning and dimensionality reduction \citep{Han_2011, guyon2003}. This paradigm is rooted in an era dominated by classical modeling methods -- where datasets were relatively small, the interpretability of inferential statistical coefficients was paramount, and, given the common techniques, parsimony was the primary defense against overfitting. In this foundational period, GIGO was a necessary precondition for modeling, one that was optimized for an inferential focus and the relevant information-theoretic constraints. However, in the context of big data and modern ML, where the primary goal is often accurate prediction rather than parameter inference, the traditional paradigm faces two critical limitations: (1) it can be impractical to execute at scale and (2) it neglects a more fundamental, structural challenge inherent in the data-generating process itself that is operationally distinct from data errors.

We began development of the theory presented in this manuscript to retrospectively explain the seemingly paradoxical success of one such engineering case study carried out at Cleveland Clinic Abu Dhabi (CCAD) -- a 66-month pseudo-prospective study predicting initial stroke or myocardial infarction \citep{leestjohn2024}. This study purposefully operated under the contrarian engineering hypothesis ``that high-D and correlated predictor-spaces can yield a predictive mechanism that remains robust to errors in individual predictive pathways... (because) multiple complementary pathways... (exist) for the prediction signal to traverse.'' Traditional predictive models typically rely heavily on content knowledge and curated data, creating static tools that align with domain expertise. In contrast, the CCAD case study operated using empirically based, automated feature selection, operating without manual data cleaning, drawing directly from the real-world data stream produced by the hospital's day-to-day operations. Ultimately, this deployment scaled to encompass \(588,105\) patients, more than \(3.4\) million patient-months, and utilized several thousand error-prone predictors, achieving significantly higher performance than that expected from the traditional risk model that served as the standard of care. Given that real-world medical data represent an extreme test case -- electronic health records are high-D, sparse, collinear, and notoriously error-ridden -- these results suggest that the contrarian engineering hypothesis was at least not wildly unreasonable. Despite this success, however, the CCAD case study stopped short of formalizing the exact ``why'' and ``how,'' leaving its success positioned as an empirical curiosity.

Critically, the phenomenon observed at CCAD is not unique and should not be viewed as an isolated event -- it resonates with similar achievements in the same field \citep{rajkomar2018}, as well as those in other fields as diverse as financial forecasting \citep{gu2020} and antibiotic discovery \citep{stokes2020}. Thus, this phenomenon is not limited to a single domain; it has been observed in enterprise warehouse ``data swamps'' (vast repositories of raw, error-prone data) across industries. Viewing these results collectively as manifestations of a common, cross-domain principle creates a need for a new theoretical perspective -- one that provides the formal language to explain why such successes are able to defy the GIGO paradigm, thereby enabling a move from empirical curiosity to predictable engineering.

While valuable theories have already emerged that provide partial explanation (e.g., ``Blessing of Dimensionality'' and ``Benign Overfitting''; \citealp{bartlett2020, donoho2000, hastie2022, shamir2022, tsigler2023}), they typically treat the data distribution as a fixed, static input and view predictor-space noise as a monolithic nuisance. By collapsing distinct forms of deviation into a single undifferentiated error term, such frameworks implement a ``Flat Topological Operationalization,'' modeling under the implicit assumption that predictors map directly to the outcome. However, in the presence of a hierarchical data-generating process, such flattening obscures the reality that predictors are not definitive objects, but instead are imperfect proxies of deeper latent structures. Consequently, because such methods do not mathematically distinguish between measurement error and structural ambiguity, they remain structurally agnostic to the architectural benefits of triangulation, focusing instead on predictor-space solutions that are model- (e.g., projection, explicit regularization; \citealp{hastie2009elements}) or algorithm-centric (e.g., implicit regularization; \citealp{neyshabur2017pac, zhang2021understanding}).

While ML theories of Representation Learning \citep{bengio2013representation} argue that modern high-capacity models (e.g., Deep Learning) often succeed by escaping this flat topology -- learning latent structures as a byproduct of monolithic error minimization -- this remains an emergent phenomenon rather than an engineered feature. We reserve a formal discussion of this objective-capability gap for Section~\ref{The Objective-Capability Gap and Representation Learning}, after we have defined the architectural principles that drive it.

To provide the structural foundation required for engineered robustness, this manuscript introduces a data-architectural theory that derives predictive performance from a first-principles analysis of a plausible, latent-hierarchical data-generating structure. Unlike prior works that treat noise as a monolith, we formally partition predictor-space noise into two components: ``Predictor Error'' and ``Structural Uncertainty.'' Predictor Error represents measurement discrepancy. Conversely, Structural Uncertainty refers to the information deficit inherent in the predictors themselves -- the degree to which even a set of perfectly measured predictors fails to fully capture the underlying latent drivers of the system. This partitioning allows us to show that these two types of noise obey different information-theoretic limits. By establishing this differentiation, we redefine ``high-quality'' data in the context of big data, shifting the focus from the fidelity of individual variables toward the integrity of the portfolio-level architecture.

This ``Garbage to Gold'' framework (G2G) potentially offers distinct theoretical and practical utility. At a high conceptual level, it establishes that robustness is not merely a property imposed by a model onto data, but rather is a potential inherent in a specific data architecture that a sufficient model can unlock. Furthermore, by formalizing the structural logic and implications of the data-generating process, the downstream behavior of the model becomes a predictable consequence of the underlying ``chemistry'' of the information. This shift allows robustness to be actively engineered rather than merely sought through post-hoc algorithmic regularities. Operationalizing this at a lower level, G2G provides a more precise toolkit for the ML practitioner. For example, when troubleshooting, this data-architectural perspective explains a model's failure on a new dataset by hypothesizing that the data lack a known, requisite internal structure -- a more specific diagnosis than the general statistical discrepancies cited by model- or algorithm-centric views. It also provides clear targets for designing novel models and algorithms by identifying specific structural mechanisms in the data the new methods should aim to exploit. Finally, it informs specific data engineering strategies (acquisition and feature selection), enabling practitioners to proactively induce predictive robustness through strategic design choices.

\vspace{0.5\baselineskip}
\begin{figure}
\centering
\includegraphics[width=.90\textwidth,keepaspectratio]
{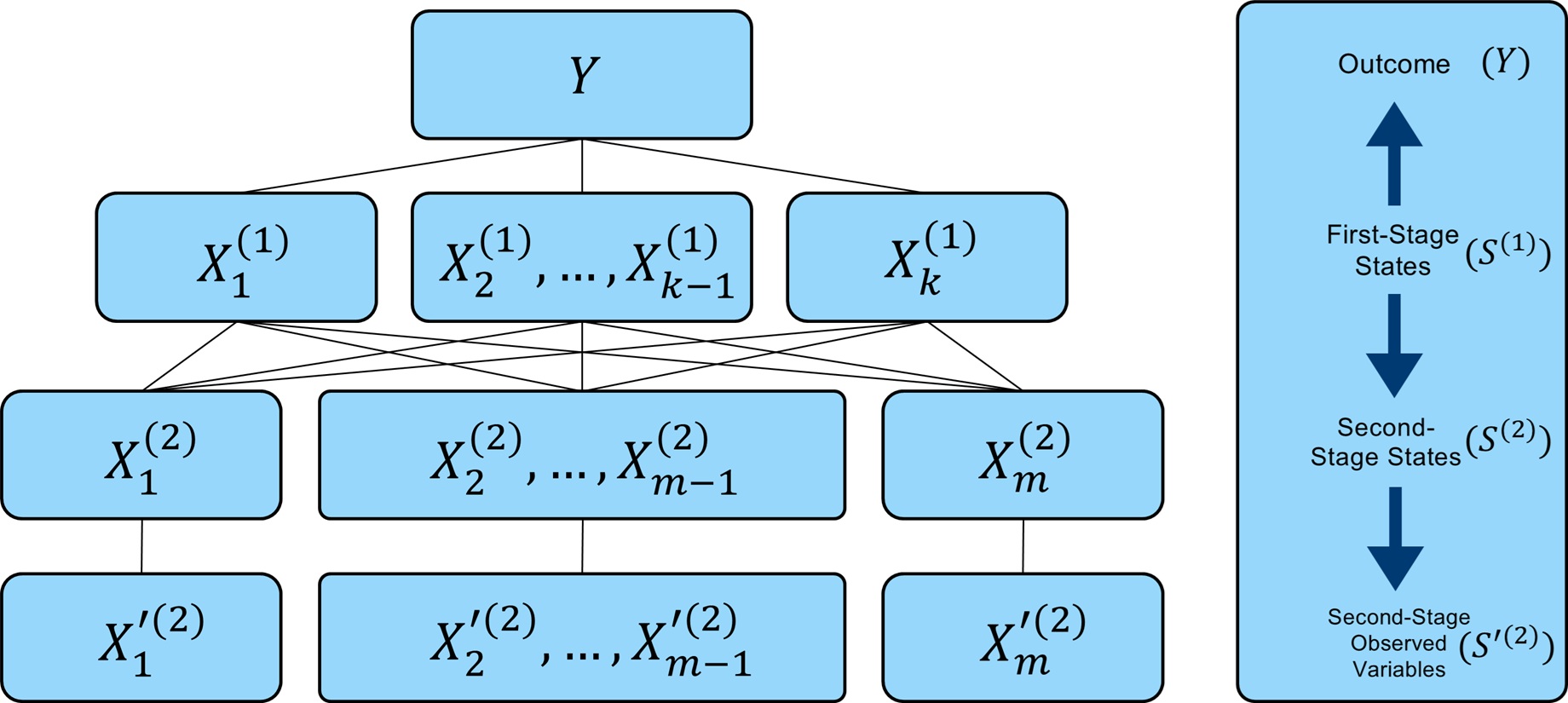}
\caption{The Primary Structure.}
\label{fig: Primary Structure}
\end{figure}

We formalize G2G by analyzing a hierarchical data-generating structure (\(Y \leftarrow S^{(1)} \rightarrow S^{(2)} \rightarrow S^{\prime(2)}\), Fig.~\ref{fig: Primary Structure}) that aligns with common, latent factor models \citep{spearman1904} and the Manifold Hypothesis (the theory that high-D data often lie on low-D manifolds; \citealt{cayton2005, fefferman2016}). We show how the primary structure has the inherent potential to mitigate Predictor Error and Structural Uncertainty, two fundamental challenges when predicting the outcome (\(Y\)) from the observed predictors (\(S^{\prime(2)}\)). In doing so, we reinterpret high-D ``Breadth'' and a specific type of collinearity (``informative'') as architectural assets that enable predictive robustness by allowing suitable models to triangulate a stable representation of the primary latent layer (\(S^{(1)}\)) which, in this structure, is the sole driver of \(Y\). We define ``Informative Collinearity'' as dependencies among predictors that arise from their shared latent causes in \(S^{(1)}\), where predictors remain ``distinct'' provided that the probabilistic realizations of their generative pathways (\(S^{(1)} \rightarrow S^{(2)}\); Structural Uncertainty) and error processes (\(S^{(2)} \rightarrow S^{\prime(2)}\); Predictor Error) are conditionally independent. Breadth also relies on distinct predictors, but is, as a concept, uncoupled from their inter-dependencies. Our analysis yields several key contributions:

\begin{itemize}
\item
\textbf{The Superiority of Breadth in Contexts Characterized by Structural Uncertainty:} We contrast a high-D Breadth strategy with a repeated-measures ``Depth'' strategy (analogous to improving observational fidelity via data cleaning), proving that while Breadth can asymptotically eliminate both Predictor Error and Structural Uncertainty, Depth remains fundamentally bounded by the Structural Uncertainty of the fixed feature set.
\item
\textbf{Efficiency and Reliability:} We derive the finite-sample convergence rate (``efficiency''), identifying the distinct roles of ``novelty'' (for completeness) and Informative Collinearity (for reliability) in accelerating latent state recovery.
\item
\textbf{Statistical Feasibility of Breadth:} We demonstrate how the latent architecture averts the apparent conflict between high-D Breadth and the Curse of Dimensionality. We analyze the interplay of feature set dimensionality (\(m\)) with sample size (\(N\)), showing that the architecture creates a ``Prevalence Floor'' so that data sparsity is bounded even as \(m\) increases.
\item
\textbf{The Necessity of Interactions:} We prove (in the binary context) that optimally exploiting this architecture requires interactions/non-linearities, providing a data-architectural explanation for the success of modern, flexible models in big data contexts.
\item
\textbf{Systematic Error and Assumption Violations:} We derive the theoretical limits imposed by Systematic Error Regimes and, relatedly, provide a data-architectural justification for why flexible, unsupervised models can theoretically mitigate violations of core assumptions (e.g., local independence and independent errors).
\item
\textbf{Connection to Benign Overfitting (BO):} We prove that the latent architecture also provides a generative origin for the ``low-rank-plus-diagonal'' covariance characteristics linked to BO (a phenomenon related to overcoming outcome variable error; \citealt{bartlett2020, hastie2022, tsigler2023}), showing that this favorable structure is a natural consequence of the underlying data-generating mechanism -- specifically, the sparsity of the latent signal relative to the high-D noise. Although our analysis overlaps with important recent work linking factor models to BO \citep{bing2021, bunea2022}, our contribution is the synthesis. The novel connection we highlight, born from a single architecture, provides a first step towards a unified understanding of robustness encompassing both Outcome Error and predictor-space noise.
\item
\textbf{``Proactive Data-Centric AI'' (P-DCAI):} Overall, G2G aligns with Data-Centric AI’s (DCAI) call to prioritize data quality over model iteration \citep{ng2021, polyzotis2021datacentric}. By redefining how high-quality data are conceptualized, however, we extend this data-centric focus by introducing P-DCAI, an operationally distinct data-centric strategy. While DCAI is primarily operationalized via label cleaning (aligning with GIGO), P-DCAI is primarily operationalized via data engineering choices. Specifically, P-DCAI employs data acquisition and feature selection strategies that identify efficient sets of predictors rather than indiscriminately utilizing everything available, thereby shifting the burden of robustness from post-hoc cleaning to a priori architectural design of the dataset.

However, we do not argue for a strict departure from DCAI. Rather, we provide an architectural argument describing where and why DCAI's focus on label curation remains distinctly powerful (Section~\ref{The Risk of Malignant Overfitting}), while maintaining that G2G mitigates predictor-side noise.

Thus, though this framework provides a rationale for the use of uncurated data in certain contexts, it also defines the operational boundary where the highlighted architectural resilience must meet label curation (DCAI) to guarantee robust learning.
\end{itemize}

We clarify that our framework is born of pragmatism. It in no manner advocates for data errors over high fidelity. In an unconstrained, idealized setting, ``clean Breadth'' -- a high-D set of distinct, error-free predictors -- remains the gold standard. However, real-world applications often force a trade-off between the perfection of individual variables and the comprehensiveness of the variable set -- manual data cleaning acts as a bottleneck on dimensionality so that practitioners are effectively forced to choose between ``Clean Parsimony'' (a high fidelity, low-D set) and ``Dirty Breadth'' (a low fidelity, high-D set). By proving that for any fixed set of predictors there exists a ``Performance Floor'' determined by the Structural Uncertainty inherent to that set, this framework demonstrates that Breadth is the only mechanism that can break through this fundamental barrier. This Performance Floor represents the extent to which the set of predictors fails to fully capture the underlying latent drivers -- a limit below which no amount of reduction in Observational Error can reduce uncertainty in the prediction. Consequently, this manuscript shows that Dirty Breadth can outperform Clean Parsimony not because Observational Error is in any way beneficial, but because the architectural advantage of comprehensive, redundant coverage can outweigh the penalty imposed by data errors.

Finally, this architectural resilience suggests a more democratic paradigm for ML deployment -- a shift to a ``Local Factory'' model where rather than exporting a static, universal model (``Model Transfer'') that degrades across sites, a robust methodology is exported (``Methodology Transfer'') that allows local institutions to refine and model their own raw data streams, effectively turning the varied local contexts (which risk low generalizability under the Model Transfer paradigm) into an asset by exploiting site-specific patterns.

Having outlined our core claims, we now describe the strategy employed in this manuscript. To establish G2G, we adopt an analytical approach common in the formalization of new theories: we begin within an idealized and tractable context (binary variables and independent Predictor Error) to provide the mathematical clarity necessary to isolate fundamental mechanisms and prove core principles. We then expand the analysis to incorporate systematic errors and violations of key assumptions. We also identify which results readily generalize to continuous variables, as well as those whose generalization requires further, non-trivial exposition. Lastly, we provide a simulation study that demonstrates core mechanisms.

The uniqueness of this work lies not in the invention of the math, high-level ideas, structures, or tools we leverage, which clearly originate from multiple, well-established fields (as outlined in Section~\ref{Background}). Instead, our contribution is the synthesis -- the application of classic disciplines to a modern context in order to formalize a novel framework. Specifically, we utilize Information Theory \citep{cover2006, shannon1948} to translate structures, concepts, and logic from Latent Factor Models \citep{spearman1904, Thurstone1947} and Psychometrics/Item Response Theory \citep{lord1968, hambleton1991} to the high-D, error-prone context of big data, providing the formal theoretical language necessary to explain why modern ML increasingly yields gold predictions from data traditionally dismissed as garbage.

Despite covering many theoretical and methodological topics, however, this manuscript should neither be viewed as a complete theoretical treatise nor as a presentation of an off-the-shelf method. Rather, our aim is simply to provide a rigorous, introductory presentation of G2G -- one meant to both (1) elevate the understanding of the role of data architecture in predictive robustness, and (2) spur numerous lines of future theoretical, empirical, and methodological investigation and development (as outlined in Section~\ref{Limitations and a Future Research Agenda}).

\section{Background}
\label{Background}

To contextualize this manuscript, it is essential to understand the tabular ML paradigm it challenges. In addition to the undeniable, negative impact of data errors, the conventional pre-processing perspective is built on two concerns: collinearity and the ``Curse of Dimensionality'' \citep{bellman1961adaptive, belsley1980}. Collinearity is defined as the presence of linear (or near-linear) dependencies among predictors. While its impact on overall predictive accuracy is understood to be complex and not always negative, conventional wisdom emphasizes its detrimental effects on model interpretability and the stability of parameter estimates. The Curse of Dimensionality describes the numerous difficulties that arise when modeling in high-D spaces. Central among these is unbounded data sparsity, which classically implies that the sample size (\(N\)) required to reliably characterize the predictor-space grows exponentially with its dimensionality (\(m\)) \citep{donoho2000}, leading to an increased risk of overfitting. Notably, it has been understood that Predictor Error only serves to compound these issues \citep{belsley1980}.

The pre-processing response is often twofold. First, data cleaning is prioritized. Second, data reduction and feature selection are generally prescribed \citep{guyon2003}. Popular supervised algorithms, such as Minimum Redundancy Maximum Relevance (MRMR) \citep{Peng2005}, are a direct formalization of this second goal, designed to select features that are highly relevant to the outcome while being minimally redundant with each other. Such pre-processing strategies remain valid for their original scope, as they reflect pragmatic concerns tied to the capabilities of traditional modeling methodologies which struggle to extract useful signal from high-D, collinear, error-prone data.

While this conventional paradigm remains exceptionally influential in many ML applications, its tenets are not the only viewpoints. For example, Psychometrics has long embraced an alternative philosophy -- Classical Test Theory (CTT) \citep{lord1968} and Item Response Theory (IRT) \citep{hambleton1991} are built on the principle of aggregating multiple, intercorrelated, imperfect indicators to attenuate idiosyncratic errors and yield reliable estimates of latent constructs. Within these measurement-centric fields, a high-D set of distinct indicators with shared variance is not a nuisance but is considered vital evidence of a common latent construct. Importantly, in these applications the latent construct is the target of inference; the goal is to achieve a precise estimate and interpretation of an unobserved trait (e.g., intelligence). We adapt this measurement machinery for a different objective: robustly predicting a downstream, observable outcome -- our ultimate focus is on the information the latent factors carry about the outcome.

This Psychometric perspective highlights a critical distinction often missed by the GIGO paradigm: the difference between reliability and validity. Effectively, Depth strategies improve the reliability of specific indicators (ensuring measurement is precise and consistent) but do not necessarily improve construct validity (ensuring measurement comprehensively captures the driving constructs). If a fixed set of indicators is structurally deficient (e.g., missing an aspect of the latent drivers), cleaning those indicators yields a precise measurement of an incomplete picture -- Structural Uncertainty represents this incompleteness. Partitioning predictor-space noise in this manner readily clarifies that no amount of effort dedicated solely to indicator reliability can compensate for a structural deficit. Conversely, by expanding the predictor set, the deficit in validity is reduced (i.e., Structural Uncertainty moves towards zero).

Also in contrast to the conventional paradigm, \cite{donoho2000} juxtaposed the Curse of Dimensionality with its ``blessing,'' arguing that the same high-D settings that challenge classical modeling methods can also exhibit regularity. This blessing is rooted in the mathematical phenomenon of concentration of measure, which dictates that in high-D, certain random quantities are tightly clustered, leading to predictable geometric structures \citep{vershynin2018high}. This has been invoked to explain the enhanced linear separability of data in high-D spaces and the remarkable generalization of overparameterized models \citep{gorban2018, tan2023}, often formalized in modern contexts via the Manifold Hypothesis. A key contribution of our framework (Section~\ref{Interplay of N and m}) is to demonstrate how, under the proposed latent architecture, sample size requirements are fundamentally mitigated by high-D, resolving the dissonance between the theoretical benefits of high-D Breadth and the pragmatic limits of finite samples.

Closely related to the Manifold Hypothesis is the theory of Representation Learning \citep{bengio2013representation}. While the Manifold Hypothesis posits that high-D data concentrate near low-dimensional structures, Representation Learning describes the mechanism by which deep models explicitly disentangle these structures. The central premise is that raw data (e.g., pixels or noisy indicators) are generated by the interaction of latent, semantic factors. A successful model must ostensibly invert this generative process, mapping the observed high-D surface back to the underlying low-D latent variables \citep{goodfellow2016}. In the context of our framework, this establishes the theoretical possibility that a sufficiently capable model can recover the primary latent layer (\(S^{(1)}\)) from the observed predictors (\(S^{\prime(2)}\)), provided the data architecture retains sufficient structural redundancy to make this inversion mathematically feasible.

While the Manifold Hypothesis addresses the geometry of high-D data, recent theoretical work on BO has focused on its spectral properties. This manuscript also furthers this line of inquiry by defining a specific data-generating structure that aligns with the Manifold Hypothesis, and showing that this structure gives rise to the ``favorable'' spectral properties linked to BO. Our analysis converges with and extends recent work linking factor models to BO \citep{bing2021, bunea2022}. While these works focus on risk analysis of interpolating predictors, our contribution is to synthesize this result into our broader framework, where highlighting this connection suggests a single architecture as a unified origin for robustness to both Outcome Error and predictor-space noise.

Lastly, we situate G2G within statistical domains that grapple with latent constructs and measurement error. The primary structure we analyze can be specified as a Structural Equation Model (SEM) \citep{bollen1989, kline2015}, and it shares foundational concepts with Measurement Error Modeling/Errors-in-Variables Models (MEM/EIV) \citep{carroll2006}. Specifically, SEM provides the established formalism for defining Structural Uncertainty as the probabilistic gap between latent drivers and observed variables. However, our framework differs from these traditions in two critical ways. First, traditional SEM estimation methods often struggle to scale to the high-D environments we address. Second, MEM/EIV and SEM view latent structures through the lens of covariance, aiming primarily for consistent parameter estimation and causal confirmation. In contrast, our core arguments adopt a fundamentally different analytic perspective. Instead of minimizing covariance discrepancies, we apply a more general Information-Theoretic lens to explain how the data's architecture can be exploited to achieve robust prediction of \(Y\) in the face of both Predictor Error and Structural Uncertainty. Furthermore, while our framework shares the foundational Econometric premise that observed variables serve as imperfect proxies for unobserved latent constructs (a core concern of EIV models) \citep{wooldridge2010econometric}, we diverge from that tradition's primary focus on leveraging these proxies for causal parameter identification, focusing instead on leveraging high-D redundancy to maximize predictive robustness.

\section{Framework Model and Notation}
\label{Framework Model and Notation}

Having established the theoretical context, we now formally define the data-generating structure that forms the foundation of our analysis. The framework uses a latent variable, data-generating structure. Figure~\ref{fig: Primary Structure} provides a visual representation of ``the primary structure'' (\(Y \leftarrow S^{(1)} \rightarrow S^{(2)} \rightarrow S^{\prime(2)}\)) where a set of latent first-stage states (\(S^{(1)}\)) acts as a common cause for both the outcome (\(Y\)) and a set of second-stage predictor states (\(S^{(2)}\)) which are observed with error as \(S^{\prime(2)}\).

We note that the principle of conceptualizing latent concepts as drivers of observable, correlated variables is a cornerstone across numerous applied disciplines outside of the methodological fields discussed in Sections~\ref{Intro} and \ref{Background} \citep{pearl2009causality}. In medicine, for instance, a latent disease process like Metabolic Syndrome is understood not as a single measurable entity, but as a high-level construct that gives rise to a cluster of observable, correlated conditions like insulin resistance and hypertension. Similarly, in macroeconomics, a Business Cycle is inferred from a portfolio of co-moving indicators such as industrial production and employment rates. In conceptual alignment with the Manifold Hypothesis, such latent drivers are not considered mere statistical conveniences, but are understood to reflect an underlying reality. We have chosen to utilize a latent structure to reflect this same understanding, thereby enhancing G2G's plausible applicability across a range of domains.

\subsection{Generation of Latent States}

A set of \(k\) binary, latent, first-stage states is defined as \(S^{(1)} = \{ X_{1}^{(1)},\ldots,X_{k}^{(1)}\}\). These states are the common causes in the framework. We assume \(k\) is fixed and finite \((1 \leq k < \infty)\) for a given application.

The states in \(S^{(1)}\) jointly generate a binary outcome \(Y\), where the process is governed by fixed conditional probabilities of \(Y\) taking a certain value (e.g., \(0\) or \(1\)) given the \(p^{\text{th}}\) configuration of all states in \(S^{(1)}\), \(s_{p}\). We assume variation in these conditional probabilities exists across configurations, as this ensures \(S^{(1)}\) carries information pertinent to \(Y\) (i.e., \(I(Y;S^{(1)}) > 0\)).

\[P(Y = 1 | s_{p}) = \delta_{s_{p} \rightarrow Y}\]
\[P(Y = 0 | s_{p}) = 1 - \delta_{s_{p} \rightarrow Y}\]

{\raggedright
\vspace{0.5\baselineskip}
There is a maximum of \(2^{k}\) conditional probabilities for \(Y\), one for each possible configuration of \(S^{(1)}\). In practice, the number of probabilities is determined by the number of configurations realized with a non-zero probability, \(K_{rlzd}\). Additionally, a more practical measure of the latent space's complexity is the ``effective'' number of configurations, \(K_{eff}\). This measure, formally defined by the joint entropy of the latent states (\(K_{eff} = 2^{H(S^{(1)})}\)), quantifies the statistical spread of the distribution (see Appendix~\ref{latent sparsity}). A highly correlated or sparse \(S^{(1)}\) layer will have a \(K_{eff}\) that is substantially smaller than both \(K_{rlzd}\) and the theoretical maximum of \(2^k\). This low ``Effective Latent Complexity'' is a key property we will leverage in our analysis of statistical feasibility (Section~\ref{Interplay of N and m}) and the connection to BO (Section~\ref{Benign Overfitting Link: Towards a Unified Understanding of Robustness}).}

The states in \(S^{(1)}\) also jointly generate a set of \(m\) distinct, binary, true, underlying, second-stage predictor states \(S^{(2)} = \{ X_{1}^{(2)},\ldots,X_{m}^{(2)}\}\). The size of this set, \(m\), can be arbitrarily large (potentially infinite). In an identical manner to the generation of \(Y\), the generation of \(X_{j}^{(2)}\) (the \(j^{\text{th}}\) state in \(S^{(2)}\)) from \(S^{(1)}\) is governed by fixed conditional probabilities of \(X_{j}^{(2)}\) taking a certain value (e.g., 0 or 1) given \(s_{p}\).

\[P(X_{j}^{(2)} = 1 | s_{p}) = \gamma_{s_{p} \rightarrow X_{j}^{(2)}}\]
\[P(X_{j}^{(2)} = 0|s_{p}) = 1 - \gamma_{s_{p} \rightarrow X_{j}^{(2)}}\]

{\raggedright
\vspace{0.5\baselineskip}
Again, while there are a maximum of \(2^{k}\) conditional probabilities for \(X_{j}^{(2)}\), the actual number required is \(K_{rlzd}\). Let \(\varphi_{S^{(1)} \rightarrow X_{j}^{(2)}}\) be the set of \(K_{rlzd}\) conditional probabilities for \(X_{j}^{(2)}\), where the \(p^{\text{th}}\) value in the set is associated with \(s_{p}\). As with \(Y\), we assume values in \(\varphi_{S^{(1)} \rightarrow X_{j}^{(2)}}\) are \(> 0.0\) and \(< 1.0\), making the model explicitly non-deterministic/probabilistic. The strength of the relationship between the \(S^{(1)}\) states and \(X_{j}^{(2)}\) depends on how the values in \(\varphi_{S^{(1)} \rightarrow X_{j}^{(2)}}\) vary across the \(K_{rlzd}\) configurations.}

Crucially, we label the resulting uncertainty in the predictor-space produced by the probabilistic \(S^{(1)} \rightarrow S^{(2)}\) step as ``\textbf{Structural Uncertainty}.'' At the set level, Structural Uncertainty refers to the ambiguity that remains in the recovery of \(S^{(1)}\) given the feature set of interest is error-free, which is determined by how well the set of predictors covers the latent layer. This keystone concept to our theory acts as a source of inherent noise in the system that obscures the latent signal, a structural ambiguity that remains even if measurements were perfect.

The condition \(k \ll m\) is assumed to hold in most domains, where a substantially smaller set of primary latent states is understood to drive a significantly larger set of predictors.

\subsubsection{``Causal Consistency'' and Parametric Constraints}
\label{Causal Consistency and Parametric Constraints}

We posit that in many contexts, the conditional probabilities in \(\varphi_{S^{(1)} \rightarrow X_{j}^{(2)}}\) that generate \(Y\) and \(S^{(2)}\) are likely parametric functions of the latent \(S^{(1)}\) states, where this function determines the parameters of the distributions from which the specific conditional probabilities for each generated variable are drawn. When such functions are present, this produces ``Causally Consistent Variables'' -- where \(Y\) and \(S^{(2)}\) variables tend to predictably reflect the dynamics of the \(S^{(1)}\) states. While Causal Consistency here does not imply the parametric function is strictly linear, we posit that in natural systems, these functions are typically monotonic (or near-monotonic) -- they predominantly possess stable directional influences (e.g., the presence of a latent risk factor consistently increases the probability of an adverse indicator, even if the rate of increase varies). Regardless of the function, however, consistency here is more generally used to imply that the rules governing the relationship between \(S^{(1)}\), \(\delta_{s_{p} \rightarrow Y}\), and \(\varphi_{S^{(1)} \rightarrow X_{j}^{(2)}}\) are stable.

We also clarify that Causal Consistency is a within-variable property, not a between-variable constraint. Within-variable consistency implies that each generated variable follows a stable, parametric rule in its response to \(S^{(1)}\) -- effectively, that the variable has a consistent internal logic. It does not imply that different variables must behave similarly to one another; for example, \(X_{1}^{(2)}\) may respond linearly to a latent driver, while \(X_{2}^{(2)}\) responds inversely, and \(X_{3}^{(2)}\) behaves as a non-linear threshold detector. 

Conversely, if we assume that the conditional probabilities in \(\varphi_{S^{(1)} \rightarrow X_{j}^{(2)}}\) are fully independent random draws for each configuration, this implies a variable devoid of Causal Consistency -- one that effectively treats the generative process as a distinct, arbitrary ``lookup table'' for every \(s_{p}\). In such a ``Chaotic Generative Variable,'' the relationship lacks topological structure -- the presence of a latent driver in one configuration (e.g., \(X_1^{(1)}\) in \(\{1,0\}\)) provides no predictive information about its effect in a neighboring configuration (e.g., \(\{1,1\}\)). This effectively decouples such a variable from the individual \(S^{(1)}\) states, instead tying it solely to the \(S^{(1)}\) configurations.

Intuitively, this can be visualized as the difference between a smooth manifold and a fractured surface. In a Causally Consistent Variable, if two latent configurations are close in the latent space, their generated outputs generally will be statistically similar. The parametric function defines the shape of this manifold. For a Chaotic Generative Variable, the smooth manifold is replaced with a fractured surface consisting of discrete neighborhoods; moving to the nearest neighbor in latent space results in a completely uncorrelated output distribution, much like jumping between unrelated indices in a hash table.

Critically, the validity of the core theoretical arguments presented throughout this work -- specifically the information-theoretic derivation of the benefits of Breadth (Statements 1-3) and the statistical analysis of efficiency -- do not depend on the presence of Causally Consistent Variables. The principles of Information Theory, such as the Data Processing Inequality and the sub-additivity of entropy, govern the flow of information through the primary structure \(Y \leftarrow S^{(1)} \rightarrow S^{(2)}\) regardless of the function defining the conditional probabilities. Whether the generation process is highly structured at the state level, effectively random at the configuration level, or mixed, the principles of Information Theory apply universally. Instead, the concept of a Causally Consistent Variable serves only to explain why such variables arise in nature and to justify specific methodological choices (like Factor Analysis) without restricting the applicability of the broader data-architectural bounds.

Major implications of Causal Consistency are discussed in Sections \ref{Assessing Framework Applicability}, \ref{Benefits of Causal Consistency}, and \ref{The Generative Origin of Low Rank: Effective Latent Complexity and Causal Consistency}.

\subsection{Conditional Independence of Second-Stage Predictor States (Local Independence, LI)}

The model implies \(P(S^{(2)} | S^{(1)}) = \prod_{j = 1}^{m}{P(X_{j}^{(2)}|S^{(1)})}\). The common cause structure \(Y \leftarrow S^{(1)} \rightarrow S^{(2)}\) in turn implies the Markov property \(Y \bot S^{(2)}|S^{(1)}\).

While LI is critical for the clarity of our core analysis, we discuss in Section~\ref{Limitations and a Future Research Agenda} and demonstrate in Appendix~\ref{equivalence for LI viol} that the framework's principles can theoretically be extended to contexts where this assumption is violated, provided a sufficiently flexible model is employed.

\subsection{Baseline (Independent) Error Model for Second-Stage States (Defining \(S^{\prime(2)}\))}

Conceptually, let ``Observational Error'' be discrepancies between true and observed values due to various sources, including instrument/measurement limitations, human error, data gaps, or rater biases. \(X_{j}^{(2)}\) is observed with Observational Error as \(X_{j}^{\prime(2)}\). Focusing on the context where each \(X_{j}^{(2)}\) is observed once as \(X_{j}^{\prime(2)}\), the size of the set of observed second-stage variables, \(S^{\prime(2)}\), is also \(m\).

Observational ``fidelity'' in the second stage is defined by fixed conditional probabilities of \(X_{j}^{\prime(2)}\) taking a certain value (e.g., 0 or 1) given a specific true value of \(X_{j}^{(2)}\). Assuming the baseline model, errors are conditionally independent across observed predictors given \(S^{(2)}\).

\[\alpha_{X_{j}^{\prime(2)}} = \alpha_{j} = P(X_{j}^{\prime(2)} = 1 | X_{j}^{(2)} = 1)\]
\[\beta_{X_{j}^{\prime(2)}} = \beta_{j} = P(X_{j}^{\prime(2)} = 0 | X_{j}^{(2)} = 0)\]

\vspace{0.5\baselineskip}
The \(\alpha\) and \(\beta\) parameters correspond to sensitivity and specificity, respectively, and can be conceptualized as draws from distributions ranging \(> 0.0\) to \(< 1.0\), making perfect observational fidelity strictly unachievable. While Observational Error conceptually applies to both outcomes and predictors, our framework directly focuses on the predictor-space. We therefore refer to the uncertainty produced by the \(S^{(2)} \rightarrow S^{\prime(2)}\) links specifically as ``Predictor Error.''

When \(\alpha_{j} = \beta_{j} = 0.5\), Predictor Error completely ``saturates'' \(X_{j}^{\prime(2)}\), fully masking the true \(X_{j}^{(2)}\) signal. High fidelity (low error) occurs as these values both approach \(1.0\). High inverted fidelity (i.e., high stable error) occurs as these values both approach \(0.0\) -- in this scenario, the signal from \(X_{j}^{(2)}\) is largely preserved in \(X_{j}^{\prime(2)}\), albeit in an inverted form.

\subsection{Taxonomy of Uncertainty: Defining ``Noise,'' ``Dirty,'' and ``Clean''}
\label{Taxonomy of Uncertainty}

To avoid ambiguity in subsequent discussions, we now formally define our terminology for describing uncertainty:

\begin{itemize}
    \item \textbf{Noise:} We utilize ``noise'' as an umbrella term that encompasses all forms of variance that obscure the relationship between the predictors and the outcome. This includes both Observational Error and Structural Uncertainty. 
    \item \textbf{Dirty vs. Clean:} We reserve the terms ``dirty'' and ``clean'' to refer strictly to the magnitude of Observational Error.
    \begin{itemize}
        \item \textbf{Dirty:} Refers to data with low observational fidelity (high Observational Error).
        \item \textbf{Clean:} Refers to data with high observational fidelity (low Observational Error).
    \end{itemize}
\end{itemize}

Crucially, under these definitions, a dataset can be clean (perfectly measured) yet still be noisy (due to Structural Uncertainty).

\subsection{Framework Flexibility and Scope}

The primary structure can accommodate a variety of scenarios. This allows for broad representation and enhances its applicability to diverse data-generating processes. For example, the structure allows for both correlated and uncorrelated \(S^{(1)}\) states. Additionally, the generation of \(S^{(2)}\) can accommodate varied dependency structures. While \(X_{j}^{(2)}\) is described as being generated from the complete set of \(S^{(1)}\) states via \(\varphi_{S^{(1)} \rightarrow X_{j}^{(2)}}\), if \(X_{j}^{(2)}\) is influenced by only a subset of \(S^{(1)}\) states (or even just a single \(X_{i}^{(1)}\)), as is likely posited in many applications, the values within \(\varphi_{S^{(1)} \rightarrow X_{j}^{(2)}}\) can flexibly reflect this sparsity.

Furthermore, while the core derivations in this manuscript utilize binary variables for tractability, the information-theoretic principles regarding uncertainty reduction generalize to continuous variables (see Appendix~\ref{continuous vars}).

Finally, we also explicitly analyze the framework's robustness across various causal topologies (e.g., mediation, common cause), identifying a key misalignment with ``mid-stream'' collider structures where \(S^{(1)}\) is caused by \(Y\) and \(S^{(2)}\) (see Appendix~\ref{Robustness of the Framework Across Alt Causal Structures}).

\subsection{The Source of ``Informative Collinearity'': Unconditional Correlation Arising in \(S^{(2)}\)}

Reichenbach's Common Cause Principle states that if two variables share a common cause (and are not a direct cause or effect of each other), they will be correlated unless very specific canceling conditions occur \citep{reichenbach1956}. Thus, though LI results from the primary structure (Section~\ref{Framework Model and Notation}), this also implies that unconditional correlations among \(S^{(2)}\) states will likely arise due to their shared sources in \(S^{(1)}\).

We use the term ``Informative Collinearity'' to describe these specific correlations when they manifest in \(S^{\prime(2)}\). In the language of Psychometrics and Factor Analysis, this concept is equivalent to the shared variance (or common variance) that arises among indicators influenced by the same latent factor \citep{spearman1904, Thurstone1947}. We adopt the term Informative Collinearity to deliberately reframe this phenomenon as a valuable signal for robustness, contrasting it with both trivial redundancies and the generally negative view of collinearity in the conventional paradigm.

Crucially, it is the combination of this shared causal origin (driving correlation) and the conditional independence of the realizations of uncertainty in the generative pathways and error mechanisms that ensures these ``distinct'' indicators provide conditionally unique information, allowing the collinearity to be ``informative'' rather than ``trivial.'' Examples of trivial collinearity include predictors that are simple duplicates, linear transformations, or linear combinations of one another. In contrast to Informative Collinearity, such redundancies can be formally defined as trivial because they offer no conditional information gain (as discussed in Section~\ref{Core Theoretical Analysis}).

\subsection{Assessing Framework Applicability: From Empirical Patterns to Plausibility}
\label{Assessing Framework Applicability}

Formal statistical validation of the \(Y \leftarrow S^{(1)} \rightarrow S^{(2)} \rightarrow S^{\prime(2)}\) structure is inherently challenging in practice because Predictor Error confounds the results of standard diagnostic tests (e.g., global fit indices such as CFI and RMSEA; \citealp{hu1999cutoff}). This creates a critical practical paradox: the robustness mechanisms we highlight in this manuscript are often valuable in the situations where core assumptions are hardest to verify empirically. Therefore, until robust, formal diagnostic methods are developed (a critical direction for future research), a pragmatic approach is required. As a preliminary assessment, a two-step heuristic process that combines data-driven exploration with domain-specific knowledge can be utilized.

\begin{enumerate}
\item 
Empirically look for evidence of a non-random, clustered correlation structure.

The existence of distinct clumpy collinearity or correlational clusters -- where some groups of predictors are more highly correlated with each other than with predictors outside the group -- is potentially the empirical footprint of an underlying \(S^{(1)}\) layer. Each cluster suggests a shared common cause, which is precisely what the presence of an \(S^{(1)}\) layer implies. Conversely, the absence of such distinct clusters may suggest that a simpler, flat model (where predictors are assumed to directly cause \(Y\)) is a more appropriate representation of the data -- one that aligns more closely with traditional, less flexible predictive models.

This exploration can be performed using visual methods like reordered correlation matrix heatmaps \citep{friendly2002corrgrams} or spectral heuristics. We note that systematic errors and direct causal links among \(S^{(2)}\) states (thereby violating LI) can also induce correlational clusters. Related discussion is provided in Sections \ref{Benign Overfitting Link: Towards a Unified Understanding of Robustness} and \ref{Violating LI}, as well as in Appendix~\ref{equivalence for LI viol}.

\begin{raggedright}
\textbf{Spectral Heuristic (The Scree Test):} A primary test stems from the framework's connection to BO (Section~\ref{Benign Overfitting Link: Towards a Unified Understanding of Robustness}), where we show that the primary structure naturally produces a ``low-rank-plus-diagonal'' covariance matrix. A practitioner can calculate a scree plot of \(S^{\prime(2)}\) to estimate the ``Effective Signal Rank'' (\(r\)) via the location of the plot's elbow.
\end{raggedright}

\begin{itemize}
\item
\textbf{Interpretation:} The elbow represents the dimensionality of the signal subspace that propagates from \(S^{(1)}\). In this framework, \(r\) serves as a proxy for the Effective Latent Complexity (\(K_{eff}\)) -- the effective number of distinct latent configurations the model must resolve.
\item
\textbf{Diagnostic:} A clear elbow followed by a flat tail is consistent with a latent structure that may be suitable for G2G, though this spectral signature can arise from other data-generating processes as well. The lack of an elbow (linear decay) suggests the data lack the requisite structural redundancy or that \(N\) and/or \(m\) are insufficient to distinguish signal from noise.
\item
\textbf{Note on Linearity:} While standard scree plots measure linear covariance, and the underlying generative process may be non-linear, we demonstrate in Section~\ref{The Generative Origin of Low Rank: Effective Latent Complexity and Causal Consistency} why this linear metric remains a valid heuristic for estimating the number of distinct latent configurations the model must resolve.
\end{itemize}

\begin{raggedright}
\textbf{Model-Based Performance Heuristic:} Practitioners can also employ a model-based performance heuristic to indirectly examine evidence of a latent hierarchy. This involves performing a direct, competitive model comparison. As our analysis in Section~\ref{Primacy of the S(1) Layer and the Role of Interactions} argues, optimally exploiting the G2G architecture requires flexible models capable of Representation Learning and interaction modeling. A practitioner can therefore compare the cross-validated predictive performance of a model that should be able to better capture this hierarchy (e.g., an autoencoder-based model) against a flat model that should struggle to capture this structure. A significant and robust performance gap, where the hierarchical model substantially outperforms the flat model, implies that the latent \(S^{(1)}\) structure is not just present, but is critical for predictive accuracy.
\end{raggedright}

\item
Assess if any uncovered empirical patterns align with domain theory.

This can be guided by two lines of questions. (a) Do the clusters correspond to known latent constructs? Does domain theory already posit the existence of core underlying concepts that would explain the observed clusters? (b) Is the hierarchical structure plausible? This involves comparing the plausibility of a hierarchical structure against a flat, non-\(S^{(1)}\) alternative.
\end{enumerate}

When the results of both assessments point toward a hierarchical latent structure, a researcher has a reasonable justification for proceeding.

\subsection{Glossary of Terms and Concepts}
\label{Glossary}

\begin{longtable}{@{} >{\small\bfseries\raggedright}p{0.40\linewidth} p{0.54\linewidth} @{}}
\caption{Glossary of Framework Terminology and Concepts.}\\
\toprule
\normalsize\textbf{Term/Concept} & \normalsize\textbf{Definition} \\
\midrule
\endfirsthead
\normalsize\textbf{Term/Concept} & \normalsize\textbf{Definition} \\
\midrule
\endhead
\bottomrule
\endfoot

\multicolumn{2}{c}{\textbf{Structural Architecture}} \\
\midrule
Primary Structure & The hierarchical data-generating process \(Y \leftarrow S^{(1)} \rightarrow S^{(2)} \rightarrow S^{\prime(2)}\), where the \(S^{(1)}\) layer drives both the outcome and the predictors. \\
\(Y\) (Outcome/Label/Target) & The outcome being predicted. In this framework, it is driven solely by the latent layer \(S^{(1)}\). \\
\(S^{(1)}\) (The Layer of First-Stage States/Latent States/Latent Drivers) & The set of \(k\) latent, unobserved drivers (causes) of the system. \\
\(s_p\) (Latent Configuration of \(S^{(1)}\)) & A specific combination of values of the \(k\) Latent States in \(S^{(1)}\) (e.g., \(\{1, 0, 1\}\)). There are \(2^k\) possible configurations, but only \(K_{rlzd}\) are realized. \\
\(S^{(2)}\) (The Layer of Second-Stage States/True Predictors/True Features) & The set of \(m\) true, underlying predictor states generated by \(S^{(1)}\). These are the ``perfect'' versions of the predictor variables before Observational Error. \\
\(S^{\prime(2)}\) (The Layer of Observed Predictors/Observed Features) & The set of \(m\) observed predictor variables -- the true predictors after being corrupted by Observational Error. \\
Chaotic Generative Variable & A variable where the relationship between \(S^{(1)}\) and \(S^{(2)}\) is random for every latent configuration (a ``lookup table''), lacking stable directional rules. Note: The presence of these types of variables is initially assumed for all core analyses to demonstrate G2G mechanisms under the worst-case scenario. \\
Causal Consistency & A property where an \(S^{(2)}\) variable is generated by a parametric function of \(S^{(1)}\) states. Strict linearity is not required, but we posit that in natural systems, these functions are typically monotonic (or near-monotonic), predominantly possessing stable directional influences. This allows predictors to act as proxies for latent states directly rather than unique identifiers for configurations of latent states. \\

\midrule
\multicolumn{2}{c}{\textbf{Uncertainty and Noise Taxonomy}} \\
\midrule
Noise & An umbrella term that encompasses all forms of variability that obscure the relationship between the observed predictors and the outcome. In the predictor-space, this includes both Predictor Error and Structural Uncertainty. \\
Observational Error & Discrepancies introduced during the measurement process (\(S^{(2)} \rightarrow S^{\prime(2)}\)), such as typos, sensor noise, or recording errors. G2G focuses on Observational Error that affects predictors (Predictor Error), but this concept applies to the outcome as well (Outcome Error). \\
Structural Uncertainty & The probabilistic ambiguity inherent in the \(S^{(1)} \rightarrow S^{(2)}\) generation step. The degree to which even a perfectly measured predictor variable is an imperfect proxy for the latent drivers. At the set level, it refers to the ambiguity remaining about \(S^{(1)}\) even when all predictor variables in the set are error-free.\\
Dirty Data & Data characterized by low observational fidelity (high Observational Error). \\
Clean Data & Data characterized by high observational fidelity. Note: clean data may still be noisy due to Structural Uncertainty. \\
Flat Topological Operationalization & A model operationalization that collapses Predictor Error and Structural Uncertainty into a single monolithic error term. This confines optimization strictly within a single probabilistic graphical model, implicitly treating observed predictors as ground-truth inputs that map directly to the outcome. This ignores the underlying latent hierarchy and obfuscates the distinction between data errors and informational deficits. \\
Systematic Error Regime (SERg) & A scenario where a subset of observed predictors share common Observational Error parameters (governed by a latent regime \(L\)), while error realizations remain conditionally independent given the regime. \\

\midrule
\multicolumn{2}{c}{\textbf{Strategies and Properties}} \\
\midrule
Breadth & A strategy of adding more distinct predictors (increasing \(m\)) to the predictor set. Breadth is the only mechanism capable of reducing Structural Uncertainty. \\
Depth & A strategy of improving the fidelity of a fixed set of predictors (e.g., repeated measures, cleaning). Depth can eliminate Observational Error but prediction remains bounded by the Structural Uncertainty inherent in the fixed set. \\
Performance Floor & The fundamental limit on predictive accuracy for a fixed set of predictors, determined by the set's Structural Uncertainty. Depth strategies cannot pass this floor; only Breadth can lower it. \\
Prevalence Floor & The lower bound on statistical feasibility in a high-D system. As \(m\) increases, predictor-space degeneracy vanishes, ensuring the sample size required for latent inference is limited by the prevalence of the latent state (\(P(s_{p})\)), not the rarity of the specific predictor pattern. \\
Absorption Hypothesis & The proposition that a sufficiently flexible model can mitigate violations of Local Independence or Systematic Error by implicitly learning an expanded latent representation (including pseudo-states) that restores conditional independence. \\
Distinctness & A statistical property where the realizations of uncertainty for a predictor (both Structural and Observational) are conditionally independent of those for other predictors. This is the prerequisite for Information Aggregation. \\
Structural Strength & The discriminative power of the \(S^{(1)} \rightarrow S^{(2)}\) link. Determined by how strongly the true predictor is influenced by the latent driver, distinct from observational fidelity. \\
Informative Collinearity & Correlations among predictors arising from shared latent causes in \(S^{(1)}\). An asset that improves reliability and convergence efficiency in recovering \(S^{(1)}\). \\
Trivial Collinearity & Redundancy in predictors arising from duplicates, linear combinations, or deterministic transformations of aspects of the predictor-space. Unlike Informative Collinearity, it offers no conditional information gain because uncertainty realizations are dependent. \\
Informative Redundancy & When a distinct predictor shares a similar functional relationship (\(\gamma\)) to \(S^{(1)}\) as an existing predictor, reinforcing the signal to improve reliability. \\
Novelty & When a predictor has a different functional relationship to \(S^{(1)}\) than the existing set, providing coverage of previously unmapped latent space (Completeness). \\
Proactive Data-Centric AI (P-DCAI) & A strategy of designing data acquisition and feature selection to maximize architectural robustness (Breadth, Novelty, Informative Redundancy) rather than solely minimizing Observational Error or prioritizing bivariate mutual information between \(Y\) and selected features. \\

\midrule
\multicolumn{2}{c}{\textbf{Rank and Spectral Definitions}} \\
\midrule
Effective Latent Complexity (\(K_{eff}\)) & The entropic ``spread'' of the latent layer \(S^{(1)}\). It quantifies the effective number of configurations the latent system actually assumes. Constrained by prevalence and correlation of the \(S^{(1)}\) states. \\
Algebraic Rank & The strict mathematical dimension of the signal matrix. In non-linear systems, this can be very high (bounded by \(K_{rlzd}-1\)). \\
Effective Signal Rank, \(r\) & The statistical dimensionality of the ``signal subspace,'' observed as the elbow in a scree plot. With Chaotic Generative Variables, this strictly scales with \(K_{rlzd}\) and effectively with \(K_{eff}\). Under monotonic Causal Consistency, this collapses towards \(k\). \\
Spectral SNR & The ratio of the smallest signal eigenvalue to the largest noise eigenvalue (\(\lambda_r/\lambda_{r+1}\)). It quantifies the distinctness of the latent structure (the ``eigengap'') in the covariance matrix. \\
Malignant Overfitting & A critical failure mode where common systematic error affects both the outcome and predictor variables (e.g., Common Method Variance), creating a pseudo-state correlated with both, causing the model to learn a bias as a predictive signal. \\
\midrule
\multicolumn{2}{c}{\textbf{Deployment Paradigms}} \\
\midrule
Model Transfer & The ML deployment paradigm of exporting a static, pre-trained model (fixed weights and parameters) from an original source environment to an external one. This approach relies on the assumption of external validity and often degrades when the external environment's data distribution or systematic error structures differ from the source. \\
Methodology Transfer & The ML deployment paradigm where the methodological blueprint is exported rather than the model parameters. This strategy operationalizes a ``Local Factory'' -- a system that ingests local data -- to generate a bespoke model optimized for the specific latent structures and artifactual quirks of the deployment site. \\
Intersectional Constraint & A theoretical limitation inherent to universal (foundation) models. Because such models must be transportable across diverse sites, they are mathematically restricted to using only the intersection of feature sets across those sites. This constraint forces the removal of site-specific predictors, reducing dimensionality and thereby raising the floor of Structural Uncertainty relative to a Local Factory that can utilize the full local set. \\
\bottomrule
\end{longtable}

\section{Core Theoretical Analysis}
\label{Core Theoretical Analysis}

Having defined the framework's structure and terminology, we now analyze its information-theoretic properties. The core analysis evaluates the properties of the primary structure from first principles. We use the foundational concepts of conditional entropy (\(H(A|B)\)) and conditional mutual information (\(I(A;B|C)\)) \citep{cover2006} to formalize the theoretical impact of predictor counts and quality on predictive uncertainty (\(H(Y| \bullet)\)). This analysis focuses on the information content of the data architecture, implicitly assuming sufficient sample size to realize these theoretical limits. The interplay between sample size and dimensionality is analyzed in Section~\ref{Interplay of N and m}.

The following list details the assumptions required for the core analysis (all of Section~\ref{Core Theoretical Analysis}).

\begin{flushleft}
\textbf{A. Structural Assumptions}
\end{flushleft}

\begin{enumerate}
\item
\textbf{The Primary Structure:} The data-generating process adheres to the \(Y \leftarrow S^{(1)} \rightarrow S^{(2)} \rightarrow S^{\prime(2)}\) hierarchical structure. This defines the information flow, implies the Markov properties (e.g., \(Y\bot S^{(2)}|S^{(1)}\)), and includes:
\begin{enumerate}
\item
\textbf{Local Independence (LI):} The second-stage states are conditionally independent of each other given the first-stage states.
\end{enumerate}
\item
\textbf{Finite Latent Dimensionality:} The number of first-stage states is finite (i.e., \(1 \leq k < \infty\)).
\item
\textbf{Low Dimensionality of \(S^{(1)}\) Relative to \(S^{(2)}\):} The number of first-stage states is substantially smaller than the number of second-stage states (i.e., \(k \ll m\)).
\item
\textbf{Baseline (Independent) Error Regime:} Predictor Error is conditionally independent across predictors.
\end{enumerate}

\begin{raggedright}
Note: Assumptions 1.a and 4 are critical for the factorization of probabilities and the analysis of information aggregation.
\end{raggedright}

\begin{flushleft}
\textbf{B. Informational Assumptions:} These ensure the system is probabilistic and the relationships are meaningful.
\end{flushleft}

\begin{enumerate}
\def\labelenumi{\arabic{enumi}.}
\setcounter{enumi}{4}
\item
\textbf{Non-Determinism:} The generative links (\(Y \leftarrow S^{(1)}\), \(S^{(1)} \rightarrow S^{(2)}\), \(S^{(2)} \rightarrow S^{\prime(2)}\)) are probabilistic/non-deterministic.
\item
\textbf{Relevance of} \(S^{(1)}\) \textbf{States:} The first-stage states carry information pertinent to the outcome (i.e., \(I(Y;S^{(1)}) > 0\)).
\item
\textbf{Unsaturated Error:} Observational Error does not completely mask \(S^{(2)}\).
\end{enumerate}

\begin{flushleft}
\textbf{C. Asymptotic Assumptions:} Required specifically for the analysis of the Infinite Breadth limit (\(m \rightarrow \infty\)).
\end{flushleft}

\begin{enumerate}
\def\labelenumi{\arabic{enumi}.}
\setcounter{enumi}{7}
\item
\textbf{Identifiability of \(S^{(1)}\):} Different configurations of \(S^{(1)}\) produce distinct distributions over the infinite sequence of \(S^{(2)}\) states, such that the average KL divergence between these distributions is bounded away from zero (Appendix~\ref{asymp Breadth v Depth}).
\end{enumerate}

\subsection{Impact of Increasing the Number of \(S^{(2)}\) States and \(S^{\prime(2)}\) Variables}

\vspace{0.5\baselineskip}
\textbf{Statement 1:} \(H(Y|S^{(1)}, S^{(2)})\) and
\(H(Y|S^{(1)}, S^{\prime(2)})\) are invariant to \(m\).

\begin{quote}
\textbf{Argument for Statement 1:} Markov property \(Y\bot S^{(2)}|S^{(1)}\) implies \(I(Y;S^{(2)} | S^{(1)}) = 0\), so \(H(Y|S^{(1)}, S^{(2)}) = H(Y|S^{(1)})\). This property propagates to observed variables: \(H(Y|S^{(1)}, S^{\prime(2)}) = H(Y|S^{(1)})\).

\textbf{Conclusion for Statement 1:} All \(S^{(2)}\) states and \(S^{\prime(2)}\) variables are redundant for predicting \(Y\) given full knowledge of \(S^{(1)}\). Thus, increasing \(m\) cannot reduce uncertainty about \(Y\) beyond that already achieved by \(S^{(1)}\).
\end{quote}

\subsection{Impact of Predictor Properties on \(S^{(1)}\) Inference}

\vspace{0.5\baselineskip}
\textbf{Statement 2a:} \(H(X_{i}^{(1)}|S^{\prime(2)})\) is non-increasing with \(m\).

\begin{quote}
\textbf{Argument for Statement 2a:} Consider incorporating the \(({m + 1)}^{\text{th}}\) variable (\(X_{(m + 1)}^{\prime(2)}\)) into \(S^{\prime(2)}\) forming \(S_{(m + 1)}^{\prime(2)}\). The change in conditional entropy about \(X_{i}^{(1)}\) is \(H(X_{i}^{(1)} | S^{\prime(2)}) - H(X_{i}^{(1)} | S_{(m + 1)}^{\prime(2)}) = I(X_{i}^{(1)};X_{(m + 1)}^{\prime(2)}|S^{\prime(2)})\). Mutual information is non-negative. Therefore, \(H(X_{i}^{(1)} | S^{\prime(2)}) \geq H(X_{i}^{(1)}|S_{(m + 1)}^{\prime(2)})\).

\textbf{Conclusion for Statement 2a:} Each additional variable that carries conditionally unique information about \(X_{i}^{(1)}\) -- information not already present in \(S^{\prime(2)}\) -- will reduce the uncertainty about \(X_{i}^{(1)}\). Thus, the ability to recover \(X_{i}^{(1)}\) tends to improve as the number of \(S^{\prime(2)}\) variables increases. In turn, the ability to estimate \(S^{(1)}\) overall also tends to improve with \(m\).

We take this to its theoretical ideal in Appendix~\ref{asymp Breadth v Depth} in the limit of Infinite Breadth, where we show that as \(m \rightarrow \infty\), (1) an explicit error mitigating link arises, and (2) the Structural Uncertainty in the \(S^{(1)} \rightarrow S^{(2)}\) step is resolved. Specifically, we show that the asymptotic limit (as \(m \rightarrow \infty\) forming \(S_{\infty}^{\prime(2)}\)) of the conditional entropy is \(H(S^{(1)} | S_{\infty}^{\prime(2)}) = 0\). This is a critical result: as the number of distinct error-prone predictors approaches infinity, \(S^{(1)}\) can be perfectly recovered. Thus, when a Breadth strategy is utilized, high-D has the inherent potential to mitigate the impact of both Predictor Error and Structural Uncertainty, allowing for the perfect recovery of \(S^{(1)}\) despite the probabilistic nature of the \(S^{(1)} \rightarrow S^{(2)}\) step.

This result, derived here through an information-theoretic lens, converges with a key finding in the high-D statistics literature on latent factor models. Notably, \cite{fan2013embracing} formally proved that with a sufficiently large number of variables, latent factors can be estimated with a precision that matches an ``oracle'' procedure that has direct access to the factors themselves. Their work provides an independent confirmation of the ``Infinite Breadth'' principle -- that perfect asymptotic recovery of the latent layer is theoretically possible.
\end{quote}

\begin{flushleft}
\textbf{Statement 2b:} \(H(X_{i}^{(1)}|X_{j}^{\prime(2)})\) is non-increasing as the correlation between \(X_{i}^{(1)}\) and \(X_{j}^{(2)}\) strengthens.
\end{flushleft}

\begin{quote}
\textbf{Argument for Statement 2b:} The
\(\varphi_{S^{(1)} \rightarrow X_{j}^{(2)}}\) parameters contribute to determining the phi correlation between \(X_{i}^{(1)}\) and \(X_{j}^{(2)}\) (\(\phi(X_{i}^{(1)},X_{j}^{(2)})\)) -- holding other statistical properties of the variables constant, increasing the magnitude of the relevant \(S^{(1)} \rightarrow S^{(2)}\) links strengthens \(\phi(X_{i}^{(1)},X_{j}^{(2)})\) (see Appendix~\ref{phi corr}). The observation process (\(X_{j}^{(2)} \rightarrow X_{j}^{\prime(2)}\)) is a separate error mechanism. The variables form the Markov chain \(S^{(1)} \rightarrow X_{j}^{(2)} \rightarrow X_{j}^{\prime(2)}\). If the phi correlation between two binary variables strengthens (approaches \(-1\) or \(1\)), then mutual information increases \citep{Linfoot1957, cover2006}. Thus, for a fixed \(X_{j}^{(2)} \rightarrow X_{j}^{\prime(2)}\) error mechanism, and assuming unsaturated error, the end-to-end mutual information (\(I(X_{i}^{(1)};X_{j}^{\prime(2)})\)) increases as \(\phi(X_{i}^{(1)},X_{j}^{(2)})\) strengthens. Consequently, \(H(X_{i}^{(1)}|X_{j}^{\prime(2)})\) is non-increasing as \(\phi(X_{i}^{(1)},X_{j}^{(2)})\) strengthens.

It is crucial to recognize, however, that the magnitude of \(\phi(X_{i}^{(1)},X_{j}^{(2)})\) is not solely determined by the direct signal strength of the \(S^{(1)} \rightarrow S^{(2)}\) link. As Appendix~\ref{phi corr} shows, the observable correlation is a nuanced interplay between this signal strength and the base-rate prevalence of both the first-stage state (\(X_{i}^{(1)}\)) and the true predictor (\(X_{j}^{(2)}\)). This interplay has direct implications for feature selection, suggesting that a predictor's value is also determined by the statistical properties of the phenomena it measures.

\textbf{Conclusion for Statement 2b:} As the correlation between \(X_{i}^{(1)}\) and \(X_{j}^{(2)}\) strengthens -- an increase in ``Structural Strength'' -- \(X_{j}^{\prime(2)}\) becomes a better proxy for \(X_{i}^{(1)}\), reducing uncertainty about it. In turn, the ability to estimate \(S^{(1)}\) overall also tends to improve as Structural Strength increases.
\end{quote}

\begin{flushleft}
\textbf{Statement 2c:} \(H(X_{i}^{(1)}|X_{j}^{\prime(2)})\) is non-increasing as the fidelity of the \(X_{j}^{(2)} \rightarrow X_{j}^{\prime(2)}\) observational process strengthens.
\end{flushleft}

\begin{quote}
\textbf{Argument for Statement 2c:} The fidelity of the \(X_{j}^{(2)} \rightarrow X_{j}^{\prime(2)}\) observation process determines the phi correlation between \(X_{j}^{(2)}\) and \(X_{j}^{\prime(2)}\) (\(\phi(X_{j}^{(2)},X_{j}^{\prime(2)})\)). The variables form the Markov chain \(S^{(1)} \rightarrow X_{j}^{(2)} \rightarrow X_{j}^{\prime(2)}\). If \(\phi(X_{j}^{(2)},X_{j}^{\prime(2)})\) strengthens (approaches \(-1\) or \(1\)), then \(I(X_{j}^{(2)};X_{j}^{\prime(2)})\) increases. For a given \(\phi(X_{i}^{(1)},X_{j}^{(2)})\), mutual information \(I(X_{i}^{(1)};X_{j}^{\prime(2)})\) also increases as \(\phi(X_{j}^{(2)},X_{j}^{\prime(2)})\) approaches \(-1\) or \(1\). Consequently, \(H(X_{i}^{(1)}|X_{j}^{\prime(2)})\) is non-increasing as this observational fidelity increases.

\textbf{Conclusion for Statement 2c:} As the observational fidelity strengthens, \(X_{j}^{\prime(2)}\) becomes a better proxy for \(X_{i}^{(1)}\), reducing uncertainty about it. In turn, the ability to estimate \(S^{(1)}\) overall also tends to improve as \(\phi(X_{j}^{(2)},X_{j}^{\prime(2)})\) approaches \(-1\) or \(1\). This result is commonly understood in the general sense that Predictor Error can negatively affect prediction. This negative effect is most pronounced when the effect of the error moves towards saturation. Counterintuitively, we note that this also explains why high inverted fidelity can be beneficial -- because such stable, albeit inverted, relationships are still highly informative for inferring \(X_{j}^{(2)}\), and subsequently \(S^{(1)}\).

We note that while higher fidelity accelerates the recovery of \(S^{(1)}\), the critical insight of the framework is that this acceleration is not strictly necessary; Breadth (Statement 2a) provides the mechanism to compensate for low or unknown observational fidelity.
\end{quote}

\subsection{Observed Manifestation of Alignment between \(S^{(1)}\) and \(S^{(2)}\) States}

\textbf{Statement 2d:} For a fixed set of marginal distributions for the observed variables in \(S^{\prime(2)}\), the strengthening of \(S^{(1)} \rightarrow S^{(2)}\) linkages across multiple indices \(j\), which increases ``Informative Collinearity'' in \(S^{\prime(2)}\), strictly reduces the joint entropy \(H(S^{\prime(2)})\).

\begin{quote}
\textbf{Argument for Statement 2d:} This relationship can be formalized by decomposing the joint entropy using the concept of Total Correlation (TC), a measure of the total redundancy among a set of variables \citep{watanabe1960information}. The identity is:

\[H(S^{\prime(2)}) = \sum_{j = 1}^{m}{H(X_{j}^{\prime(2)})} - TC(S^{\prime(2)}).\]

\vspace{0.5\baselineskip}
The first term is the sum of the marginal entropies, representing the total uncertainty if all variables were independent. The second term (\(TC(S^{\prime(2)})\)) quantifies the reduction in uncertainty due to the variables' dependencies.

Informative Collinearity is the statistical signature of the shared dependence on a common cause in \(S^{(1)}\). An increase in this collinearity directly leads to an increase in the shared information, or redundancy, among the \(S^{\prime(2)}\) variables, thus increasing the value of \(TC(S^{\prime(2)})\).

Under the condition of fixed marginal distributions, \(\sum_{j = 1}^{m}{H(X_{j}^{\prime(2)})}\) is constant. Thus, any increase in Informative Collinearity results in a larger \(TC(S^{\prime(2)})\) which mathematically necessitates a decrease in \(H(S^{\prime(2)})\).

\textbf{Conclusion for Statement 2d:} Increased Informative Collinearity in \(S^{\prime(2)}\), originating from stronger and more pervasive \(S^{(1)} \rightarrow S^{(2)}\) linkages, leads to a more structured and less random joint distribution of the observed predictors when marginal distributions are constant.
\end{quote}

\subsection{Impact of Predictor Properties on Observed Outcome Inference}

\textbf{Statement 2e:} \(H(Y|S^{\prime(2)})\) is non-increasing with \(m\) and with increased Informative Collinearity in \(S^{\prime(2)}\).

\begin{quote}
\textbf{Argument for Statement 2e:} Incorporating \(X_{(m + 1)}^{\prime(2)}\) into \(S^{\prime(2)}\) forming \(S_{(m + 1)}^{\prime(2)}\) yields \(H(Y|S^{\prime(2)}) \geq \ H(Y|S_{(m + 1)}^{\prime(2)})\). Similarly, changes in the underlying data-generating process that result in increased Informative Collinearity in \(S^{\prime(2)}\) (e.g., stronger \(S^{(1)} \rightarrow S^{(2)}\) linkages, as shown in Statements 2b and 2d) will also lead to a reduction in \(H(Y|S^{\prime(2)})\). This follows because a more accurate estimation of \(S^{(1)}\) states cannot increase, and will generally decrease, the uncertainty about \(Y\).

\textbf{Conclusion for Statement 2e:} Increasing \(S^{\prime(2)}\) dimensionality and increasing Informative Collinearity of \(S^{\prime(2)}\) tends to decrease uncertainty about \(Y\) by enabling a more robust estimation of \(S^{(1)}\).
\end{quote}

\subsection{``High-Quality'' Variables, ``Uniqueness,'' and ``Efficiency''}
\label{``High-Quality'' Variables, ``Uniqueness,'' and ``Efficiency''}

Statement 2a and Appendix~\ref{asymp Breadth v Depth} show that high-D alone enables perfect recovery of \(S^{(1)}\) in the asymptotic limit. Alternatively, Statements 2b and 2c suggest that the rate of improvement in the recovery of \(S^{(1)}\) as \(m\) increases is determined by the interplay of the strengths of the \(S^{(1)} \rightarrow S^{(2)}\) linkages and the observational fidelities.

Conceptually, the rate at which increasing \(m\) reduces uncertainty about \(S^{(1)}\) (the efficiency) is driven by the ``quality'' of the incorporated predictors. A ``high-quality'' predictor provides a large amount of conditionally unique information about \(S^{(1)}\). To formalize how quality contributes to uniqueness and efficiency (see Appendix~\ref{finite convergence}), we define the following concepts in relation to the framework's structure and parameters (\(\gamma\)'s, \(\alpha\)'s, and \(\beta\)'s defined in Section 3), distinguishing between properties of the realizations of uncertainty and properties of the parameters governing the pathways.

Recall that the \(\gamma\) parameters for an indicator \(X_{j}^{(2)}\) form a vector \(\varphi_{S^{(1)} \rightarrow X_{j}^{(2)}}\) containing the conditional probabilities \(P(X_{j}^{(2)}=1 | s_p)\) for each realized configuration \(s_p\).

\begin{itemize}
\item 
\textbf{Distinctness (The Prerequisite for Information Gain):} Distinctness relates to the realizations of uncertainty. An indicator is ``distinct'' if the realizations of its Structural Uncertainty (\(S^{(1)} \rightarrow S^{(2)}\)) and Predictor Error (\(S^{(2)} \rightarrow S^{\prime(2)}\)) are conditionally independent of the corresponding realizations of uncertainty for other indicators. This requirement is guaranteed by the assumptions of Local Independence (Assumption 1.a) and Independent Errors (Assumption 4). Distinctness is the foundational statistical property that allows information to aggregate (via the factorization of probabilities).
\end{itemize}

\begin{raggedright}
Assuming indicators are distinct, the nature of their information gain is determined by the relationship between their \(\gamma\) vectors (\(\varphi\)):
\end{raggedright}

\begin{itemize}
\item
\textbf{Novelty (for Completeness):} Novelty relates to the diversity of the pathways. A candidate predictor is novel if its \(\gamma\) vector (\(\varphi\)) is weakly correlated (or uncorrelated) with the \(\gamma\) vectors of predictors already in the set. Thus, a predictor is novel if it responds to the \(S^{(1)}\) configurations in a fundamentally different way than existing predictors. Adding a novel predictor to a set expands coverage of the latent \(S^{(1)}\) layer.
\item
\textbf{Informative Redundancy (for Reliability):} Redundancy relates to the similarity of the pathways. A new predictor is redundant if its \(\gamma\) vector (\(\varphi\)) is highly correlated -- either positively or negatively -- with the \(\gamma\) vector of an existing predictor. This means the predictors respond similarly (or inversely) to the \(S^{(1)}\) configurations. Crucially, this redundancy is ``informative'' only if the indicator is distinct. Two predictors with perfectly correlated (or anti-correlated) \(\gamma\) vectors still provide unique information because their realizations of uncertainty are independent draws. Informative redundancy reinforces existing signals in the latent \(S^{(1)}\) layer.
\end{itemize}

\begin{raggedright}
The magnitude of the information gain (the overall signal quality) is determined by two factors:
\end{raggedright}

\begin{itemize}
\item
\textbf{Structural Strength:} This relates to the discriminative power of the \(S^{(1)} \rightarrow S^{(2)}\) link, determined by the \(\gamma\) vector. Strength is maximized when the values within the \(\gamma\) vector vary significantly across different \(S^{(1)}\) configurations (ensuring the indicator helps distinguish configurations) and deviate substantially from \(0.5\) (as probabilities near \(0\) or \(1\) are highly informative). Stronger structural links maximize the information flow from the first-stage latent layer (Statement 2b).
\item
\textbf{Observational Fidelity:} This relates strictly to the precision of the \(S^{(2)} \rightarrow S^{\prime(2)}\) process, determined by the \(\alpha\) and \(\beta\) parameters. High fidelity (or high inverted fidelity) minimizes the information loss during observation (Statement 2c).
\end{itemize}

\begin{raggedright}
It is critical to recognize that these item-level properties are the micro-foundations of the global covariance structure. Specifically, Structural Strength contributes to the magnitude of the signal eigenvalues (\(\lambda_1, \dots, \lambda_r\), where \(r\) is the Effective Signal Rank), pushing the ``spike'' of the scree plot upward. Conversely, observational fidelity contributes to the variance of the idiosyncratic noise, helping set the ``floor'' for the noise tail (\(\lambda_{r+1}, \dots, \lambda_m\)). \textbf{Therefore, we can also conceptualize high-quality variables as those that amplify the ``Spectral SNR'' (\(SNR_{spectral} = \lambda_r/\lambda_{r+1}\)), creating the clear eigengap required for the latent signal to be more readily distinguished from the high-D, noisy predictor-space (see Section~\ref{Spectral SNR Definition}).}
\end{raggedright}

In summary, distinctness is a structural prerequisite. Novelty and informative redundancy describe what information the indicator captures relative to the existing set. Structural Strength and observational fidelity determine how clearly that information is captured. In an ideal setting, a high-quality predictor possesses both high Structural Strength and high observational fidelity. However, this framework focuses on scenarios where observational fidelity is inherently low or practically immutable (e.g., when the scale of the data makes manual cleaning unfeasible). In such contexts, we prioritize predictors based on their structural assets: Structural Strength, Novelty, and Informative Redundancy. One of the framework's core mechanisms compensates for low observational fidelity by aggregating these structural assets across a large Breadth of predictors. Aligned with this focus, the below discussion treats observational fidelity as fixed when examining the roles novelty and informative redundancy play.

\subsubsection{Information Gain and Efficiency}

To analyze the recovery of \(S^{(1)}\) more formally, we revisit the reduction in uncertainty from adding a new variable \(X_{(m + 1)}^{\prime(2)}\) to the existing set \(S^{\prime(2)}\), first discussed in Statement 2a. Focusing on the set-level, this equals the conditional mutual information that \(X_{(m + 1)}^{\prime(2)}\) provides:

\[H(S^{(1)}|S^{\prime(2)}) - H(S^{(1)}|S^{\prime(2)} , X_{(m + 1)}^{\prime(2)}) = I(S^{(1)};X_{(m + 1)}^{\prime(2)}|S^{\prime(2)}).\]

{\raggedright
\vspace{0.5\baselineskip}
\(I(S^{(1)};X_{(m + 1)}^{\prime(2)}|S^{\prime(2)})\) is the ``information gain'' we have discussed, which describes the effect a new predictor has on the recovery of \(S^{(1)}\), and also dictates the rate of convergence towards perfect \(S^{(1)}\) recovery (across multiple new predictors).}

This formulation clarifies our definition of ``quality'' as a portfolio-level construct. A variable's quality is not determined solely at the item level, but rather its utility in this context is tied to the information it provides above and beyond the other features in the selected set. This also further informs our understanding of the difference between informative and trivial redundancy. It is the conditioning on the existing predictor set (\(S^{\prime(2)}\)) that defines the concept of ``uniqueness.'' Thus, a new variable is trivially collinear with the existing predictor variables when this conditional mutual information is zero. Conversely, Informative Collinearity is the observable signature of informative redundancy scaled across many variables; it arises when predictors are unconditionally correlated due to a shared cause in \(S^{(1)}\) but also provide unique information due to the conditional independence of their uncertainty realizations. While this concept aligns with ``Redundant'' and ``Synergistic'' information in Partial Information Decomposition Theory \citep{williams2010}, we define it here strictly through the generative parameters of the primary structure.

Analyzing the information flow from \(S^{(1)}\) to \(S^{\prime(2)}\), we can intuitively understand the role of Informative Collinearity. Recall that information flows along the Markov chain \(S^{(1)} \rightarrow S^{(2)} \rightarrow S^{\prime(2)}\), so information can only be lost at each step according to the Data Processing Inequality \citep{cover2006}. Focusing on changes to the \(S^{(1)} \rightarrow S^{(2)}\) linkages, as these linkages weaken, the information gain that \(X_{(m + 1)}^{\prime(2)}\) provides becomes smaller if the error mechanisms are fixed. Alternatively, as these linkages strengthen, the information gain that \(X_{(m + 1)}^{\prime(2)}\) provides becomes larger for fixed error mechanisms. Thus, information gain is positively associated with levels of Informative Collinearity.

Efficiency can be formalized by analyzing the finite-\(m\) convergence rate -- specifically, how quickly the probability of error in estimating \(S^{(1)}\) decreases as the number of predictors \(m\) grows. This analysis provides a concrete link between predictor quality and the speed of perfect \(S^{(1)}\) recovery. As detailed in Appendix~\ref{finite convergence}, we analyze the probability of error in estimating \(S^{(1)}\) from \(m\) observations (\(P_{e}(m)\)) utilizing Chernoff Information (CI) \citep{chernoff1952measure} -- a measure of the statistical distinguishability between the distributions induced by different latent configurations -- to bound this error. The analysis demonstrates that \(P_{e}(m)\) decreases exponentially with \(m\):

\[P_{e}(m) \leq \exp(- m \cdot R).\]

{\raggedright
\vspace{0.5\baselineskip}
Here, \(R\) is the ``Asymptotic Efficiency Rate,'' defined as the average CI between the least distinguishable pair of \(S^{(1)}\) configurations based on the current data (the ``weakest link''). The Asymptotic Identifiability assumption guarantees that \(R > 0\). Furthermore, by Fano's Inequality \citep{fano1961transmission}, the conditional entropy (\(H(S^{(1)}|S^{\prime(2)})\)) is upper bounded by a function of \(P_{e}(m)\). Therefore, the \(H(S^{(1)}|S^{\prime(2)})\) uncertainty also converges exponentially towards zero at a rate governed by \(R\).}

This result helps substantiate Statement 2b and the role of Informative Collinearity as it relates to efficiency -- \(R\) increases when new predictors raise the average CI for the weakest link, and this occurs when new predictors provide either novel information or informatively redundant information related to the weakest link.

This analysis highlights that achieving efficiency requires a dual approach. \textbf{First}, it is critical to acquire variables from diverse predictors to enhance novelty for the sake of complete coverage, ensuring all aspects of \(S^{(1)}\) are represented. \textbf{Second}, particularly in contexts where observational fidelity is inherently low or the vastness of the data renders direct data cleaning impractical, it is essential to leverage informative redundancy (which manifests as Informative Collinearity) to ensure that the covered aspects of the weakest link are reliably represented, effectively compensating for the errors in individual indicators.

We note that information gain always contributes to overall reliability of the \(S^{(1)}\) representation (i.e., enhances confidence we have in distinguishing specific pairs of configurations). However, efficiency (the asymptotic convergence rate \(R\)) is specifically governed by the least distinguishable pair (the weakest link). Thus, a new predictor might significantly improve reliability without increasing efficiency. This fosters a nuanced understanding of Informative Collinearity: it drives reliability of representing \(S^{(1)}\) broadly, and when targeted at the weakest link, enhances asymptotic efficiency.

The concepts of information gain and uniqueness suggest that the principle of diminishing returns also applies to increasing \(m\). Each additional variable contributes by reducing uncertainty based on the unique information it provides about the \(S^{(1)}\) layer. However, as \(m\) grows, it becomes increasingly likely that a new variable will offer less unique information -- this is empirically demonstrated in Section~\ref{Sim}. Therefore, in practice, the benefit of continuing to collect additional variables must be weighed against the point at which the rate of \(S^{(1)}\) recovery becomes marginal.

This principle of diminishing returns is the core justification for strategic data architecture. In a world of infinite model capacity and computational power, one might simply ``use everything.'' But in practice, every predictor added incurs a computational cost and strains the capacity of a finite model to isolate the true signal. The goal of P-DCAI, therefore, is to prioritize the inclusion of predictors that offer the highest information gain (novelty and informative redundancy), ensuring the fastest possible convergence rate and maximizing robustness within a feasible budget. See Section~\ref{Towards Proactive Data-Centric AI} for related discussion.

Lastly, while the Asymptotic Efficiency Rate (\(R\)) governs the perfect recovery of the \(S^{(1)}\) layer, practical prediction of \(Y\) allows for a refined definition: ``Predictive Efficiency.'' In this context, not all configuration errors are equal. The distinguishability requirement for a specific pair of configurations \((s_p, s_\omega)\) should be weighted by the Mutual Information that the distinction between them provides about the \(Y\) (roughly proportional to the divergence between \(P(Y|s_p)\) and \(P(Y|s_\omega)\)). Consequently, the effective weakest link is not the pair of configurations that is hardest to distinguish statistically, but is the pair that is hard to distinguish and carries significant, conflicting information about \(Y\). A supervised feature selection strategy that targets this weighted metric allows for robust prediction of \(Y\) with a smaller \(m\) than is required for the total structural recovery of \(S^{(1)}\). However, this strategy relies on \(Y\) being observed (in practice as \(Y^{\prime}\)) with high observational fidelity (see Section~\ref{Automated Feature Selection}).

\subsection{The Mechanism of Robustness: Triangulating \(S^{(1)}\) with Redundant and Novel Information}

The core theoretical arguments converge on a crucial insight for achieving robust prediction. In the presence of fixed Predictor Error mechanisms, the power of a high-D, informatively collinear predictor-space lies not in its ability to recover \(S^{(2)}\) from \(S^{\prime(2)}\), but instead in its ability to enable robust representation of \(S^{(1)}\). Effectively, the data architecture provides the inherent potential for a model to triangulate \(S^{(1)}\). This process relies on aggregating information from distinct indicators, leveraging two complementary strategies: utilizing novelty (diverse \(\gamma\) parameters) to ensure the resulting picture of \(S^{(1)}\) is complete, and utilizing informative redundancy (similar \(\gamma\) parameters with independent realizations of uncertainty) to ensure the picture is reliable. The efficiency of this triangulation is driven by the indicators' Structural Strength. It is this mechanism that allows for both Predictor Error and Structural Uncertainty to be mitigated in the presence of finite data.

\subsubsection{Triangulating the Truth: The Detective Analogy}
\label{Triangulating the Truth: The Detective Analogy}

To offer an analogy, imagine a detective trying to determine if a subject committed a crime (\(Y\)). During the investigation, no hard physical evidence directly linking the subject to the crime was recovered -- in theoretical terms, the available data do not exhibit a flat topology. Instead, the detective must rely on latent conditions to determine the likelihood of \(Y\): these are the \(S^{(1)}\) states -- for example, the suspect's motive, character, etc.

Focusing on resolving character, the detective undertakes an investigation, gathering multiple measurements of character. Each is distinct, with its own limited vantage point (Structural Uncertainty) and its own potential for error (Predictor Error). These are the \(S^{\prime(2)}\) variables. \textit{Testimonial Evidence from Friends}: She interviews people in the suspect's social circle, knowing testimony only reflects social contexts and could also be colored by personal bias or faulty memory. \textit{Historical Police Evidence}: She pulls official police records, aware that these only reflect acute types of high-stress moments and might also be incomplete or contain clerical errors. \textit{Behavioral Evidence from Professional Settings}: She reviews job performance reports, recognizing that professional conduct only reflects workplace behavior and that such reports may also reflect bias or contain clerical errors.

Each measurement is a qualitatively different type of information -- these \(S^{(2)}\) states and their corresponding observed variables are not merely re-measurements of \(S^{(1)}\) using the same tool. Instead, all are truly distinct indicators, each containing both novel and informatively redundant information about the same underlying \(S^{(1)}\) state of interest. Because no single piece of information fully captures the subject's character (even if perfectly error free), the detective can only form a complete and reliable judgment by integrating the totality of these disparate and flawed pieces of evidence.

Once this synthesis is performed, the character assessment is more robust than if it had been based on any single variable. Once similar investigations are executed to assess the other \(S^{(1)}\) states (e.g., motive), the detective uses the set of assessments to estimate the likelihood that the suspect committed the crime (\(Y\)).

\subsection{Primacy of the \(S^{(1)}\) Layer and the Role of Interactions}
\label{Primacy of the S(1) Layer and the Role of Interactions}

The framework implies that when data align with the primary structure, a modeling approach capable of representing \(S^{(1)}\) will achieve a better prediction of \(Y\) than a simpler model that attempts to predict \(Y\) without first explicitly or implicitly accessing \(S^{(1)}\) (i.e., a flat model).

We prove this assertion by demonstrating that even under ideal observational circumstances -- an infinite number of predictors all with independent errors -- a model that attempts to predict \(Y\) without representing \(S^{(1)}\) cannot reduce the uncertainty about \(Y\) to the level achievable by those that first access \(S^{(1)}\).

{\raggedright
\vspace{0.5\baselineskip}
\textbf{Statement 3: \(H(Y|S^{(1)}) \leq H(Y|S^{\prime(2)})\)}}

\begin{enumerate}
\def\labelenumi{(\alph{enumi})}
\item
In the asymptotic limit (``Infinite Breadth,'' \(m \rightarrow \infty\)), \(H(Y | S^{(1)}) = H(Y | S_{\infty}^{\prime(2)})\), where \(S_{\infty}^{\prime(2)}\) denotes the infinite sequence of observed variables.
\item
For a finite number of predictors \(m\), \(H(Y|S^{(1)}) < H(Y|S^{\prime(2)})\).
\end{enumerate}

\begin{quote}
{\raggedright
\textbf{Argument 1 for Statement 3:}}

\begin{enumerate}
\item
The primary structure \(Y \leftarrow S^{(1)} \rightarrow S^{\prime(2)}\) (a simplification of the full chain) establishes that \(Y\) and \(S^{\prime(2)}\) are conditionally independent given \(S^{(1)}\). This implies \(I(Y;S^{\prime(2)} | S^{(1)}) = 0\).
\item
Using the chain rule for mutual information, we can expand the joint mutual information \(I(Y;(S^{(1)},S^{\prime(2)}))\) in two ways:
\begin{itemize}
\item
\(I(Y;(S^{(1)},S^{\prime(2)})) = I(Y;S^{(1)}) + I(Y;S^{\prime(2)} | S^{(1)})\).

Since the second term is zero (Step 1), this simplifies to \(I(Y;S^{(1)})\).
\item
\(I(Y;(S^{(1)},S^{\prime(2)})) = \ I(Y;S^{\prime(2)}) + I(Y;S^{(1)} | S^{\prime(2)})\).
\end{itemize}
\item
Equating these two expressions gives: \(I(Y;S^{(1)}) = I(Y;S^{\prime(2)}) + I(Y;S^{(1)} | S^{\prime(2)})\).
\item
Since mutual information is non-negative, \(I(Y;S^{(1)} | S^{\prime(2)}) \geq 0\). This means \(I(Y;S^{(1)}) \geq I(Y;S^{\prime(2)})\). This is the Data Processing Inequality. Translating this to entropy yields \(H(Y|S^{(1)}) \leq H(Y|S^{\prime(2)})\). The strictness of the inequality depends on whether \(m\) is finite or infinite.

As shown in Appendix~\ref{asymp Breadth v Depth}, in the asymptotic limit, \(S^{(1)}\) can be perfectly recovered from \(S_{\infty}^{\prime(2)}\). Thus, \(H(S^{(1)} | S_{\infty}^{\prime(2)}) = 0\), which implies \(I(Y;S^{(1)} | S_{\infty}^{\prime(2)}) = 0\). Therefore, \(I(Y;S^{(1)}) = I(Y;S_{\infty}^{\prime(2)})\) when \(m \rightarrow \infty\).

Alternatively, because the pathway \(S^{(1)} \rightarrow S^{\prime(2)}\) is non-deterministic, when \(m\) is finite, \(S^{\prime(2)}\) cannot perfectly determine \(S^{(1)}\) (i.e., \(H(S^{(1)} | S^{\prime(2)}) > 0\)). This implies \(I(Y;S^{(1)} | S^{\prime(2)}) > 0\). Therefore, \(I(Y;S^{(1)}) > I(Y;S^{\prime(2)})\) when \(m\) is finite.
\item
The general relationship between mutual information and conditional entropy is \(I(A;B) = H(A) - H(A|B)\). Substituting this into the results from Step 4 yields:
\begin{itemize}
\item
Asymptotic Limit: \(H(Y) - H(Y | S^{(1)}) = H(Y) - H(Y | S_{\infty}^{\prime(2)})\).
\item
Finite \(m\): \(H(Y) - H(Y | S^{(1)}) > H(Y) - H(Y | S^{\prime(2)})\).
\end{itemize}
\item
By basic algebraic manipulation: \(H(Y | S_{\infty}^{\prime(2)}) = H(Y|S^{(1)})\) when \(m \rightarrow \infty\) and \(H(Y | S^{\prime(2)}) > H(Y|S^{(1)})\) when \(m\) is finite.
\end{enumerate}

This formally establishes \(H(Y|S^{(1)})\), or equivalently \(H(Y | S_{\infty}^{\prime(2)})\), as the theoretical ideal predictive performance level, showing that idealized prediction can be obtained by utilizing infinite data (a justification for high-D Breadth).

However, in practice, in addition to the size of \(m\), this result is also dependent on the model's capabilities to recover the optimal prediction function.

{\raggedright
\vspace{0.5\baselineskip}
\textbf{Argument 2 for Statement 3 -- The Necessity of Interactions:}}

\vspace{0.5\baselineskip}
{\raggedright
The goal is to determine the functional form of the optimal predictor, which utilizes the true conditional distribution \(P(Y | S^{\prime(2)})\). This is derived by marginalizing over \(S^{(1)}\):}

\[P(Y | S^{\prime(2)}) = \sum_{S^{(1)}}{P(Y | S^{(1)})P(S^{(1)} | S^{\prime(2)})}.\]

{\raggedright
\vspace{0.5\baselineskip}
The key component is the posterior inference \(P(S^{(1)} | S^{\prime(2)})\).}

Recall that \(S^{(1)}\) consists of \(k\) binary states, there are \(K_{rlzd}\) realized configurations of \(S^{(1)}\), and \(s_{p}\) denotes the \(p^{\text{th}}\) configuration. We analyze the posterior \(P(s_{p} | S^{\prime(2)})\) using Bayes Theorem:

\[P(s_{p} | S^{\prime(2)}) = \frac{P(S^{\prime(2)}|s_{p})P(s_{p})}{P(S^{\prime(2)})}.\]

{\raggedright
\vspace{0.5\baselineskip}
Utilizing conditional independence, we obtain:}

\[P(S^{\prime(2)} | s_{p}) = \prod_{j = 1}^{m}{P(X_{j}^{\prime(2)}|s_{p})}.\]

{\raggedright
\vspace{0.5\baselineskip}
Examining the log of the joint probability for configuration \(s_{p}\):}

\[LPJ_{p}(S^{\prime(2)}) = \log(P(S^{\prime(2)},s_{p})) = \log(P(s_{p})) + \sum_{j = 1}^{m}{\log(P(X_{j}^{\prime(2)}|s_{p}))}.\]

{\raggedright
\vspace{0.5\baselineskip}
For binary predictors, the term \(\log(P(X_{j}^{\prime(2)}|s_{p}))\) is linear in \(X_{j}^{\prime(2)}\). Therefore, \(LPJ_{p}(S^{\prime(2)})\) is an affine (linear + constant) function of the input vector \(S^{\prime(2)}\) which can be expressed compactly in vector notation as \(LPJ_{p}(S^{\prime(2)}) = b_{p} + \mathbf{w}_{p}^{T}S^{\prime(2)}\), where \(b_{p}\) is the constant/bias term and \(\mathbf{w}_{p}\) is a weight vector specific to configuration \(s_{p}\).}

The posterior probability \(P(s_{p} | S^{\prime(2)})\) is obtained by exponentiating the log of the joint probability and normalizing over realized configurations using the normalizing constant \(P(S^{\prime(2)}) = \sum_{c = 1}^{K_{rlzd}}{\exp(LPJ_{c}(S^{\prime(2)}))}\):

\[P(s_{p} | S^{\prime(2)}) = \frac{\exp(LPJ_{p}(S^{\prime(2)}))}{\sum_{c = 1}^{K_{rlzd}}{\exp(LPJ_{c}(S^{\prime(2)}))}}.\]

{\raggedright
\vspace{0.5\baselineskip}
This is precisely the Softmax function \citep{bishop2006pattern} applied to the vector of linear evidence scores (reducing to the logistic function when \(k = 1\)). Thus, the Softmax function is the mechanism that introduces interactions. While its inputs are linear, the output is generally non-linear. The non-linearity arises because the denominator couples the evidence scores from all configurations. The probability assigned to the configuration \(s_{p}\) depends on the relative evidence \(LPJ_{p}\) compared to all other \(LPJ_{c}\). The effect of an input variable \(X_{j}^{\prime(2)}\) on the posterior probability depends on the magnitude of the denominator, which, in turn, depends on the values of all other input variables. This coupling in the denominator introduces inherent non-linearities and dependencies analogous to high-order interactions, where the predictive weight of one variable is modulated by the state of others.}

Substituting the Softmax formulation and vector notation back into the optimal prediction function, \(P(Y | S^{\prime(2)})\), we obtain:

\[P(Y = 1 | S^{\prime(2)}) = \sum_{p = 1}^{K_{rlzd}}{P(Y = 1 | s_{p})\frac{\exp(b_{p} + \mathbf{w}_{p}^{T}S^{\prime(2)})}{\sum_{c = 1}^{K_{rlzd}}{\exp(b_{c} + \mathbf{w}_{c}^{T}S^{\prime(2)})}}}.\]

{\raggedright
\vspace{0.5\baselineskip}
Thus, the optimal prediction function, being a weighted sum where the weights are the non-linear Softmax outputs, is generally non-linear with respect to \(S^{\prime(2)}\).}

{\raggedright
\vspace{0.5\baselineskip}
\textbf{Conclusion for Statement 3:}}

{\raggedright
\vspace{0.5\baselineskip}
The practical implication of this analysis is that the Infinite Breadth strategy allows for idealized prediction from \(S_{\infty}^{\prime(2)}\) (perfect recovery of \(S^{(1)}\) from \(S_{\infty}^{\prime(2)}\)), thereby overcoming both Predictor Error and Structural Uncertainty. However, to realize this ideal, one must utilize a modeling methodology that has the capacity (leverages interactions) to accurately model the optimal prediction function. Without such capacity, a predictive model cannot fully exploit the information in \(S^{\prime(2)}\) to reconstruct \(S^{(1)}\), and its performance will fall short of the theoretical ceiling, \(H(Y|S^{(1)})\), even in the presence of infinite data.}

This provides a justification for pursuing more complex, hierarchical modeling strategies when the data align with the primary structure. It also explains how architectures from highly flexible modeling methodologies (e.g., neural networks, random forests) can reconstruct \(S^{(1)}\) when predicting \(Y\).

The work of \cite{gu2020} explicitly attributes the superior performance of ML models like neural networks and regression trees to their ability to model ``non-linear predictor interactions missed by other methods.'' Our analysis aligns with this by offering a data-architectural explanation for why interactions are fundamentally important -- because they provide a mechanism to represent different configurations of \(S^{(1)}\), making them implicit proxies for \(S^{(1)}\).

Finally, we acknowledge that this Softmax derivation explicitly relies on the binary variables model. For continuous variables, the conclusion that modeling the latent hierarchy requires interactions depends on the form of the underlying data model and its variables' distributions. Full examination of this result for continuous variables is a clear direction for future theoretical development.
\end{quote}

\section{A Model of Systematic Predictor Errors and Its Impact}
\label{A Model of Systematic Predictor Errors and Its Impact}

The core analysis assumes the Baseline (Independent) Error Regime. Real-world data often contain systematic errors, which can be broadly categorized based on how they affect the independence of error realizations.

\begin{itemize}
\item 
``Systematic Error Realizations'' occur when the error realizations themselves are dependent across predictors, typically due to an unobserved common cause in the measurement process (e.g., a shared instrument artifact or batch effect). This violates the core assumption of conditionally independent errors (Assumption 4), meaning the indicators are no longer distinct. If unaddressed, Breadth strategies may not asymptotically overcome this type of error, as aggregating correlated realized errors converges to the shared bias rather than eliminating it. We discuss potential mechanisms for overcoming this challenge in Section~\ref{Absorption Hypothesis}, where we show why a sufficiently flexible model can theoretically absorb such dependencies as ``pseudo-states.'' This model-centric capability is justified by a formal data-architectural principle (detailed in Appendix~\ref{equivalence for LI viol}). Specifically, that such dependencies are mathematically equivalent to an expanded latent structure where conditional independence is restored.
\item
A ``Systematic Error Regime'' (SERg) occurs when a systemic factor influences the parameters of the error-generating process, but the error realizations remain conditionally independent given the regime. This section introduces a generative model for Systematic Error Regimes based on a latent error-regime-switching variable, \(L\). This tractable framework allows us to derive the boundaries imposed by ambiguity regarding the error-generating process.
\end{itemize}

\subsection{Incorporating Systematic Error Regimes: The Generalized (Mixed) Error Model}

Define a systematic Predictor Error regime as a subject-level condition that determines whether a common observational process governs a set of the second-stage variables and their observations. This can be modeled as follows:

\begin{itemize}
\item
Let \(L \in \{0,1\}\) be a binary, latent, error-regime indicator.

\item
Let \(\pi = P(L = 1)\) be the prior probability that the SERg is active.

Note that \(\pi\) can be interpreted either as the prevalence of the SERg across subjects, or, equivalently, as the unconditional probability that a given subject is governed by the SERg.

\item
Let \(A \subseteq \{1,\ldots,m\}\) denote the (possibly empty) index set of affected variables; indices in \(A^{C}\) are unaffected. Crucially for this model of SERg, we assume the realizations of the errors are conditionally independent across variables given \((S^{(2)},L)\). Therefore, the indicators remain distinct even when they share the same error parameters. Though unconditionally systematic (as marginalizing over \(L\) induces dependence), conditional independence of realizations is maintained.
\end{itemize}

{\raggedright
\textbf{Baseline (independent error) regime, \(L = 0\):} Each observed variable follows its own, potentially asymmetric, observational mechanism as defined in Section~\ref{Framework Model and Notation}. Adding this error-regime notation, the baseline observational process for the \(j^{\text{th}}\) state is defined by \((\alpha_{j},\beta_{j})\).}

\[\alpha_{j} = P(X_{j}^{\prime(2)} = 1 | X_{j}^{(2)} = 1,L = 0),\]
\[\beta_{j} = P(X_{j}^{\prime(2)} = 0 | X_{j}^{(2)} = 0,L = 0)\]

\vspace{0.5\baselineskip}
, so that

\[P(S^{\prime(2)}|S^{(2)},L = 0) = \prod_{j = 1}^{m}{P(X_{j}^{\prime(2)}|X_{j}^{(2)},L = 0)}.\]

{\raggedright
\vspace{0.5\baselineskip}
\textbf{Systematic Error Regime, \(L = 1\):} All affected variables \(j \in A\) share the same parameters that define the observational mechanism \((\alpha_{c},\beta_{c})\):}

\[\alpha_{c} = P(X_{j}^{\prime(2)} = 1 | X_{j}^{(2)} = 1,L = 1),\]
\[\beta_{c} = P(X_{j}^{\prime(2)} = 0 | X_{j}^{(2)} = 0,L = 1).\]

{\raggedright
\vspace{0.5\baselineskip}
Unaffected variables \(j \in A^{C}\) remain under their baseline regime defined by \((\alpha_{j},\beta_{j})\). Hence,}

\[P(S^{\prime(2)}|S^{(2)},L = 1) = (\prod_{j \in A}{P(X_{j}^{\prime(2)} | X_{j}^{(2)})}_{\alpha_{c},\beta_{c}})(\prod_{j \in A^{C}}{P(X_{j}^{\prime(2)} | X_{j}^{(2)})}_{\alpha_{j},\beta_{j}}).\]

{\raggedright
\vspace{0.5\baselineskip}
\textbf{Mixture over the regime:} Marginalizing over \(L\) gives the observed mechanism:}

\[P(S^{\prime(2)} | S^{(2)}) = \pi P(S^{\prime(2)} | S^{(2)},L = 1) + (1 - \pi)P(S^{\prime(2)} | S^{(2)},L = 0).\]

{\raggedright
\vspace{0.5\baselineskip}
This subject-level construction induces cross-variable dependence among \{\(X_{j}^{\prime(2)}\):\(j \in A\}\) after marginalizing over \(L\). When the SERg is active, all affected variables share the same observational mechanism defined by the common \((\alpha_{c},\beta_{c})\). When it is not, each variable follows its own Predictor Error mechanism defined by \((\alpha_{j},\beta_{j})\).}

We note that asymmetry and inverted fidelity are fully allowed in either regime. Additionally, this model can be applied concurrently allowing multiple \(L\) variables to exist simultaneously, aligning with scenarios where different sources of systematic error affect different sets of predictors, as is common in real-world applications.

\subsection{General Decomposition of Uncertainty}

We derive the general expression for the conditional entropy \(H_{L}(m) = H(S^{(1)}|S^{\prime(2)})\), where \(S^{\prime(2)}\) is a set of \(m\) observed variables generated under the generalized (mixed) error model defined in Section~\ref{A Model of Systematic Predictor Errors and Its Impact}. We then analyze \(H_{L}(m)\) in the absence of the SERg to confirm it reduces to the baseline error model assumed in the core analysis. We then discuss the asymptotic limit of Infinite Breadth considering the impact of the number of variables affected by the SERg, although the related derivation is deferred to Appendix~\ref{asymp Breadth v Depth}.

\begin{flushleft}
\textbf{Setup:}
\end{flushleft}

{\raggedright
Partition the \(m\) observed variables in \(S^{\prime(2)}\) into two disjoint sets.}

\begin{itemize}
\item
\(S_{A}^{\prime(2)}\): The subset of affected variables (indices in \(A\)). Let \(m_{A} = |A|\).
\item
\(S_{A^{C}}^{\prime(2)}\): The subset of unaffected variables (indices \(A^{C}\)). Let \(m_{A^{C}} = m - m_{A}\).
\end{itemize}

{\raggedright
We define the conditional entropies under known regimes.}

\begin{itemize}
\item
\(H_{0}(m) = H(S^{(1)}|L = 0,S^{\prime(2)})\): Uncertainty about \(S^{(1)}\) given \(m\) variables under the baseline regime.
\item
\(H_{1}(m) = H(S^{(1)}|L = 1,S^{\prime(2)})\): Uncertainty about \(S^{(1)}\) given \(m\) variables under the SERg.
\end{itemize}

\begin{flushleft}
\textbf{Derivation of \(H_{L}(m)\):}
\end{flushleft}

{\raggedright
Use the information-theoretic identity relating conditional entropy and conditional mutual information, \(H(A | B) = H(A | B,C) + I(A;C|B)\). Set \(A = S^{(1)}\), \(B = S^{\prime(2)}\), and \(C = L\) to obtain:}

\[H_{L}(m) = H(S^{(1)} | S^{\prime(2)}) = H(S^{(1)} | S^{\prime(2)},L) + I(S^{(1)};L|S^{\prime(2)}).\]

{\raggedright
\vspace{0.5\baselineskip}
This equation provides a fundamental decomposition of the total uncertainty. It states that the uncertainty about \(S^{(1)}\) given the observed predictors is equal to the uncertainty assuming the error regime \(L\) is known, plus a penalty term related to the uncertainty of the error regime \(L\).}

\vspace{0.5\baselineskip}
Analyzing the components:

{\raggedright
\vspace{0.5\baselineskip}
Term 1: \(H(S^{(1)} | S^{\prime(2)},L)\) (Expected Entropy Given the Regime)}

\vspace{0.5\baselineskip}
This is the expected conditional entropy of \(S^{(1)}\) when \(L\) is known, averaged over the prior distribution of \(L\).

\[H(S^{(1)} | S^{\prime(2)},L) = \sum_{l \in \{0,1\}}{P(L = l)H(S^{(1)}|L = l,S^{\prime(2)})} = \pi H_{1}(m) + (1 - \pi)H_{0}(m)\]

{\raggedright
\vspace{0.5\baselineskip}
Term 2: \(I(S^{(1)};L|S^{\prime(2)})\) (Cost of Systematic Error Regimes)}

\vspace{0.5\baselineskip}
This term quantifies the dependency between \(S^{(1)}\) and \(L\) after accounting for the information in \(S^{\prime(2)}\). It represents the additional uncertainty about \(S^{(1)}\) introduced because \(L\) is unknown. If needed, it can be expanded as:

\[I(S^{(1)};L | S^{\prime(2)}) = H(L | S^{\prime(2)}) - H(L|S^{(1)},S^{\prime(2)}).\]

\vspace{0.5\baselineskip}
Substituting these terms into the fundamental decomposition, we obtain the general expression for \(H_{L}(m)\):

\[H_{L}(m) = \pi H_{1}(m) + (1 - \pi)H_{0}(m) + H(L | S^{\prime(2)}) - H(L|S^{(1)},S^{\prime(2)}).\]

\vspace{0.5\baselineskip}
If there is no SERg, the error model is fixed. This corresponds to the scenario where the SERg prevalence is zero (i.e., \(\pi = 0\)) and so \(L = 0\). Since \(L\) is determined, entropy about \(L\) is zero. Thus, \(H(L | S^{\prime(2)}) = 0\) and \(H(L | S^{(1)},S^{\prime(2)}) = 0\).

Therefore, when no SERg exists:

\[H_{L}(m) = H_{0}(m).\]

{\raggedright
\vspace{0.5\baselineskip}
This confirms that the general expression reduces to the standard conditional entropy under the baseline (independent) error model.}

\subsection{Asymptotic Analysis (Infinite Breadth Revisited)}

This subsection examines the behavior of the conditional entropy \(H_{L}(m)\) in the theoretical limit of Infinite Breadth (\(m \rightarrow \infty\)). This analysis reveals the fundamental limits imposed by the SERg on the recovery of \(S^{(1)}\).

In addition to the assumptions listed at the beginning of Section~\ref{Core Theoretical Analysis}, the asymptotic analysis assumes \textbf{Distinct Conditional Error Regimes} -- the error mechanisms corresponding to \(L = 0\) and \(L = 1\) are statistically distinct for the affected variables (set \(A\)) when conditioned on \(S^{(1)}\), such that the average KL divergence between the induced distributions over \(A\) is bounded away from zero.

\textbf{Theorem 1 (Asymptotic Convergence of Uncertainty):} As the number of predictors \(m \rightarrow \infty\), the uncertainty about \(S^{(1)}\) given the observed data and the error regime converges to zero. That is, \(H(S^{(1)} | S_{\infty}^{\prime(2)},L) = 0\). Consequently, the total asymptotic conditional entropy is given by:

\[H_{L}(\infty) = \pi(0) + (1 - \pi)(0) + H(L | S_{\infty}^{\prime(2)}) - H(L | S^{(1)},S_{\infty}^{\prime(2)}) = I(S^{(1)};L|S_{\infty}^{\prime(2)}).\]

{\raggedright
\vspace{0.5\baselineskip}
For proof, see Appendix~\ref{asymp Breadth v Depth}.}

Theorem 1 demonstrates that, in the limit of Infinite Breadth, both Predictor Error and Structural Uncertainty are overcome, provided the error regime is known. The residual uncertainty remaining is equal to the information that the error regime provides about \(S^{(1)}\) after observing the infinite data sequence. This result now depends on how the number of affected predictors (\(m_{A}\)) scales with \(m\).

\textbf{Corollary 1.1 (Localized SERg):} If the SERg is localized, meaning \(m_{A}\) remains finite as \(m \rightarrow \infty\), then the asymptotic conditional entropy is zero:

\[H_{L}(\infty) = 0.\]

{\raggedright
\vspace{0.5\baselineskip}
For proof, see Appendix~\ref{asymp Breadth v Depth}.}

In the localized scenario, the number of unaffected variables (\(m_{A^{C}}\)) grows infinitely large. Since these variables follow the baseline regime regardless of \(L\), they provide sufficient information to perfectly recover \(S^{(1)}\), rendering the impact of the finitely many affected variables negligible.

\textbf{Corollary 1.2 (Pervasive SERg):} If the SERg is pervasive, meaning the number of affected predictors \(m_{A} \rightarrow \infty\) as \(m \rightarrow \infty\) (e.g., a fixed fraction of variables are affected, or the universal case where \(m_{A} = m\)), the asymptotic conditional entropy is equal to the uncertainty remaining about the error regime after observing the data:

\[H_{L}(\infty) = H(L | S_{\infty}^{\prime(2)}) = H_{b}(\pi) - I(L;S_{\infty}^{\prime(2)})\]

{\raggedright
\vspace{0.5\baselineskip}
, where \(H_{b}(\pi) = - \pi\log(\pi) - (1 - \pi)\log(1 - \pi)\). For proof, see Appendix~\ref{asymp Breadth v Depth}.}

In the pervasive scenario, the residual uncertainty is determined solely by the distinguishability of the error regimes from the observations alone.

\begin{itemize}
\item
If the regimes are perfectly identifiable, then \(H_{L}(\infty) = H(L | S_{\infty}^{\prime(2)}) = 0\).
\item
If the regimes are indistinguishable (i.e., \(I(L;S_{\infty}^{\prime(2)}) = 0\)), then \(H_{L}(\infty) = H(L | S_{\infty}^{\prime(2)}) = H_{b}(\pi)\).
\item
If the regimes are partially identifiable (i.e., \(I(L;S_{\infty}^{\prime(2)}) > 0\)), then \(H(L | S_{\infty}^{\prime(2)}) < H_{b}(\pi)\).
\end{itemize}

\subsection{Implications: The Power of Infinite Breadth and the Limits of SERg}

The asymptotic analysis under the generalized error model provides crucial insights into the theoretical capabilities and limitations of the Infinite Breadth strategy.

\textbf{1. The Profound Power of Infinite Breadth:} Theorem 1 reinforces a central tenet of this framework: high-D Breadth is a powerful mechanism for mitigating uncertainty. It demonstrates that, asymptotically, the Breadth can overcome not only Predictor Error introduced during observation (\(S^{(2)} \rightarrow S^{\prime(2)}\)) but also the inherent probabilistic nature (Structural Uncertainty) of the relationship between the latent drivers and their true underlying manifestations (\(S^{(1)} \rightarrow S^{(2)}\)). This ability to overcome Structural Uncertainty is the key differentiator from Depth strategies, which can only overcome Predictor Error but remain bounded by Structural Uncertainty (Appendix~\ref{asymp Breadth v Depth}).

\textbf{2. Resilience to Localized SERg:} Corollary 1.1 offers a strong theoretical reassurance for practitioners. In many real-world scenarios, an SERg may stem from specific causes or events affecting only a subset of the predictor-space. The analysis shows that if such regimes are localized (finite \(m_{A}\)), the Infinite Breadth strategy remains asymptotically optimal. The signal from the vast number of predictors with independent errors effectively overwhelms the localized SERg, allowing for perfect recovery of \(S^{(1)}\).

\textbf{3. Fundamental Limits Imposed by Pervasive Systematic Error Regimes:} Corollary 1.2 defines the theoretical boundary when an SERg is widespread. When \(m_{A}\) is infinite, the ability to perfectly recover \(S^{(1)}\) hinges entirely on the ability to identify the active error regime (\(L\)) from the data itself (\(S_{\infty}^{\prime(2)}\)). If the error regimes are statistically indistinguishable, this ambiguity imposes a fundamental limit on performance that cannot be overcome by simply adding more variables.

\textbf{4. The Counterintuitive Role of Prevalence (\(\pi\)) and the Cost of Ambiguity}: The analysis reveals a nuanced relationship between the subject-level prevalence of the SERg (\(\pi\)) and the theoretical performance limit when it is pervasive. The maximum potential penalty, bounded by \(H_{b}(\pi)\), occurs when \(\pi = 0.5\), as this represents the highest ambiguity regarding which error regime is active. Conversely, the penalty is minimized (zero) when the regime is certain (\(\pi = 1\) or \(\pi = 0\)). This leads to the counterintuitive implication that, assuming the regimes are difficult to distinguish from the data, a near-universal SERg (e.g., \(\pi = 0.99\)) imposes a less severe theoretical limit than an intermittent one (\(\pi = 0.5\)). The fundamental limitation of the SERg stems not just from its presence, but from the uncertainty about the error-generating process itself (regime ambiguity).

\textbf{5. Strategic Implications for Data Acquisition and Modeling:} These findings highlight the importance of strategic data acquisition (Section~\ref{Towards Proactive Data-Centric AI}). Acquiring data from diverse measurement sources or using multiple measurement modalities is beneficial primarily because it helps ensure that an SERg remains localized rather than pervasive. This creates a principled rationale for ``Heterogeneity of Measurement.'' Using \(100\) sensors from the same manufacturer may be risky (due to the potential pervasive SERg), whereas using \(100\) sensors from \(10\) different manufacturers is potentially more robust (due to the potential of localized SERg) even if the individual sensors are of lower quality. Furthermore, if a pervasive SERg is suspected, the results emphasize the value of developing models capable of explicitly inferring the active error regime from the data, as identifying the regime is the key to maximizing performance in this challenging scenario. This also suggests that strategies that track measurement sources (assuming sources provide information related to the error regime) and incorporate such metadata into \(S^{\prime(2)}\) have the potential to mitigate the effect of an SERg.

%\textbf{6. The ``Smearing'' of the Spectral Gap:} While formal spectral analysis is detailed in Section~\ref{Benign Overfitting Link: Towards a Unified Understanding of Robustness}, it is worth noting here that Systematic Error Realizations and SERg both have a destructive effect on the data's geometry. By introducing correlations that are neither fully signal (latent \(S^{(1)}\) drivers) nor fully independent errors, systematic errors introduce eigenvalues that populate the space between the signal spike and the noise floor. This ``smears'' the spectral gap, reducing the \(SNR_{spectral}\) and making it increasingly difficult for any model to distinguish the true latent structure from the systematic artifacts of the measurement process.

\section{A Statistical Treatment of Inferential Power and Breadth: The Interplay of Sample Size (\(N\)) and Predictor Set Dimensionality (\(m\))}
\label{Interplay of N and m}

This section directly addresses the statistical feasibility of the high-D Breadth strategy, resolving its apparent conflict with the Curse of Dimensionality. We discuss the non-interchangeable roles played by predictor dimensionality and sample size. Specifically, increasing \(m\) serves to resolve predictor ambiguity by making latent \(S^{(1)}\) configurations observationally distinguishable. In contrast, a sufficient \(N\) is required to resolve outcome ambiguity by providing the statistical power to model the \(S^{(1)}\rightarrow Y\) link. We discuss why, under this architecture, increasing \(m\) does not lead to unbounded data sparsity but instead creates a ``Prevalence Floor,'' fundamentally reducing the statistical burden of latent inference and making the Breadth strategy viable with finite samples.

\subsection{Inferring \(S^{(1)}\): The Core Trade-Off}

At the individual sample level, the goal is the robust inference of the configuration of the \(S^{(1)}\) layer from an observed configuration of the \(S^{\prime(2)}\) layer. Let \(s_1\) and \(s_2\) be two unique configurations of \(S^{(1)}\), and \(g_a\) be a configuration of \(S^{(2)}\). By performing a thought experiment where \(s_1\), \(s_2\), and \(g_a\) are known, we can understand that the ability to make this inference with confidence is a function of a strategic trade-off between two primary resources: sample size (\(N\)) and predictor-space dimensionality (\(m\)) or Breadth.

The primary obstacle to \(S^{(1)}\) inference is ``configuration-level degeneracy:'' the case where a lack of Breadth (low \(m\)) causes distinct \(S^{(1)}\) configurations to become observationally indistinguishable, collapsing into a single \(S^{(2)}\) configuration (e.g., \(g_a\) can originate from either \(s_1\) or \(s_2\)). This degeneracy creates a fundamental choice for the researcher regarding the inference of \(S^{(1)}\), revealing an inverse relationship between \(N\) and \(m\):

\begin{enumerate}
\item
\textbf{Sample Size (\(N\)):} For a given level of degeneracy caused by a fixed \(m\), the challenge centers on statistical estimation. We must precisely quantify the ambiguity by sampling a degenerate \(S^{(2)}\) configuration (e.g., \(g_a\)) with sufficient \(N\) to confidently estimate the conditional probability distributions of interest (e.g., \(P(s_1 | g_a)\) and \(P(s_2 | g_a)\)).
\item
\textbf{Breadth Dimensionality (\(m\)):} For a fixed \(N\), the challenge centers on resolving the degeneracy by design. By increasing \(m\), the mapping from \(S^{(1)}\) configurations to \(S^{(2)}\) configurations approaches one-to-one. This resolution of degeneracy is a direct consequence of the framework's core assumptions: The Asymptotic Identifiability assumption guarantees that for any two distinct \(S^{(1)}\) configurations (e.g., \(s_a\), \(s_b\)), an indicator \(X_j^{(2)}\) exists that can distinguish them. The Local Independence assumption then provides the mathematical factorization needed for this new, distinct information to break the prior ambiguity. The task thus shifts from statistical estimation to simple deterministic confirmation, dramatically reducing the \(N\) required.
\end{enumerate}

\begin{raggedright}
For the specific goal of resolving \(S^{(1)}\) configuration-level degeneracy, these two strategies act as complementary levers, where Breadth reduces the sample complexity required for latent inference. Resources spent on increasing \(m\) reduce the \(N\) required for robust inference of \(S^{(1)}\), and vice versa. However, as analyzed in Section~\ref{final bottleneck}, this interchangeability does not hold for the ultimate goal of predicting \(Y\).
\end{raggedright}

\subsection{Opposing Forces of Sample Cost: \(n_{target}\) vs. \(P(g_{rarest})\)}

To understand this trade-off, we must calculate the total sample size required to fully characterize the \(S^{(1)}\) system, \(N_{upper}\). This is driven by the rarest \(S^{(2)}\) configuration: we need to collect \(n_{target}\) examples of the rarest \(S^{(2)}\) configuration, \(g_{rarest}\). Then, 

\[N_{upper} \approx \frac{n_{target}}{P(g_{rarest})}.\]

\vspace{0.5\baselineskip}
\begin{raggedright}
This ratio, however, creates a conflict: as we increase \(m\), the numerator and denominator have opposing effects on \(N_{upper}\).
\end{raggedright}

\begin{enumerate}
\item 
\textbf{The Numerator (\(n_{target}\)): The Statistical Cost of Ambiguity}

\(n_{target}\) is the number of samples needed to resolve the ambiguity within a single \(S^{(2)}\) configuration (e.g., \(g_a\)). The statistical problem is that when we observe \(g_a\), we do not know the configuration of \(S^{(1)}\) from which it originated (e.g., \(s_1\) or \(s_2\)).

\(n_{target}\) represents the sample size required to confidently estimate the relevant conditional probabilities (e.g., \(P(s_1 | g_a)\) and \(P(s_2 | g_a)\)). In the simplest case of a \(2\)-way split (\(w = 2\)), the worst-case ambiguity is a \(50/50\) split (\(p = 0.5\)). To estimate this proportion with a desired margin of error (\(ME\)) and confidence level (\(z\)-score, \(z\)), the standard sample size formula is:

\[n_{target} = (\frac{z}{ME})^2 \times p(1 - p).\]

\vspace{0.5\baselineskip}
For example, for a \(95\%\) confidence interval (\(z = 1.96\)) and a \(\pm2.5\%\) margin of error (\(ME = 0.025\)), the worst-case (\(p = 0.5\)) cost is \(\approx 1,537\). If the degeneracy is worse than a \(2\)-way split (e.g., \(w > 2\)), the problem becomes multinomial, and the required \(n_{target}\) to achieve the same precision is higher.

As we discuss in Section~\ref{Resolving Conflict: Prev Floor}, as \(m\) increases, Breadth resolves this degeneracy. The ambiguity vanishes (\(w \rightarrow 1\)), and the statistical estimation problem collapses. The cost drops to a deterministic confirmation cost of \(n_{target} = 1\).

\item
\textbf{The Denominator (\(P(g_{rarest})\)): The Prevalence of the Rarest Configuration}

The denominator \(P(g_{rarest})\) is the prevalence of the rarest \(S^{(2)}\) configuration. As \(m\) increases, the total number of possible \(S^{(2)}\) configurations (\(2^m\)) explodes. By the Curse of Dimensionality, it seems the prevalence of any specific \(S^{(2)}\) configuration should decrease toward zero.
\end{enumerate}

\begin{raggedright}
This seemingly creates the conflict: if increasing \(m\) leads to \(n_{target} \rightarrow 1\) and \(P(g_{rarest}) \rightarrow 0\), \(N_{upper}\) would go to infinity (\(\frac{1}{0}\)), and this contradicts our earlier claim that increasing \(m\) reduces the required \(N\).
\end{raggedright}

\subsection{Resolving the Conflict: Breadth and the Prevalence Floor}
\label{Resolving Conflict: Prev Floor}

The conflict is resolved by formally examining the denominator, \(P(g_{rarest})\), or \(P(g_a)\) more generally. The prevalence of \(g_a\) is not an independent value; it is the sum of its probabilities occurring from all possible \(S^{(1)}\) configurations. Recall \(s_p\) is the \(p^\text{th}\) configuration of the \(k\) \(S^{(1)}\) states. Then,

\[P(g_a) = \sum_{p = 1}^{K_{rlzd}}P(g_a | s_p)P(s_p).\]

\vspace{0.5\baselineskip}
This formula reveals how Breadth (increasing \(m\)) fundamentally alters \(P(g_a)\):

\begin{itemize}
\item \textbf{In a Low-\(m\) (Degenerate) System:} \(g_a\) is an ambiguous ``bucket.'' It is degenerate because it contains probability mass from multiple latent sources (e.g., \(P(g_a | s_1) > 0\) and \(P(g_a | s_2) > 0\)). Consequently, \(P(g_a)\) is the sum of these contributions and is relatively high, but the inverse inference remains uncertain.

\item \textbf{In a High-\(m\) (Resolved) System (Elimination of Degeneracy):} By adding distinct indicators, Breadth fundamentally eliminates degeneracy. As \(m \rightarrow \infty\), the distributions of predictor vectors generated by distinct latent configurations become asymptotically orthogonal. Consequently, the predictor-space partitions into distinct ``equivalence classes,'' denoted \(\mathcal{G}_{a}\). Degeneracy is eliminated because these classes become ``pure'': any observed string falling within \(\mathcal{G}_{a}\) is generated exclusively by \(s_a\) (i.e., \(P(\mathcal{G}_{a} | s_{b}) \rightarrow 0\) for all \(b \neq a\)).

While the probability of observing any single specific string \(g_a\) decays to zero (due to the Curse of Dimensionality), the total probability mass of the entire equivalence class \(\mathcal{G}_{a}\) converges to the prevalence of the latent configuration itself (see Appendix~\ref{proof:degeneracy} for the formal proof). We do not need to observe a specific, pre-defined string \(g_a\) to identify \(s_a\); we only need to observe \textit{any} string belonging to the equivalence class \(\mathcal{G}_{a}\). This shifts the denominator in the sample size calculation from the vanishing probability of a specific string to the fixed prevalence of the latent state:

\[\lim_{m\to\infty} P(\mathcal{G}_a) = P(s_a)\]

Thus, by resolving degeneracy, Breadth ensures the Prevalence Floor is set not by the rarity of the specific predictor pattern, but by the rarity of the underlying latent driver. Consequently, the prevalence of the rarest equivalence class (\(P(\mathcal{G}_{rarest})\)) is limited by the rarest \(S^{(1)}\) configuration prevalence (\(P(s_{rarest})\)).

To illustrate this trade-off, consider the analogy of forensic fingerprinting. In a low-\(m\) (low-D Breadth) scenario, the researcher possesses only a ``partial print'' -- a smudge containing just a few ridges. These data are degenerate; the partial pattern is not unique and could plausibly belong to many different individuals (latent configurations). Consequently, the researcher faces a statistical burden: they must collect a massive database (a large \(N\)) to accurately estimate the probability that this ambiguous smudge belongs to a specific suspect. In a high-\(m\) (high-D Breadth) scenario, the researcher possesses a ``full, high-resolution print.'' By increasing the surface area (adding dimensions), the number of distinct features (minutiae) increases to the point where the pattern becomes statistically unique and degeneracy collapses. The researcher no longer needs to estimate the probability of a match across a vast population; they simply need to confirm the identity. Thus, as \(m\) increases, Breadth transforms the problem from one of statistical estimation of the latent configuration (requiring high \(N\)) to one of deterministic identification.
\end{itemize}

In summary, Breadth mitigates degeneracy in two manners:

(a) \textbf{Partial Resolution:} Ambiguity is likely no longer at its maximum. Increasing \(m\) will tend to split the worst-case (e.g., \(50/50\)) ambiguity into a clearer signal (e.g., \(80/20\)), thereby reducing the sample size cost to characterize this lower-ambiguity configuration. For example, an \(80/20\) split results in a lower sample requirement: \(n_{target}(\text{for} \: 95\% \: CI, \: \pm2.5\% \: \text{error}) = \left(\frac{1.96}{0.025}\right)^{2} \times 0.8(1 - 0.8) \approx 984.\)

(b) \textbf{Complete Resolution:} As described above, for sufficiently high \(m\), degeneracy is completely resolved. When this occurs, any string belonging to the equivalence class \(\mathcal{G}_{a}\) becomes a unique, high-resolution signature for one specific \(s_p\).

This mechanism provides a tangible geometric interpretation of the theoretical Infinite Breadth limit, which in practice is approximated by finite high-D systems. While the information-theoretic analysis proves that uncertainty vanishes (\(H(S^{(1)}|S^{\prime(2)}) \to 0\)), this statistical perspective explains how: by expanding dimensionality, the system physically separates the overlapping probability distributions of the latent states until they become orthogonal. Thus, the power of Breadth is not merely the accumulation of redundant signal, but the structural disintegration of ambiguity, effectively transforming a probabilistic \(S^{(1)}\) estimation problem (requiring large \(N\)) into a deterministic identification problem (requiring minimal \(N\)).

\subsection{The Resolved Trade-Off: Comparing Sample Costs}

This Prevalence Floor resolves the conflict and clarifies the statistical impact of Breadth. We can now compare the nature of the costs for the two approaches:

\begin{itemize}
\item
\textbf{Low-\(m\) (Sample Size Strategy):}

\[N_{req} \approx \frac{n_{target}(w \ge 2)}{P(g_{a, \text{degen}})}\]

\vspace{0.5\baselineskip}
This approach is dominated by the statistical cost of ambiguity. The numerator \(n_{target}\) is large (e.g., \(\ge 1,537\)), requiring substantial sampling to resolve ambiguity within the degenerate configuration.

\item
\textbf{High-\(m\) (Breadth Strategy):}

\[N_{req} \approx \frac{n_{target}(w = 1)}{\lim_{m \rightarrow \infty} P(g_a)} = \frac{1}{P(s_{rarest})}\]

\vspace{0.5\baselineskip}
This approach eliminates the statistical cost of ambiguity, as the numerator \(n_{target}\) collapses (e.g., from \(\ge 1,537\) to \(1\)). The required sample size is now governed solely by the prevalence of the rarest latent configuration.
\end{itemize}

The central mechanism is the qualitative shift driven by the dramatic drop in the numerator (from a large statistical cost to a deterministic cost of 1). While the total required sample size \(N_{req}\) may increase or decrease depending on the specific relationship between \(P(g_{a, \text{degen}})\) and \(P(s_{rarest})\), increasing \(m\) fundamentally removes the statistical burden of ambiguity associated with inferring \(S^{(1)}\).

\subsection{A Heuristic Using \(K_{eff}\)}

For rapid feasibility assessment in a low-\(m\) scenario, we can leverage the effective number of configurations, \(K_{eff}\) (as defined in Section~\ref{Benign Overfitting Link: Towards a Unified Understanding of Robustness} and Appendix~\ref{latent sparsity}). The resulting ballpark sample size, using our baseline \(n_{target} \approx 1,537\), is:

\[N_{ballpark} \approx 1,537 \times K_{eff}.\]

\vspace{0.5\baselineskip}
This heuristic is labeled a liberal estimate (an underestimate) because it combines two opposing assumptions:

\begin{enumerate}
\item
\textbf{A Conservative Numerator:} The \(n_{target} \approx 1,537\) term assumes worst-case \(50/50\) ambiguity (\(w = 2\)). As \(m\) increases, any remaining ambiguity is likely less severe (e.g., an \(80/20\) split, which requires a lower \(n_{target}\)). This component of the formula is a conservative overestimate of the cost.
\item
\textbf{A Liberal Multiplier:} The \(K_{eff}\) multiplier is a proxy for \(1/P_{avg}\). The true bottleneck multiplier is \(1/P(s_{rarest})\). By definition, \(P(s_{rarest}) \ll P_{avg}\), which means the true multiplier is orders of magnitude larger than \(K_{eff}\). This component is potentially an extremely optimistic underestimate.
\end{enumerate}

The \(N_{ballpark}\) formula is considered liberal overall because the underestimate from the \(K_{eff}\) multiplier (an exponential-scale error) is assumed to be far more significant than the overestimate from the \(n_{target}\) term (a linear-scale error). It is best used for order of magnitude assessments.

Crucially, this \(K_{eff}\) multiplier is not arbitrary; it corresponds directly to the Effective Signal Rank of the signal component in the covariance matrix. Thus, \(K_{eff}\) dictates the index at which the ``Spectral SNR'' cliff occurs (i.e., the transition from signal \(\lambda_r\) to noise \(\lambda_{r+1}\), see Section~\ref{Spectral SNR Definition}), establishing the dimensionality of the subspace the model must resolve.

\subsection{The Impact of Predictor Error}

All preceding calculations operate in the context of the underlying \(S^{(1)} \rightarrow S^{(2)}\) processes, ignoring Predictor Error. In reality, such error acts as a powerful confounding factor, actively counteracting the benefit of Breadth. It blurs the distinctions gained from a high \(m\), causing distinct \(S^{(2)}\) configurations to be misread as one another. This artificially increases degeneracy, pushes observed probabilities back towards ambiguity, re-inflates the required \(n_{target}\), obscures the Prevalence Floor, and works against the conditional probabilities of interest from reaching their ideal limits (e.g., \(P(g_a | s_a) < 1\) and \(P(g_a | s_b) > 0\)).

Therefore, any sample size calculated from this framework -- even the \(N_{ballpark}\) -- should be considered a liberal minimum when low observational fidelity is expected. If perfect recovery of \(S^{(1)}\) configurations is paramount, these sample size estimates must be padded significantly to provide the statistical power needed to overcome not only the system's inherent degeneracy but also the ambiguity introduced by Predictor Error.

\subsection{The Final Bottleneck: Statistical Cost of Predicting \(Y\)}
\label{final bottleneck}

The preceding analysis focused on the high cost of the first inferential challenge: recovering \(S^{(1)}\) configurations from \(S^{\prime(2)}\). We showed that increasing Breadth is the key to solving this, causing the statistical cost (\(n_{target}\)) for this task to collapse (e.g., from \(\ge 1,537\) to \(1\)). However, this does not mean the total sample requirement for the ultimate goal of predicting \(Y\) also collapses.

Solving the \(S^{\prime(2)} \rightarrow S^{(1)}\) ambiguity simply reveals the second and final statistical bottleneck: the \(S^{(1)} \rightarrow Y\) link. To build a robust predictive model, we must now statistically estimate the conditional probability \(P(Y | s_p)\) for every \(s_p\).

Since \(Y\) is a binary outcome in this treatment, this is a new set of binomial estimation problems. The ``weakest link'' for this specific task is the rarest \(S^{(1)}\) configuration, \(s_{rarest}\). The worst-case ambiguity for this new problem occurs if \(P(Y = 1 | s_{rarest}) = 0.5\). Therefore, the statistical cost discussed above re-emerges. Here though, we must define a new target, \(n_{target, Y}\), which is the sample size needed to reliably estimate this \(S^{(1)} \rightarrow Y\) probability with \(95\%\) confidence and a \(\pm 2.5\%\) margin of error: \(n_{target, Y} \approx 1,537\).

Thus, the statistical cost does not disappear with Breadth; it transfers from the \(S^{\prime(2)} \rightarrow S^{(1)}\) problem to the \(S^{(1)} \rightarrow Y\) problem, where it is more manageable (strictly scaling with \(K_{rlzd}\) rather than \(m\)). Using a \(\pm 2.5\%\) margin of error, the total sample size required for robust prediction of \(Y\), assuming \(m\) is high enough to solve predictor degeneracy, is therefore:

\[N_{Y, \text{req}} \approx \frac{n_{target, Y}}{P(s_{rarest})} \approx \frac{1,537}{P(s_{rarest})}\]

This reveals the complete, non-interchangeable roles of Breadth and sample size: a high \(m\) is required to solve predictor ambiguity, while a high \(N\) is required to solve outcome ambiguity. Both are necessary to achieve robust end-to-end prediction.

\subsection{Benefits of Causal Consistency: Efficiency and Interpolation}
\label{Benefits of Causal Consistency}
The preceding analysis implicitly focuses on the most difficult version of the inference problem: resolving configuration-level degeneracy in the absence of underlying structural relationships (i.e., inference in the context of a Chaotic Generative Variables). However, variables often exhibit monotonic Causal Consistency, and this shifts the dynamics fundamentally, where the requirement to isolate a unique string \(g_a\) is relaxed.

Specifically, when the latent \(S^{(1)}\) states exert stable directional forces within variables, predictors are not merely unique identifiers for specific configurations, but rather function as distinct bits of evidence for the \(S^{(1)}\) states directly. We no longer face a brute-force ``barcode'' matching problem -- which scales strictly with \(K_{rlzd}\) (or effectively with \(K_{eff}\)) -- but rather a standard latent ``signal recovery'' problem that scales only with \(k\) (when Causal Consistency is near-linear). Thus, Causal Consistency increases the efficiency with which increasing \(m\) resolves degeneracy: the effective Breadth (\(m\)) required to resolve ambiguity in the presence of Causally Consistent Variables is significantly lower than that required in the presence of worst-case Chaotic Generative Variables.

Causal Consistency also offers a second benefit regarding the sample size (\(N\)). As established above, the effective complexity of the latent space (\(K_{eff}\)) imposes a theoretical baseline for \(N\), representing the necessity to observe the latent landscape with sufficient granularity for recovery of \(Y\). However, if \(Y\) is also Causally Consistent (a reasonable assumption in many applications), the parametric nature of the \(Y\) generative process introduces the potential for interpolation, which relaxes sample size requirements. If \(Y\) is a Chaotic Generative Variable, the learner is forced to ``memorize'' the signature of every distinct \(S^{(1)}\) configuration independently. Here, a rare configuration absent from the training data (\(N_{train} = 0\)) is fundamentally unpredictable, as its signature shares no logic with observed configurations. Conversely, if \(Y\) is Causally Consistent, the \(S^{(1)}\) latent drivers exert stable forces on \(Y\). This stability allows for \textbf{parameter sharing}: a model can learn the directional influence of a specific latent state from common configurations and successfully apply that learned ``coefficient'' to predict rare or unseen configurations where the \(S^{(1)}\) state is active. This property enables models to generalize to the ``tails'' of the latent distribution when predicting \(Y\). By learning the \textit{physics of the latent components} rather than memorizing the configurations, the model effectively interpolates the structure of rare configurations, partially relaxing the strict sample size constraints imposed by configuration-based latent complexity (like those discussed above).

\section{Benign Overfitting Link: Towards a Unified Understanding of Robustness}
\label{Benign Overfitting Link: Towards a Unified Understanding of Robustness}

An important implication of this framework is its potential to serve as a unifying, data-architectural explanation for robustness to both Outcome Error and predictor-space noise. While our framework's primary mechanism addresses Predictor Error and Structural Uncertainty, it also provides a plausible causal origin for the conditions that enable robustness to Outcome Error via the BO mechanism.

Theories of BO have established that overparameterized models can generalize well despite fitting noisy outcomes if the predictor covariance matrix has ``favorable'' spectral properties, such as a low-rank-plus-diagonal structure \citep{bartlett2020, hastie2022, tsigler2023}. The connection between latent factor models and these favorable covariance structures has been productively explored in recent risk analysis literature, which has analyzed the performance of interpolating predictors specifically in such settings \citep{bing2021, bunea2022}.

Our contribution is distinct and complementary: we propose that our specific data-generating architecture -- which we have already argued is robust to Predictor Error and Structural Uncertainty -- is also a plausible causal origin for these very factor structures, thereby laying the initial groundwork for the unification of distinct theories of robustness under a single architectural principle. This connection is further strengthened by the Manifold Hypothesis, the foundational idea that high-D data are often concentrated on or near a low-dimensional manifold-space. While it is hypothesized that such structures arise from latent hierarchical processes, our framework posits a specific hierarchical model that can generate such a manifold, whose covariance structure, in turn, enables BO.

We formalize this connection by analyzing the spectral properties of the covariance matrix of the observed binary predictors (\(\Sigma_{S^{\prime(2)}}\)) generated by the primary structure. The covariance structure arises from the shared dependence of the \(m\) observed variables (\(S^{\prime(2)}\)) on the \(S^{(1)}\) layer. We can analyze the structure of \(\Sigma_{S^{\prime(2)}}\) using the Law of Total Covariance, which allows us to decompose the total covariance of \(S^{\prime(2)}\) conditioned on \(S^{(1)}\):

\[\Sigma_{S^{\prime(2)}} = Cov(E[ S^{\prime(2)}|S^{(1)} ]) + E[ Cov(S^{\prime(2)} | S^{(1)}) ].\]

{\raggedright
\vspace{0.5\baselineskip}
This decomposition naturally partitions the covariance matrix into two components: a signal component and a noise component.}

\subsection{The Signal Component: \(C_{signal} = Cov(E[ S^{\prime(2)}|S^{(1)} ])\)}
\label{The Signal Component}
This term represents the covariance among the observed variables that is driven by their shared dependence on \(S^{(1)}\) -- it captures the signal propagated from the latent layer. The crucial insight lies in the rank of this component. \(S^{(1)}\) consists of \(k\) binary variables, with \(K_{rlzd}\) realized configurations. Consequently, the vector of conditional expectations (\(E[ S^{\prime(2)}|S^{(1)} ]\)) takes \(K_{rlzd}\) unique values. The rank of a covariance matrix is strictly bounded by the number of unique values the underlying random vector can take (minus one). Therefore, the Algebraic Rank of \(C_{signal}\) is \(K_{rlzd} - 1\).

This strict rank definition notwithstanding, Effective Latent Complexity (\(K_{eff}\)) is most relevant to the spectral properties related to BO, and is typically lower than \(K_{rlzd} - 1\). Low \(K_{eff}\) relative to \(K_{rlzd}\) -- sparsity in the latent configuration space -- arises from two complementary forces. First, the prevalence rate of each individual \(S^{(1)}\) state provides a foundational level of structure. A state that is either extremely rare or extremely common has very little variance and low individual entropy, making any configuration that requires it to be in its less likely state inherently improbable. This ensures that the probability mass of the latent configuration space is unevenly distributed.

Second, as the framework permits, correlations among the \(S^{(1)}\) states induce further sparsity. As detailed in Appendix~\ref{latent sparsity}, correlation among these states reduces the joint entropy of the \(S^{(1)}\) space, exponentially decreasing \(K_{eff}\). This reflects the reality that the underlying states in complex systems are rarely independent. This position is supported by the foundational field of Factor Analysis, where it is well-established that forcing latent factors to be orthogonal often distorts the representation of the underlying structure, and that allowing factors to correlate (e.g., via oblique rotation) provides a more realistic and stable representation of data \citep{Thurstone1947, Browne2001}.

Together, these two mechanisms -- skewed individual prevalences and inter-state correlations -- push \(K_{eff}\) lower, potentially allowing for the low-rank-plus-diagonal link to BO to persist even if the \(k\) is moderately large.

\subsection{The Noise Component: \(C_{noise} = E[ Cov(S^{\prime(2)} | S^{(1)}) ]\)}
\label{The Noise Component}
This term represents the expected variance of the observed variables that remains even when the latent states \(S^{(1)}\) are known. This \(C_{noise}\) component is the combined result of two distinct sources of uncertainty defined by our framework: idiosyncratic variance arising from both Structural Uncertainty and Predictor Error. The core assumptions of LI and the baseline independent error regime imply that the realizations of noise from both are conditionally independent across predictors given \(S^{(1)}\). Therefore, both sources of noise contribute only to the diagonal structure of the \(m\) by \(m\) \(C_{noise}\) matrix, which represents the idiosyncratic noise variance for each predictor.

An implication of this decomposition is that the favorable low-rank-plus-diagonal structure is not merely an artifact of dirty data. Instead, it is a fundamental characteristic of data generated by the probabilistic, hierarchical latent structure. Even in an idealized scenario devoid of Predictor Error, the \(C_{noise}\) term would still be non-zero, generated entirely by the system's inherent Structural Uncertainty. This further strengthens our proposition that the BO phenomenon is an innate feature of data that aligns with the hierarchical nature of the underlying structure, not just a consequence of imperfect measurement.

Additionally, this clarifies the meaning of the Spectral Heuristic discussed in Section~\ref{Assessing Framework Applicability}. The Effective Signal Rank estimated from a scree plot (\(r\)) provides an estimate of the dimensionality of the dominant signal subspace. When the system is comprised of Chaotic Generative Variables, \(r\) strictly scales with \(K_{rlzd}\), and so is distinct from \(K_{eff}\), which is a lower value that discounts rare configurations. Therefore, assuming sufficient \(N\) and \(m\) to distinguish signal from noise, \(r\) should be interpreted as a practical upper limit of \(K_{eff}\). When the system is comprised of Causally Consistent Variables, \(r \rightarrow k\) as consistency becomes increasingly linear.

\subsection{The Spectral SNR (The Eigengap)}
\label{Spectral SNR Definition}

While a global Signal-to-Noise Ratio can be defined by the ratio of total signal and total noise variances, the robustness of the primary structure is more precisely governed by the \textit{transition} between the signal subspace and the noise floor. Accordingly, we define the ``Spectral SNR'' to quantify the severity of the elbow in the eigenvalue spectrum.

Let \(\lambda_1 \ge \dots \ge \lambda_m\) be the eigenvalues of \(\Sigma_{S^{\prime(2)}}\). The Spectral SNR is the ratio of the weakest signal eigenvalue to the strongest noise eigenvalue:

\[SNR_{spectral} = \frac{\lambda_r}{\lambda_{r+1}}.\]

\vspace{0.5\baselineskip}
\begin{raggedright}
This ratio corresponds to the magnitude of the ``spectral gap.'' As \(SNR_{spectral}\) values increase, the transition from noise to signal becomes sharper, implying that \(S^{(1)}\) is increasingly differentiated from the idiosyncratic noise \(C_{noise}\).
\end{raggedright}

While this spectral SNR governs the \textit{recoverability} of the latent signal (determining if the model \textit{can} find \(S^{(1)}\)), the theory of BO posits that the robustness to the remaining noise is governed by the properties of the tail eigenvalues (\(\lambda_{r+1}, \dots, \lambda_m\)). Specifically, the noise must be distributed such that its rank is sufficiently high to dissipate the error of fitting it. Notably, our framework explains how high-D Breadth satisfies both conditions: Informative Collinearity maximizes the spectral SNR (lifting the signal), while the aggregation of distinct predictors ensures a high-D uncorrelated noise floor (expanding the tail's rank).

\subsection{The Favorable Covariance Structure}
The total covariance matrix is the sum of the two components: \(\Sigma_{S^{\prime(2)}} = C_{signal} + C_{noise}\). This is precisely the ``low-rank-plus-diagonal'' structure identified by \cite{bartlett2020} and subsequent works \citep{hastie2022, tsigler2023} as a key prerequisite for BO.

The derived structure connects this work to foundational latent variable modeling fields specialized for analyzing latent structures in binary data, such as IRT or Latent Structure Analysis \citep{goodman1974}. These foundational latent variable theories naturally give rise to the required spectral properties. The low-rank signal component \(C_{signal}\) contributes a relatively small number of potentially large eigenvalues, corresponding to the variance driven by \(S^{(1)}\). The diagonal noise component \(C_{noise}\) contributes variance across all \(m\) dimensions, creating a long tail of smaller eigenvalues.

BO analyses hinge on this interplay: a distinct eigengap (high Spectral SNR) enables the separation of the signal subspace, while the specific decay profile of the noise tail (often quantified by its rank) allows the model to absorb the idiosyncratic variance benignly. Accordingly, we frame our result in terms of these two architectural features: the latent structure concentrates variance in a low-D subspace (maximizing the gap) while keeping the residual spectrum flat and high-D (optimizing the tail), which exactly matches the spectral conditions used in BO proofs.

This connection highlights a critical complementarity: while BO theory explains how this covariance structure enables robustness to Outcome Error, our framework explains how the same architecture leverages high-D Breadth to mitigate predictor-side noise. This synthesis suggests that G2G may begin to explain a unified understanding of robustness to these different sources of uncertainty. It also highlights a testable scientific hypothesis for future research: predictor-spaces generated from the \(S^{(1)} \rightarrow S^{(2)} \rightarrow S^{\prime(2)}\) model tend to exhibit low-rank-plus-diagonal properties (for empirical demonstration see Section~\ref{Sim}), implying that robustness to both Outcome Error and predictor-space noise will be more likely to occur in such datasets.

We note that while our derivation focuses on the binary case for tractability, it is well-established that continuous latent factor models also generate analogous low-rank-plus-diagonal covariance structures, which have similarly been shown to satisfy the spectral conditions for BO \citep{bing2021, bunea2022}. Thus, the validity of this general connection extends beyond the binary nature of our analysis.

\subsection{Systematic Realized Error and Rank}
\label{Systematic Error and Rank}

The preceding derivation of a low-rank-plus-diagonal structure hinges on the baseline (independent) error model. When Systematic Realized Error is present, the foundational assumptions are altered. The Systematic Realized Error acts as a common cause for the errors of a subset of predictors, violating the conditional independence required for a diagonal noise matrix. Consequently, \(C_{noise}\) will contain non-zero off-diagonal elements, and the favorable covariance structure is, in this form, lost.

However, the source of such Systematic Realized Error (\(L_R\)) can also be conceptualized as an additional latent variable. By expanding the conditioning set to include all latent drivers, \((S^{(1)},L_R)\), the covariance decomposition can be re-expressed as:

\[\Sigma_{S^{\prime(2)}} = Cov(E[S^{\prime(2)}|S^{(1)},L_R]) + E[Cov(S^{\prime(2)} | S^{(1)},L_R)].\]

\vspace{0.5\baselineskip}
In this formulation, the noise term, \(E[Cov(S^{\prime(2)}|S^{(1)},L_R)]\), is again diagonal, as errors are conditionally independent given both the primary states and \(L_R\). The signal term, \(Cov(E[S^{\prime(2)}|S^{(1)},L_R])\), is now driven by a latent layer that has \(k+1\) states, giving it an Algebraic Rank of \(2^{k+1}-1\) (for related proof, see Appendix~\ref{equivalence for LI viol}).

This conceptual shift has profound implications. It suggests that in the presence of systematic error -- whether it manifests as Systematic Error Regimes (modeled by \(L\)) or as Systematic Error Realizations (modeled by \(L_R\)) -- the data's covariance structure can be transformed into a low-rank-plus-diagonal form, where the rank metrics incorporate both \(S^{(1)}\) and these systematic error sources.

This reframing aligns perfectly with the capabilities of the highly flexible, automated modeling methodologies like those discussed in Section~\ref{Modeling Methodology}. To such a model, the pattern induced by a systematic error source is indistinguishable from one induced by \(S^{(1)}\). We hypothesize that such a model would likely model the error source as an additional latent driver, effectively isolating it and restoring conditional independence. In this context, the model is not imposing an unprincipled solution but rather discovering the expanded latent architecture that accounts for all sources of covariance.

This suggests that the connection between our framework and BO may hold even in the presence of systematic error, provided two conditions are met: (1) a sufficiently flexible model is used that has the capacity to identify and isolate sources of systematic errors (in addition to \(S^{(1)}\)), and (2) the combined number of first-stage latent states and systematic error sources is low relative to the predictor-space dimensionality, thereby keeping the Effective Signal Rank low relative to \(m\).

We note that this mechanism is analogous to the one proposed for handling structural violations of LI in Section~\ref{Absorption Hypothesis}, where direct dependencies between predictors can theoretically be absorbed by the model as additional ``pseudo-states.'' In both cases, the robustness of the architecture relies on a sufficiently flexible model's ability to expand the latent space to account for unmodeled dependencies, whether they originate from Predictor Error or the underlying causal structure itself.

\subsection{The Generative Origin of Low Rank: Effective Latent Complexity and Causal Consistency}
\label{The Generative Origin of Low Rank: Effective Latent Complexity and Causal Consistency}

Here we contextualize the generative conditions that allow real-world data to satisfy this requirement.

As detailed in Section~\ref{The Signal Component} and Appendix~\ref{latent sparsity}, the Effective Latent Complexity is inherently limited by the statistical properties of the latent layer itself. This provides a foundational level of latent sparsity, limiting the complexity of the problem, even in the presence of Chaotic Generative Variables. However, when the system is Causally Consistent, this fundamentally reduces the dimensional scaling of the system. Here we clarify this dynamic.

\subsubsection{Distinguishing Algebraic vs. Effective Signal Rank}
\label{Distinguishing Algebraic vs. Effective Signal Rank}

We distinguish between the dimensionality of the latent states (\(k\)) and the dimensionality of the signal subspace, or the Effective Signal Rank (\(r\)). The interpretation of \(r\) depends on the generative topology of the variables:

\begin{itemize}
\item
\textbf{Causally Consistent Systems:} When predictors act as monotonic proxies for latent \(S^{(1)}\) states directly, the Effective Signal Rank is structurally compressed such that \(r \rightarrow k\). This compression increases as the functions underlying the Causal Consistency become increasingly linear -- the magnitude of the compression relies on the monotonicity of the consistency; parametric functions that are highly non-monotonic (e.g., parity functions or complex XOR interactions), while Causally Consistent, would fail to compress the rank and would spectrally resemble a Chaotic Generative Variable. This increases the Spectral SNR, widening the spectral gap between the signal subspace and noise tail.
\item
\textbf{Chaotic Generative System:} When predictors act as unique identifiers for specific latent configurations, interactions inflate the Effective Signal Rank well beyond \(k\). However, for binary predictors, these non-linearities are captured within the linear span of the configuration space. Consequently, the Effective Signal Rank scales strictly with \(K_{rlzd}\) (or effectively with \(K_{eff}\)).
\end{itemize}

\textbf{Implication for Diagnostics:} This distinction validates the use of linear diagnostic tools (like the scree plot) even on binary data. While a linear tool cannot identify the number of \(S^{(1)}\) states in a Chaotic Generative System, the Effective Signal Rank \(r\) \textit{can} be used to identify the number of \(S^{(1)}\) configurations. Since BO relies on the model's ability to span the configuration space, the linear scree plot remains the appropriate heuristic for assessing architectural learnability in this binary treatment.

\subsubsection{The Dual Geometry of Robustness: Why Breadth Beats Depth}

This spectral decomposition clarifies the specific mechanical advantage of Breadth over Depth in the context of BO, where such generalization requires satisfying two simultaneous spectral conditions: a low-D, identifiable signal subspace (to ensure learnability) and a high-D noise subspace (to dissipate the energy of overfitting). We can now explain why Breadth is the unique strategy capable of optimizing both:

\begin{itemize}
\item
The Limit of Depth: A Depth strategy improves the fidelity of a fixed set of predictors. Geometrically, this lowers the magnitude of the noise eigenvalues (lowering the noise floor), thereby improving the Spectral SNR. However, because the effective dimensionality \(m\) is fixed, Depth does not expand the noise subspace. Consequently, it cannot increase the stable rank of the noise tail. While this aids signal recoverability, it fails to unlock the mechanism required to absorb label noise without contamination.
\item
The Power of Breadth (Expansion): Increasing \(m\) fundamentally alters the covariance geometry in two complementary ways:
\begin{enumerate}
    \item
    Lifting the Signal (via Informative Collinearity): As distinct but informatively redundant predictors are added, their shared signal variance sums constructively within the signal subspace. This effectively lifts the signal eigenvalues (\(\lambda_1, \dots, \lambda_k\)), widening the spectral gap (Spectral SNR).
    \item
    Expanding the Tail (via Distinctness): Crucially, because the predictors possess conditionally independent noise, adding them contributes new dimensions to the noise tail without increasing the magnitude of the largest noise eigenvalue. This increases the stable rank of the noise tail (\(R_{eff} \propto m\)).
\end{enumerate}
\end{itemize}

Thus, Breadth works towards two necessary preconditions for robust overparameterized learning (via BO): an identifiable low-rank signal and high-rank noise, while the Spectral SNR serves as the diagnostic metric confirming the separation between these two regimes.

\section{The Objective-Capability Gap and Representation Learning}
\label{The Objective-Capability Gap and Representation Learning}

Having established G2G's theoretical foundations we now address a critical question: how do models actually exploit this potential?

To examine this question, we highlight a contradiction in how modern ML is typically operationalized: there exists a productive tension between the mathematical objective of the gradient descent algorithm and the architectural capacity of the model -– standard training paradigms operate under a dissonance. The objective function -- typically the minimization of a monolithic scalar such as Cross-Entropy or Mean Squared Error -- operationalizes a flat strategy \citep{bishop2006pattern, goodfellow2016}. It effectively treats the data as if it fills the high-D Euclidean space, ignoring the geometric reality posited by the Manifold Hypothesis and our framework -- that the true signal is concentrated on a much lower-D latent structure. We term this mismatch ``Topological Dissonance:'' the objective function penalizes the deviation between the observed outcome and the prediction (\(\hat{Y}\)), mathematically aggregating all sources of uncertainty into a single scalar loss, without explicitly partitioning Predictor Error from Structural Uncertainty or attributing specific error components to unobserved latent states. Thus, the objective function contains no explicit incentive to model a latent structure; it optimizes purely for predictive accuracy. Simultaneously, however, the highly flexible ML models employed to solve this flat objective (e.g., Deep Neural Networks, Gradient Boosting Machines) possess hierarchical capacity -- formally recognized in the literature as a capacity for Representation Learning \citep{bengio2013representation}. They are architected with layers of abstraction that allow for the non-linear folding and transformation of the input space \citep{lecun2015deep}.

This dissonance resolves through an emergent phenomenon that favors the hierarchy: high-capacity models essentially ``hack'' the flat objective function. This hacking is not a defect but an elegant discovery: when data possess hierarchical structure, the path of least resistance to minimizing the monolithic error is not to memorize the noisy surface features of \(S^{\prime(2)}\), but to implicitly reconstruct the latent drivers of the data \citep{gunasekar2017implicit, arora2019implicit, geirhos2020shortcut}. Effectively, Representation Learning theories suggest that modern ML models uncover the optimal prediction function we discussed in Section~\ref{Primacy of the S(1) Layer and the Role of Interactions} (or a version thereof), aligning with our analysis that prediction via the latent hierarchy outperforms prediction via a flat topology -- a model that can disentangle the latent causes achieves a lower global error than a model that only has the capacity to treat predictors as ground truth \citep{shwartz2017opening}. In effect, the flexibility of the model works towards its prediction function escaping the Flat Topological Operationalization of the monolithic loss term by building a hierarchical topology in the weights. From a Bayesian perspective, one might argue that the model's architecture and optimization implicitly encode priors favoring hierarchical structure, thus resolving the apparent dissonance. We agree -- this convergence of implicit architectural priors with the latent structure is precisely what enables robust learning. A contribution of G2G is to formalize the data-architectural conditions under which these implicit priors align optimally with the generative process.

Clearly, the phenomenon of Representation Learning is well-established in the literature. To researchers familiar with these theories, the observation that deep networks discover latent structure may seem self-evident. However, what remains undertheorized is the precise data-architectural conditions under which this discovery is not merely beneficial but necessary for robust prediction, and the information-theoretic limits that distinguish successful from unsuccessful latent recovery. Thus, our contribution is to identify the specific data-architectural properties (hierarchical latent structure, Informative Collinearity, high-D Breadth) that enable this discovery to overcome both Predictor Error and Structural Uncertainty, thereby linking Representation Learning to predictive robustness in a formal, information-theoretic framework.

Additionally, existing Representation Learning literature largely focuses on its application to unstructured perceptual data (e.g., pixels in images, tokens in text). Its role in tabular domains -- often characterized by heterogeneous, noisy, and discrete predictors -- also remains undertheorized \citep{borisov2021deep, grinsztajn2022tree}. Thus, we also situate G2G as a theoretical bridge that extends Representation Learning into this tabular frontier. Specifically, we posit that Informative Collinearity in high-D tabular data serves the same architectural function as spatial correlation in images: it provides the necessary structural redundancy that allows the hierarchy to emerge. Ultimately, our framework begins to define the data-architectural prerequisites required for Representation Learning to function as a robust denoising mechanism in the context of tabular data that contains both Predictor Error and Structural Uncertainty.

Finally, we note that some training paradigms -- such as Variational Autoencoders, Expectation-Maximization algorithms, or structured prediction with latent variables -- do explicitly operationalize latent structure in their objective functions \citep{dempster1977, kingma2013auto, yu2009learning}. However, the dominant supervised learning paradigm (minimizing Cross-Entropy or MSE directly) does not, relying instead on model capacity to emergently discover hierarchical representations \citep{goodfellow2016}. Formalization of how the alignment between the objective function and latent generative mechanisms further enhances efficiency of the robustness mechanisms discussed in this manuscript is a clear direction for future research.

\subsection{The Engine and Fuel Analogy, Part 1}
\label{The Engine and Fuel Analogy, Part 1}

To clarify the relationship between model capacity and data architecture, we introduce a fuel-engine analogy that helps contextualize our framework overall, as well as the P-DCAI strategies introduced in Section~\ref{Towards Proactive Data-Centric AI}.

ML models are engines. This manuscript, in contrast, merely helps begin to define the chemistry of the fuel (i.e., the architecture of the data). Older, finely-tuned engines that require specific, delicate fuel mixtures are akin to linear models and other less flexible methodologies. Conversely, modern, multi-fuel engines (e.g., Deep Learning) are designed with massive horsepower potential and are robust to a range of fuel types -- they are mechanically capable of processing both refined and unrefined energy sources. Importantly, not all multi-fuel engines are equal. Just as a consumer-grade diesel generator differs from a massive industrial turbine, modern ML models vary significantly in their combustion capacity (e.g., parameter count, depth, and architectural priors). While both may technically possess the mechanism to process crude data, a lower-capacity engine may stall (underfit) or clog (fail to converge) when facing the thickest sludge of the data swamp, whereas a high-capacity industrial engine can successfully extract power from the same mixture. Ultimately, for a given multi-fuel engine, the performance output and efficiency in achieving that output depends entirely on the energy density of the fuel provided -- performance output is predictive performance, while efficiency relates to the resource costs (sample size, predictor dimensionality, computation, training time) required to achieve such performance.

Our core analyses facilitate understanding of why raw crude oil (e.g., a real-world data swamp) may contain the inherent chemical energy that allows a sufficiently powerful multi-fuel engine to achieve high performance. Adding nuance, our examination of efficiency suggests that, when using crude oil, such performance is likely achieved inefficiently.

We can thus view varied pre-processing philosophies through this fuel lens. Different strategies applied to the same source (the raw data swamp) produce datasets with different energy densities -- some over-refined and depleted (cleaning of a parsimonious subset), others with the full inherent chemical energy but also clogged with inert matter (the ``use everything'' approach), and others strategically enriched (P-DCAI). The practical application of G2G is that by understanding how fuel chemistry affects its inherent energy (the inherent potential for comprehensive, robust prediction), practitioners can deliberately engineer datasets that allow modern ML models to reach peak performance with minimal resource cost.

We extend this analogy further in Section~\ref{The Engine and Fuel Analogy, Part 2: P-DCAI as Strategic Refinement} after first presenting P-DCAI, a data engineering philosophy (data acquisition and feature selection principles) informed by this framework's data-architectural focus.

\section{Violations of Local Independence (LI)}
\label{Violating LI}

While the core derivation of G2G relies on the assumption of LI, real-world data-generating processes are rarely so pristine. Dependencies between predictors often exist due to direct causal links or shared systematic errors. However, a key strength of the proposed data-architectural perspective is that it offers a mechanism to mitigate these violations. This section details the ``Absorption Hypothesis,'' which posits that sufficiently flexible models can internalize these structural deviations, and outlines strategies to ensure this absorption contributes to robustness rather than overfitting.

\subsection{The Absorption Hypothesis}
\label{Absorption Hypothesis}

A core assumption of the framework is LI, which posits that the second-stage states are conditionally independent of each other given the first-stage states. This assumption is what allows the joint probability to be factorized as \(P(S^{(2)} | S^{(1)}) = \prod_{j = 1}^{m}{P(X_{j}^{(2)}|S^{(1)})}\). Violating this assumption has significant theoretical and practical implications for the highlighted robustness mechanism.

A violation of LI implies that the shared cause (\(S^{(1)}\)) does not fully explain the correlation among the \(S^{(2)}\) states. This residual dependency can arise from several real-world data-generating processes not captured by the primary structure, such as direct causal links (e.g., \(X_{i}^{(2)}\ \rightarrow X_{j}^{(2)}\)) or additional unmodeled latent causes.

The primary consequence of such a violation is that the realizations of Structural Uncertainty are no longer conditionally independent. Therefore, the affected predictors are no longer distinct with respect to that dependency. This can lead to information double-counting, which degrades the benefits of informative redundancy and diminishes the efficiency of Breadth by overestimating the amount of distinct evidence available.

However, we hypothesize that this structural challenge can potentially be overcome under certain conditions that require further characterization. We note that this Absorption Hypothesis is not a departure from the data-centric framework; rather, it is derived directly from it. We demonstrate that the data architecture resulting from such a violation is mathematically equivalent to a valid, LI-abiding architecture with a more complex latent layer. This hypothesis is formalized by showing that a system with direct causal links in the \(S^{(2)}\) layer is mathematically equivalent to a system with an expanded latent layer in which LI is restored.

\begin{raggedright}
\vspace{0.5\baselineskip}
\textbf{Proposition 1.} A system where LI is violated by a direct causal link between second-stage states is distributionally equivalent to an alternative system that incorporates the parent node of the dependency as a ``pseudo-state'' into an expanded latent layer (\(S_{exp}^{(1)}\)) with respect to which LI holds. See Appendix~\ref{equivalence for LI viol} for the full constructive proof.
\end{raggedright}

\vspace{0.5\baselineskip}
This logic can be extended from a single causal link to any directed acyclic graph of dependencies within the \(S^{(2)}\) layer. Each parent node in the dependency graph can be treated as a pseudo-state in an expanded first-stage layer, restoring LI for the entire system.

This result has profound implications. This data-architecturally informed, model-based mitigation strategy is analogous to the mechanism we proposed for handling systematic error in Section~\ref{Systematic Error and Rank}. In both cases, the framework's robustness relies on a sufficiently flexible model's ability to expand the first-stage latent space to account for unmodeled dependencies, whether they originate from the underlying causal structure (the LI violation) or from the observation process (systematic error). This motivates the \textbf{Absorption Hypothesis: that a highly flexible model has the theoretical potential to mitigate LI violations if it has the capacity to expand the latent space to include both the original \(S^{(1)}\) states and pseudo-states (representing ``rogue dependencies'' not contained by the primary structure)}, though the practical conditions for successful absorption remain to be fully characterized.

This mechanism converges with long-standing approaches in Psychometrics, where Method Factors or Bifactor structures are explicitly modeled to isolate systematic error from substantive traits \citep{Podsakoff2003, Reise2012}. It also aligns with recent methodological advancements that explicitly define and validate a ``bias absorption'' hypothesis within the context of Mixture Multigroup Structural Equation Modeling (MMG-SEM) \citep{PerezAlonso2024} -- they demonstrate that when a model is specified with sufficient flexibility by allowing for group-specific measurement parameters, it can effectively absorb systematic measurement non-invariances. This prevents the systematic error from confounding the identification of structural clusters, thereby allowing for the robust recovery of latent relations despite the presence of pervasive bias. Thus, our Absorption Hypothesis is not radical; it mirrors a mathematical property of latent variable models that is well-established in the rigorous fields of Psychometrics and SEM: that a sufficiently flexible model can internalize systematic deviations as distinct latent components, effectively neutralizing their impact on the primary inference.

Applying this analysis to modern ML further elevates the practical importance of automated modeling strategies discussed in Section~\ref{Modeling Methodology}. However, this potential mechanism introduces a critical tension. While it provides a pathway to mitigate structural violations, it does so at the cost of increased Effective Latent Complexity (\(K_{eff}\)). Successful application of this strategy likely depends on the interplay of the size of the expansion of the latent layer, the number of available predictors, the sample size, the model's capacity, and other factors affecting convergence. Crucially, excessive expansion of the latent space can inflate the required sample size (as analyzed in Section~\ref{Interplay of N and m}) and may violate the low-rank requirement for the favorable covariance structures linked to BO (Section~\ref{Benign Overfitting Link: Towards a Unified Understanding of Robustness}).

The viability of this absorption strategy also depends on the principle of Causal Consistency defined in Section \ref{Causal Consistency and Parametric Constraints}. When variables are Causally Consistent, the \(S^{(1)}\) states act as structural constraints of the true signal, strictly pushing signal rank from \(K_{rlzd}\) (or effectively, \(K_{eff}\)) to \(k\) as linearity increases (Section~\ref{Benefits of Causal Consistency}). Consequently, even if the number of pseudo states required to restore LI is large, so long as combined rank is low relative to \(m\) (a feat more likely when Causal Consistency compresses signal rank towards \(k\)), the model may still be able to absorb widespread structural violations without destroying the spectral gap required for BO. Examination of the pragmatic limits of absorption -- determining when the cost of complexity outweighs the benefits of absorption -- is a clear direction for future research and theoretical expansion.

Finally, we also note that absorption further complicates interpretability, as a model may learn a mixture of true and pseudo-states, making it difficult to distinguish between genuine common causes and artifacts of the dependency structure.

\subsection{The Risk of Malignant Overfitting}
\label{The Risk of Malignant Overfitting}

While the Absorption Hypothesis suggests that flexible models can mitigate systematic Predictor Errors by isolating them as pseudo-states (\(Z\)), this mechanism creates a critical vulnerability when the same systematic error also corrupts the outcome variable. In scenarios exhibiting this ``Common Method Variance'' (e.g., a hospital's coding practice affects both the patient history predictors \(S^{\prime(2)}\) and the diagnosis label \(Y^{\prime}\)), the associated pseudo-state \(Z\) will be highly correlated with the error-prone outcome, transforming the model's capacity to absorb it into a massive liability.

This creates a semantic gap: while the model can mathematically isolate the systematic error as a distinct latent feature (pseudo-state), it cannot semantically distinguish between a correlational cluster arising from a true latent driver and one arising from a systematic instrument error; both manifest as statistically identical latent features. To navigate this, we propose two distinct, yet complementary, strategic paths.

\begin{raggedright}
\vspace{0.5\baselineskip}
\textbf{Path A: The Traditional DCAI Strategy (Label Cleaning):} To guarantee the threat of Common Method Variance is resolved, practitioners can focus exclusively on direct label cleaning. This approach aligns with the core tenets of traditional Data-Centric AI, but is uniquely empowered by the principles of G2G.
\end{raggedright}

By improving the fidelity of the observed outcome \(Y^{\prime}\) -- thereby removing the systematic error from the label -- the correlation between the outcome and the pseudo-state \(Z\) is severed. While the rogue sources of covariation persist in the predictor-space, the Absorption Hypothesis may allow the model to isolate them and restore LI. Crucially, because this absorbed pseudo-state is no longer correlated with the cleaned outcome, the risk of malignant overfitting due to Common Method Variance is fully neutralized.

\begin{raggedright}
\vspace{0.5\baselineskip}
\textbf{Path B: The ``Metadata-Adversarial'' Strategy (Internal Control):} When measurement provenance is available, practitioners can leverage the Absorption Hypothesis to turn this vulnerability into a control mechanism. By explicitly including measurement source identifiers (e.g., sensor IDs, facility codes) as predictors during unsupervised Representation Learning, the model is incentivized to isolate systematic error into a distinct, identifiable pseudo-state rather than mixing it with true latent drivers.
\end{raggedright}

\begin{enumerate}
\item \textbf{Forced Absorption:} Encourage the isolation of the error variance via metadata. Explicitly include measurement source identifiers that are expected to ideally be unrelated to \(Y\) during the unsupervised Representation Learning stage.
\item \textbf{Semantic Identification:} Confirm which identified driver correlates with metadata, identifying it as \(Z\).
\item \textbf{Coefficient Zeroing:} When training the final predictive model for \(Y\), include \(Z\) to capture the Common Method Variance, but manually set its coefficient to zero during inference (or exclude it entirely if \(Y\) is believed to be unbiased).
\end{enumerate}

When metadata are available, this approach allows the model to utilize the full Breadth of the data while surgically removing the pseudo-states that would otherwise drive overfitting in the presence of Common Method Variance.

\section{The Geometry of Outcome Error}
\label{The Geometry of Outcome Error}

Though the primary structure is readily extended to include Observational Error in the outcome (e.g., \(Y^{\prime} \leftarrow Y \leftarrow S^{(1)} \rightarrow S^{(2)} \rightarrow S^{\prime(2)}\)) and predictive modeling clearly also relies on the fidelity of the outcome variable, this manuscript focuses primarily on mitigating predictor-side noise. However, we have touched on implications of Outcome Error throughout (Sections~\ref{Benign Overfitting Link: Towards a Unified Understanding of Robustness} and \ref{The Risk of Malignant Overfitting}). Here we provide a consolidated summary of these discussions, tying Outcome Error to the architecture geometry of the predictor-space.

Because Outcome Error is varied, its impact -- and the optimal engineering strategy to address it -- is strictly determined by its geometric relationship to the signal structure. Towards this understanding, we propose a spectral taxonomy of Outcome Error that classifies it into two distinct topologies: orthogonal (stochastic) and collinear (systematic). This distinction provides a decision matrix for when practitioners may be able to rely on the BO mechanism of architectural robustness versus when they must resort either to manual curation of the outcome variable or the Metadata-Adversarial strategy.

\subsection{Orthogonal Outcome Error (Stochastic)}
Orthogonal Outcome Error refers to random deviations in the outcome variable, stochastic measurement errors that are statistically uncorrelated with the \(S^{(1)}\) drivers or sources of systematic biases (\(Z\)) of the predictors.

\begin{itemize}
\item
\textbf{The Geometry:} Because this error is uncorrelated with the drivers of the predictor-space, it manifests geometrically as variance components that are orthogonal to the low-rank signal subspace. In the spectral domain, this energy resides in the high-D noise tail of the covariance matrix.
\item
\textbf{The G2G Connection to Benign Overfitting:} As detailed in Section~\ref{Benign Overfitting Link: Towards a Unified Understanding of Robustness}, the structure leverages high-D Breadth to essentially weaponize the noise tail. By expanding \(m\) the architecture provides a vast reservoir of orthogonal dimensions. As the BO mechanism posits, a sufficiently flexible model can effectively sequester stochastic Outcome Error into these benign dimensions, fitting the training labels perfectly without distorting the low-rank signal required for generalization.
\item
\textbf{Strategic Implication: Rely on Architecture.} In regimes dominated by Orthogonal Outcome Error, aggressive label cleaning yields diminishing returns. If the BO connection to G2G is valid and active, the data architecture of the predictor-side acts as a lightning rod, dissipating the label noise into the high-D tail. Consequently, the Dirty Breadth strategy remains valid; the cost of cleaning \(Y\) can be bypassed by simply expanding \(m\) efficiently (P-DCAI, Section~\ref{Towards Proactive Data-Centric AI}).
\end{itemize}

\subsection{Collinear Outcome Error (Systematic)}
Collinear Outcome Error refers to systematic deviations in the outcome variable that are correlated with specific patterns in the predictor-space (e.g., Common Method Variance, shared instrument bias, or demographic coding artifacts).

\begin{itemize}
\item
\textbf{The Geometry:} Because this type of Outcome Error correlates with observed predictors, it does not populate the noise tail. Instead, it masquerades as a low-rank signal component, residing in the same dominant subspace as the latent drivers.
\item
\textbf{The Failure Mode (Malignant Overfitting):} This creates a Semantic Gap. To the model, the vector of systematic error is mathematically indistinguishable from a vector driven by \(S^{(1)}\). No amount of architectural Breadth can filter this, as the Outcome Error mimics the spectral signal's geometry. The model will efficiently learn this bias as a primary predictive feature, leading to robust but invalid predictions.
\item
\textbf{Strategic Implication:} In this regime, G2G's architectural robustness fails spectacularly. Intervention is mandatory. The practitioner must deploy one of two strategies as discussed in Section~\ref{Absorption Hypothesis}: label curation (the DCAI Approach) or Metadata-Adversarial Modeling.
\end{itemize}

\subsection{A DCAI-Aligned Resource Shortcut: ``Dirty Inputs, Clean Targets''}
\label{A DCAI-Aligned Resource Shortcut: ``Dirty Inputs, Clean Targets''}
This taxonomy reveals a massive asymmetry in the economics of data cleaning. Because high-D Breadth can architecturally absorb predictor-side noise, the marginal return on cleaning features is low. Conversely, because systematic label noise can induce Malignant Overfitting (Section~\ref{The Risk of Malignant Overfitting}), the marginal return on cleaning the outcome is infinite. Therefore, the optimal strategy is asymmetric: \textbf{``Dirty Inputs, Clean Targets.''}

By focusing curation resources exclusively on the single outcome column to eliminate systematic bias, the practitioner is liberated from the prohibitive task of cleaning a high-D predictor-space. Instead, resources are allocated to the high-fidelity curation of \(Y\), creating a clean label that allows the model to safely utilize a massive, uncurated predictor-space to mitigate predictor-side noise via the G2G mechanisms.

This resource shortcut resolves the tension between DCAI and P-DCAI, positioning label cleaning not as a rejection of our theory, but as a powerful enabling constraint that allows the architectural robustness of G2G to function safely in contexts characterized by Common Method Variance.

\section{Proactive Data-Centric AI: Implications Related to Data Collection and Feature Selection}
\label{Towards Proactive Data-Centric AI}

This framework presents a central tension: our theoretical analysis proves that robustness approaches an ideal limit as predictor dimensionality approaches infinity, suggesting a ``use everything'' strategy. However, our analysis of efficiency reveals that the rate of convergence is dependent on predictor quality. This efficiency rate is the critical factor for real-world applications, which are constrained not by theory but by finite computational budgets and finite model capacity.

P-DCAI is proposed as a pragmatic approach to bridge the gap between this theoretical ideal and engineering reality, though its effectiveness requires empirical validation across domains. It is a strategy focused on two simultaneous objectives: completeness (maximizing latent coverage to minimize Structural Uncertainty) and efficiency (accelerating the convergence rate). P-DCAI aims to strategically select or collect a set of predictors that delivers the highest density of novelty for completeness (construct validity) and informative redundancy for reliability, thereby achieving maximum robustness for a given computational cost.

Ultimately, computational considerations should not be viewed as limitations of G2G. Rather, they are a central motivation. It is precisely because computational resources and model capacity are finite that a naive ``use everything'' approach is impractical and often suboptimal. This reality elevates P-DCAI from a simple best practice to a practical necessity, justifying the upfront investment in G2G-informed feature selection strategies that prioritize the efficiency of robustness convergence.

Overall, G2G suggests a philosophical shift from a reactive stance (cleaning and removing Observational Error) to a proactive one (strategically designing data acquisition and feature selection strategies in order to build a dataset whose inherent potential can be unlocked by a sufficiently flexible model).

\begin{longtable}{@{}
>{\raggedright\arraybackslash}p{(\linewidth - 6\tabcolsep) * \real{0.15}}
>{\centering\arraybackslash}p{(\linewidth - 6\tabcolsep) * \real{0.29}}
>{\centering\arraybackslash}p{(\linewidth - 6\tabcolsep) * \real{0.35}}
>{\centering\arraybackslash}p{(\linewidth - 6\tabcolsep) * \real{0.29}}@{}}
\caption{Comparative Analysis of Proposed Framework vs. Conventional Approaches.} \label{Table: Comparative Analysis of Proposed Framework vs. Conventional Approaches}\\
\toprule\noalign{}
\begin{minipage}[b]{\linewidth}\centering
\textbf{Aspect}
\end{minipage} & \begin{minipage}[b]{\linewidth}\centering
\textbf{Conventional Wisdom or Approach}
\end{minipage} & \begin{minipage}[b]{\linewidth}\centering
\textbf{Proposed Reinterpretation or Approach}
\end{minipage} & \begin{minipage}[b]{\linewidth}\centering
\textbf{Key Distinguishing Feature or Novelty Claim}
\end{minipage} \\
\midrule\noalign{}
\endhead
\bottomrule\noalign{}
\endlastfoot
\textbf{High-D}
& Often viewed as a curse; leads to data sparsity, overfitting, computational burden. Aim to limit \(m\).
& Can be an asset by providing more opportunity for novel and/or redundant information to improve the recovery of \(S^{(1)}\).
& Reinterprets high-D as beneficial for \(S^{(1)}\) recovery, not as inherently problematic.
\\
\hline
\textbf{Collinearity}
& Though not always harmful to prediction, considered primarily problematic for model interpretation. Mitigated by feature removal or transformation.
& Informative Collinearity is beneficial because it can both improve reliability of recovering \(S^{(1)}\) and increase the efficiency of asymptotic convergence. This is distinct from trivial collinearity.
& Views a specific type of collinearity as a signal that can be exploited for robustness, rather than solely a statistical nuisance.
\\
\hline
\textbf{Predictor Error}
& A primary problem requiring aggressive data cleaning, validation, imputation, or removal of error-prone features.
& Can be mitigated through thoughtful predictor-space design (P-DCAI) and appropriate modeling techniques when cleaning is impractical to scale.
& Shifts focus from error removal to error mitigation through strategic use of data characteristics and model choice.
\\
\hline
\textbf{Structural Uncertainty}
& Not explicitly considered. Instead is treated as part of a monolithic noise term.
& Formally separated from Observational Error as an independent source of predictive limits.
& Proves that for a fixed set, Structural Uncertainty is a hard limit that persists despite perfect cleaning; only Breadth resolves it.
\\
\hline
\textbf{Feature Selection Goal}
& Minimal set of relevant, ideally uncorrelated features that directly and strongly predict \(Y\). Prioritizes parsimony and interpretability.
& Strategic selection of sufficiently numerous, novel, and informatively redundant predictor variables. When the outcome is known and has high fidelity, features can be prioritized by their ability to resolve the predictive weakest link.
& Goal shifts from direct prediction of \(Y\) to robust \(S^{(1)}\) estimation.
\\
\hline
\textbf{Data Collection Strategy}
& Focus on acquiring a few clean, direct, and strong correlates of \(Y\).
& Prioritize acquiring multiple, diverse, informatively redundant predictor variables, as well as those with diverse error sources. Considers targeted redundancy for weakly estimated \(S^{(1)}\) states/configurations.
& Values a vast, diverse, and structurally informed portfolio of (potentially imperfect) predictors over a few perfectly clean measures.
\\
\hline
\textbf{Primary Modeling Goal} & Often unbiased parameter estimation, model fit,
interpretability of individual effects, or direct prediction of \(Y\) by
observed features. & Robust prediction of \(Y\), achieved primarily
through improved estimation of latent \(S^{(1)}\) states to combat Predictor Error and
Structural Uncertainty. & The primary utility of high-D and Informative Collinearity is tied to enhancing \(S^{(1)}\) estimation for the purpose of robustly predicting \(Y\).\\
\hline
\textbf{Theoretical Basis for Handling Uncertainty or Error in the Data}
& Often relies on statistical assumptions about error distributions, error correction models, or removal/down-weighting of error-prone data.
& Employs Information Theory to provide a rationale for how and why high-D can mitigate the impact of both Predictor Error and Structural Uncertainty.
& Highlights the importance of novel for complete coverage (construct validity) and redundancy for reliability in mitigating both Predictor Error and Structural Uncertainty.
\\
\hline
\textbf{Data as Fuel Metaphor}
& Garbage In, Garbage Out. Data are viewed as a static input; any impurity contaminates the result.
& Crude Oil vs. Refined Fuel. Data has inherent chemical energy; raw, dirty data (crude) contains high-density signal that modern engines can refine, whereas over-cleaning (Clean Parsimony) removes impurities but depletes the energy density.
& Reconceptualizes dirty data not as waste, but as unrefined fuel potentially containing the raw materials for robust signal reconstruction.
\\
\end{longtable}

\subsection{Implications for Feature Selection}
\label{FeatureSelection}

The implications summarized in Table~\ref{Table: Comparative Analysis of Proposed Framework vs. Conventional Approaches} argue for a paradigm shift in how researchers approach feature selection. To clarify, we do not advocate for the universal abandonment of data curation practices. Rather, we suggest a recalibration of priorities. This framework allows the focus to shift from an overwhelming emphasis on data reduction and aggressive cleaning -- often impractical at scale -- towards a nuanced strategy of leveraging the data architecture. This shift is particularly pertinent when dealing with datasets that are exceedingly complex, vast in scale, and exhibit high degrees of correlation among large numbers of error-prone variables.

Instead of conventional approaches that aim to create a minimal set of directly relevant, ideally uncorrelated and error-free predictors of \(Y\) -- a philosophy embodied by methods like Minimum Redundancy Maximum Relevance (MRMR) which explicitly penalize redundancy \citep{Peng2005} -- our framework suggests that the primary goal of feature selection should transform to the strategic leveraging of high-D novelty and informative redundancy for complete, reliable, and efficient \(S^{(1)}\) recovery.

We acknowledge that a fundamental challenge in operationalizing this framework is that the theoretical definitions of novelty and informative redundancy (Section~\ref{``High-Quality'' Variables, ``Uniqueness,'' and ``Efficiency''}) are based on the relationships between the unobserved \(\gamma\) vectors (\(\varphi\)) (which govern the \(S^{(1)} \rightarrow S^{(2)}\) pathways), but in practice, researchers only have access to the observed, error-prone data. Therefore, direct calculation of these theoretical quantities is impossible. Nevertheless, the goal of G2G-informed feature selection is to use strategies -- both theoretically driven and automated -- that serve as practical proxies for these idealized concepts, recognizing that the presence of Observational Error further obscures the true latent relationships.

\subsubsection{Theoretically Driven Feature Selection}

These strategies utilize domain knowledge for hypothesizing the latent \(S^{(1)}\) states and identifying candidate \(S^{\prime(2)}\) variables. Though manual implementation is required, the primary strength of these strategies is their direct alignment with established scientific theory, ensuring that the selected features have substantive meaning and that the resulting model is interpretable. Several such strategies can be employed:

\begin{itemize}
\item
\textbf{\(S^{(1)}\)-Centric Feature Engineering:} The primary goal is to select sets of \(S^{\prime(2)}\) variables that best estimate each hypothesized \(S^{(1)}\) state.
\item
\textbf{Prioritizing by Theoretical Signal Strength:} This strategy involves selecting features by prioritizing the theoretical correlation strength of the \(S^{(1)} \rightarrow X_{j}^{(2)}\) link (high Structural Strength) even if Predictor Error is suspected to be high. While lower Predictor Error is beneficial, it should not be the primary filter when pursuing a Breadth strategy.
\item
\textbf{Protecting Low and High Prevalence Indicators:} This strategy involves selecting features by considering not just the presumed Structural Strength, but also the prevalence rates of the indicators. Due to the formulation of the phi correlation, a variable with a low or high prevalence rate might be highly correlated with \(S^{(1)}\) states even if its theoretical signal strength is expected to be somewhat low.
\item
\textbf{Protecting Indirect Indicators:} This involves being cautious about discarding predictor variables that have weak direct correlations with \(Y\) but are strongly correlated with other variables that are strong predictors of \(Y\).
\item
\textbf{Redundancy-Preserving Aggregation within \(S^{(1)}\)-Proxy Sets:} If computational constraints force a reduction in dimensionality, one might use techniques (like Principal Component Analysis) to aggregate proxies for the same \(S^{(1)}\) state (or configuration). However, caution is warranted: manual aggregation collapses the distinct error distributions of the individual predictors, potentially destroying the fine-grained information a flexible model (e.g., a neural network) needs to isolate specific failures. Thus, this strategy is best reserved for scenarios involving linear models or strict computational limits; otherwise, providing the raw, disaggregated inputs to a flexible model is preferred to allow the model to learn the optimal aggregation function itself.
\item
\textbf{Maximizing Diversity of Error Sources:} When multiple predictor variables are believed to be comparable proxies for the same \(S^{(1)}\), but some must be discarded (e.g., due to computational limitations or cost), prioritize retaining a set that originates from different observational sources or methodologies.
\end{itemize}

\subsubsection{Automated Feature Selection}
\label{Automated Feature Selection}

While powerful, manual, theory-driven approaches can be challenging to scale and may not capture all empirically relevant relationships in the data. In contrast, automated feature selection methods, driven by empirical evidence in the data, offer a more scalable and potentially more comprehensive utilization of the available data. Such automated strategies can be designed to select variable structures that are likely to align with G2G, but may also advantageously identify variables that do not strictly fall into the framework yet nevertheless hold predictive utility (see Section~\ref{Absorption Hypothesis}). Several such strategies are particularly well-suited to this:

\vspace{0.5\baselineskip}
\begin{raggedright}
\textbf{General-Purpose Strategies (Unsupervised):} When the objective is to create a flexible dataset capable of predicting various future outcomes (e.g., a general-purpose clinical data warehouse), general-purpose strategies prioritize the complete recovery of the latent layer \(S^{(1)}\) rather than optimizing for a specific \(Y\) -- they work towards reinforcing the causal structure underlying the predictor-space. Whether such causes represent true \(S^{(1)}\) states, \(S^{(1)}\) configurations, or pseudo-states that restore LI in the presence of structural violations (Sections~\ref{Systematic Error and Rank} and \ref{Absorption Hypothesis}), ultimately, capturing all such causes is important for improving the reliability of \(S^{(1)}\) recovery.
\end{raggedright}

\begin{itemize}
\item
\textbf{Principled Iterative Expansion:} An automated process can be designed to mirror G2G’s theoretical goals: achieving both novelty for completeness and redundancy for reliability. This can be implemented as a two-phase strategy:
\begin{enumerate}
    \item
    Phase 1 (Novelty-First Anchoring): The initial goal is to select an ``anchor set'' of predictors that provides broad coverage of the latent \(S^{(1)}\) layer. Methods explicitly designed for latent structure discovery, such as selecting the highest-loading item from each factor identified via Exploratory Factor Analysis (when Causal Consistency is assumed) or utilizing spectral clustering, provide a philosophically aligned approach to anchor selection. The size of this anchor set can be guided by the understood complexity of the latent space -- for instance, by selecting a number of predictors (at least one from each cluster) approximately equal to the effective number of configurations (\(K_{eff}\)).
    \item
    Phase 2 (Redundancy-Reinforcement Expansion): Once the anchor set is established, the strategy shifts to enhancing reliability by leveraging informative redundancy. This phase iteratively expands the feature set by adding new predictors that are the most highly correlated (i.e., informatively collinear) with the predictors in the initial anchor set or those most representative of the extracted factor or cluster they represent. Such iterative expansion may also be guided by the concept of the weakest link (see Appendix~\ref{finite convergence}), where, for example, the observed centroids of clusters of the selected variables are compared to the centroids estimated from the entire predictor-space, and the weakest cluster (the one with the largest discrepancy) is targeted for reinforcement. Such procedures directly operationalize G2G principles: by adding distinct indicators that respond similarly to the \(S^{(1)}\) configurations (as evidenced by their high correlation or common cluster membership), the strategy reinforces the signals from the anchor set which were selected to provide complete coverage of \(S^{(1)}\). This reinforcement is essential for accelerating the efficiency of \(S^{(1)}\) recovery.

    As an alternative expansion strategy (once the anchor set is selected), when examining the remainder of \(S^{\prime(2)}\), a first step could be to perform a simple, computationally cheap filtering of the remaining predictor pool to remove any features that are perfect duplicates or linear combinations of others. Once these obvious sources of trivial collinearity are removed, using the remaining variables, the strategy would iterate through different subsets of \(S^{\prime(2)}\) (where the anchor set is always maintained) to find the set that maximizes the ``Relative Redundancy'' ratio: the Total Correlation (\(TC(S^{\prime(2)})\)) divided by the sum of the marginal entropies, \(\sum_{j = 1}^{m}{H(X_{j}^{\prime(2)})}\) -- see Statement 2d. This scale-invariant ratio quantifies the degree of architectural structure in the data. Due to the prohibitive computational requirements of this strategy if implemented naively, however, novel algorithmic strategies are necessary to efficiently guide iterations to maximize Relative Redundancy.
\end{enumerate}

\item
\textbf{Empirical Identification of Indicator Clusters:} This strategy also utilizes automated methods like clustering or exploratory factor analysis on the predictor correlation matrix to empirically identify groups of highly correlated \(S^{\prime(2)}\) variables. Feature selection can then simply focus on retaining the set of indicators from each identified cluster that load above a predefined threshold.
\end{itemize}

Importantly, general-purpose strategies serve a dual purpose. Beyond improving the reliability of \(S^{(1)}\) recovery, they also work to proactively amplify the favorable spectral properties linked to BO. By prioritizing internal consistency of the predictors (independent of \(Y\)), these strategies ``stack the deck'' with informative redundancy. This effectively concentrates variance into the signal subspace, widening the eigengap and enhancing the signal-to-noise ratio of the selected set relative to the raw predictor-space. This makes general-purpose strategies particularly important in scenarios where the outcome variable is suspected to contain significant Observational Error, as they work towards inducing the structural conditions required for robustness to Outcome Error via the BO mechanism.

\begin{raggedright}
\vspace{0.5\baselineskip}
\textbf{Outcome-Specific Strategies (Supervised):} Conversely, when the outcome variable is defined and available, the objective can be to maximize the predictive efficiency for that specific outcome.
\end{raggedright}

\begin{itemize}
\item
\textbf{Targeted Residual Expansion (Predictive Efficiency):} While general-purpose strategies aim to recover the complete latent structure (\(S^{(1)}\)), the most efficient path to robust prediction focuses solely on distinguishing latent configurations that provide conflicting information about \(Y\). This strategy operationalizes the concept of Predictive Efficiency, effectively weighting the value of latent recovery by the Mutual Information between specific latent configurations and the outcome.
\begin{enumerate}
    \item Proxy Inference: At each step, a proxy model is trained on the currently selected predictors to generate a prediction \(\hat{Y}\).
    \item Residual Computation: The algorithm calculates the residuals (\(Y - \hat{Y}\)), which represent the currently unexplained variance -- mathematically analogous to the ambiguity remaining in the latent space. Crucially, residuals are large only when the model conflates latent configurations that imply different outcomes (high Mutual Information with \(Y\)). If two configurations are indistinguishable but imply the same outcome (low Mutual Information with \(Y\)), the residual is negligible.
    \item Targeted Selection: The algorithm scans the candidate pool to identify the predictor that exhibits the maximum mutual information with these residuals, inherently targeting the predictive weakest link. By selecting predictors that correlate with these residuals, the algorithm identifies the specific areas where the current data architecture is structurally blind, adding the exact Breadth required to lower the local Performance Floor for \(Y\).
\end{enumerate}

This ensures that the feature set expands not just to resolve structural ambiguity, but specifically to resolve the ambiguities that possess the highest information value regarding \(Y\). Notably, this logic mirrors the mechanism of Gradient Boosting Machines (e.g., XGBoost, LightGBM) \citep{chen2016xgboost}, which iteratively fit new weak learners to the residuals of the previous ensemble. In the context of P-DCAI, Targeted Residual Expansion can be viewed as a data-centric analogue to boosting: rather than adding complexity to the model to fit the residual, we add specific information to the dataset to explain it.

However, this pursuit of predictive efficiency rests on a critical assumption: that the outcome variable is high-fidelity. This strategy operates on the premise that any large prediction residual represents a predictive weakest link. However, if \(Y\) contains significant Observational Error (is observed as a low fidelity \(Y^{\prime}\)), these residuals will reflect this error rather than a latent ambiguity. In such contexts, this strategy risks ``malignant overfitting'' -- curating a feature set specifically optimized to fit measurement artifacts in \(Y^{\prime}\) rather than the latent drivers. Therefore, targeted expansion is best reserved for applications where the outcome labels are known to be accurate, whereas general-purpose strategies are safer when the quality of \(Y\) is uncertain or known to be error-prone.
\item
\textbf{Two-Stage Selection via Latent State Validation:} This workflow combines unsupervised structure discovery with supervised validation. First, potential latent states are estimated using automated anchoring methods. Second, these states are validated against \(Y\); if predictive, the system retains the associated feature cluster, potentially expanding it via redundancy-reinforcement. Crucially, this strategy mitigates the risk of malignant overfitting associated with Targeted Residual Expansion. Because candidate features are defined by their mutual information with the predictor set rather than the outcome, the system effectively filters out features that might otherwise be selected solely for their chance correlation with Observational Errors in \(Y\).
\end{itemize}

\begin{raggedright}
\textbf{Summary: The Efficiency-Robustness Trade-off.} The choice between general-purpose and outcome-specific feature selection implies a fundamental trade-off between computational efficiency and robustness to Outcome Error.
\end{raggedright}

\begin{itemize}
\item
Use General-Purpose or the Two-Stage Selection via Latent State Validation strategies when \(Y^{\prime}\) is error-prone (if unknown use General-Purpose only). While resulting in a feature set that is computationally more demanding to model, these approaches prioritize the construction of a predictor covariance structure that works towards conditions that enable BO, potentially utilizing the inefficiency of redundant predictors as a buffer against Outcome Error.
\item
Use Targeted Residual Expansion strategy when confidence in the accuracy of \(Y^{\prime}\) is high and computational or model capacity constraints are tight. This minimizes \(m\) but assumes the weakest link is structural, not observational.
\end{itemize}

\subsection{Implications for Data Acquisition}

The principles derived from G2G also extend to data acquisition practices. There are substantial benefits in prioritizing the collection of multiple, diverse -- albeit imperfect -- proxies of \(S^{(1)}\) states. The emphasis here is on gathering a diverse portfolio of \(S^{\prime(2)}\) variables that are expected, based on domain theory or empirical evidence, to have stronger correlations with different facets of the \(S^{(1)}\) layer or with other variables that are themselves expected to be strongly related to facets of the \(S^{(1)}\) layer, thereby better ensuring complete and reliable coverage of \(S^{(1)}\). Another key principle of this philosophy is maximizing the diversity and/or tracking of error sources. This improves the likelihood of the independent error model, reduces the likelihood of pervasive systematic errors, and may aid in the identification of the systematic error source if one exists.

Overall, G2G suggests a data collection philosophy that values a richer, multi-faceted, and structurally informed representation of predictor constructs over a narrow pursuit of a few, supposedly error-free, direct measures.

\begin{itemize}
\item
\textbf{Targeted Acquisition for Reliability and Efficiency:} If domain knowledge suggests a particular latent first-stage state (\(X_{i}^{(1)}\)) is critical for predicting \(Y\), but its current observable proxies are few or highly noisy, additional distinct measures should be acquired to improve reliability. Furthermore, to maximize the efficiency of \(S^{(1)}\) recovery, acquisition efforts should prioritize variables that help resolve the weakest link as this bottleneck governs the overall convergence rate.
\item
\textbf{Multi-Instrument or Multi-Modal Measurement of \(S^{(2)}\) States:} Actively seek to measure using different instruments, raters, or data modalities.
\item
\textbf{Prioritizing Collection of Proxies with High Structural Strength:} When resources are limited, prioritize the collection of \(S^{\prime(2)}\) variables that are expected to have the strongest underlying Structural Strength (strongest correlations with their theorized \(S^{(1)}\) source states). This focus on inherent relevance often supersedes prioritizing variables with marginally higher presumed observational fidelity, as architectural strategies are designed to compensate for the latter.

\item
\textbf{Deliberate Acquisition of Error-Orthogonal Predictors:} If a known systematic error source is suspected to affect a subset of variables, make a targeted effort to acquire data for other potential proxies of the same \(S^{(1)}\) states from different sources.
\item
\textbf{Track Measurement Sources:} If observational/measurement sources are tracked, these can be used to explicitly incorporate the potential sources of error into the model.
\item
\textbf{Systematic Exploration for Latent Construct Indicators:} Actively search for sets of variables that are highly inter-correlated, as these may indicate underlying \(S^{(1)}\) states. Data collection could then focus on enriching these clusters.
\item
\textbf{Leveraging Low-Cost, High-Volume Data Sources:} The framework suggests utility in the acquisition of variables that are individually noisy but operationally inexpensive to collect at scale. Because the G2G mechanism relies on the aggregation of evidence to overcome error, the inclusion of data from automated sources -- such as web scrapers, wearable sensors, or IoT devices -- is theoretically justified even if their Structural Strength and fidelity is low. This suggests a shift in budget allocation: rather than spending heavily to perfect a small set of manual measurements, resources may be better deployed acquiring a vast Breadth of inexpensive, automated proxies to triangulate the latent signal.
\end{itemize}

\subsection{Active Learning and Optimal Design}

Our framework's emphasis on acquiring a diverse array of correlated imperfect \(S^{\prime(2)}\) variables naturally leads to considerations of active learning \citep{settles2012active} and optimal experimental design \citep{pukelsheim2006optimal}. When data collection is an ongoing process or involves choices under resource constraints (e.g., budget, time, participant burden), these principles can offer a strategic approach. Instead of randomly or exhaustively collecting all possible \(S^{\prime(2)}\) variables, active learning algorithms or optimal experimental design methodologies could be employed to guide the selection of which new variables to acquire or which specific experiments to conduct next.

The objective would be to choose those variables or experiments anticipated to yield the maximum information gain about \(S^{(1)}\). This allows for a more efficient and targeted data acquisition strategy, prioritizing measurements that are most likely to refine the \(S^{(1)}\) estimates and, consequently, enhance the robustness of predicting \(Y\), particularly when the cost or feasibility of acquiring all potentially relevant \(S^{\prime(2)}\) variables is prohibitive.

\subsection{The Engine and Fuel Analogy, Part 2: P-DCAI as Strategic Refinement}
\label{The Engine and Fuel Analogy, Part 2: P-DCAI as Strategic Refinement}

Now that the P-DCAI philosophy has been presented, here we complete the engine and fuel analogy introduced in Section~\ref{The Engine and Fuel Analogy, Part 1}.

We hypothesize that producing refined data fuel via G2G-informed strategies encourages the multi-fuel engine to utilize its capacity for Representation Learning, making the reconstruction of \(S^{(1)}\) not merely a possible solution to the flat objective, but the clear path of least resistance. This is not meant to imply prior work was mistaken in their focus on model- or algorithm-centric approaches. Rather, we are arguing that data-architectural design constitutes an orthogonal axis of optimization that has been underexplored in theoretical treatments.

Just as a high-performance engine cannot deliver peak output on low-grade gasoline, the hierarchical capacity of a deep network cannot effectively resolve latent structures if the input data lack the necessary architectural properties. To concretize this mechanism, we categorize three distinct data paradigms through the lens of fuel chemistry:

\begin{enumerate}
\item
\textbf{Traditional Cleaning (Synthesized Laboratory Gas)}: The traditional paradigm tends to over-refine the fuel. Because manual data cleaning is labor-intensive, prioritizing it acts as a bottleneck, forcing practitioners to curate a limited subset of variables. This curation process logically prioritizes novelty, discarding collinear variables to achieve maximum conceptual coverage with minimum redundancy.

While this strategy is efficient in the context of the data cleaning bottleneck, ultimately, it is energetically destructive. By systematically removing Informative Collinearity, the practitioner strips away the complex molecular chains required to triangulate latent drivers. The resulting fuel is chemically pure (free of error) but structurally depleted (low energy density), leaving the engine powerless against the Performance Floor of Structural Uncertainty. Consequently, while the multi-fuel engine runs smoothly, it hits a hard speed limit. It is effectively starved of signal, unable to reach its design potential because the fuel lacks the informational density to support the engine's full capacity.
\item
\textbf{The Data Swamp (Raw Unfiltered Crude)}: The raw, high-D data swamp is akin to extracting fuel directly from the source. It is thick, messy, and filled with the inert sludge of irrelevant variables. While the massive, complex hydrocarbon chains (the complete latent signal) are present, offering potential for maximum power, the mixture is volatile. A linear model (a finely-tuned engine) lacks the systems to handle this sludge and fails immediately. A modern ML model (the multi-fuel engine) possesses the necessary capacity to run on this crude and eventually reach maximum speed. However, this comes at a cost: the model must divert significant energy to separating the signal from sludge during operation. The process is computationally expensive, inefficient, and requires a massive volume of fuel to sustain output.
\item
\textbf{P-DCAI (Enriched Heavy Fuel)}: P-DCAI does not refine the crude into clean gas (removing error); instead, it processes the crude into a centrifuged concentrate. It spins out the inert sludge that does not contribute to coverage, but intentionally retains the dirty, complex heavy chains of Informative Collinearity. P-DCAI constructs a dataset that ensures complete and redundant coverage of the latent energy sources (\(S^{(1)}\)), tolerating the ``particulates'' of Predictor Error to preserve high-energy structure. The result is a fuel with high energy density and a more optimized signal-to-noise ratio relative to crude. With this enriched fuel, the multi-fuel engine hits maximum speed with less effort than with crude. It no longer wastes capacity filtering out inert sludge, unlocking the inherent potential of the raw swamp with less drag. This achieves robustness with significantly greater efficiency than the raw data, and higher predictive performance than the clean parsimonious data.
\end{enumerate}

\subsection{Methodology Transfer: A Democratized Deployment Paradigm}
\label{Methodology Transfer: A Democratized Deployment Paradigm}

The principles of G2G -- specifically the reliance on high-D Breadth to resolve Structural Uncertainty -- reveal a fundamental theoretical limitation of the dominant ``Model Transfer'' paradigm (including the deployment of General Purpose or Foundation Models).

The conventional goal of Model Transfer is to train a single model on data from Site A (or a centralized aggregate of sites) and deploy it to Site B. To ensure transportability, the model is necessarily restricted to the set of predictors shared by all sites. Let \(\mathcal{F}_{site}\) denote the available feature set at a specific location. The feature set available to a universal model, \(\mathcal{F}_{univ}\), is the intersection of all local sets: 
\[ \mathcal{F}_{univ} = \bigcap_{i} \mathcal{F}_{site_i}. \]

In contrast, a ``Local Factory'' (Methodology Transfer) operates on the full unique feature set of the specific deployment site, \(\mathcal{F}_{local}\). 

This distinction reveals why Local Factories are theoretically positioned to outperform universal models, regardless of the sophistication of the pre-training:

\begin{enumerate}
\item
\textbf{The Intersectional Constraint:} By definition, \(|\mathcal{F}_{univ}| \le |\mathcal{F}_{local}|\). In domains characterized by high heterogeneity (e.g., healthcare, finance), this intersection is often small, reducing the predictor space to a lowest common denominator. As proven in Statement 1 and Appendix~\ref{asymp Breadth v Depth}, reducing \(m\) reduces Structural Uncertainty. Universal models are thus brittle not necessarily because they fail to learn the training data, but because the \textit{standardization} required for their existence forces the removal of the high-D Breadth required to triangulate the latent signal.
\item
\textbf{Contextual Coefficient Stability:} Even within the shared intersection \(\mathcal{F}_{univ}\), the functional relationship between a predictor and the latent drivers often varies by locale (e.g., a specific billing code may be used aggressively at one hospital and conservatively at another). A universal model must learn a global average coefficient, which effectively acts as an additional source of noise when applied locally. A Local Factory learns the site-specific coefficients, turning local operational variance into accurate signal.
\item
\textbf{Artifacts as Features:} Perhaps most critically, Model Transfer treats site-specific idiosyncrasies (e.g., local administrative artifacts, unique sensor noise patterns) as distribution shifts to be rejected. G2G reframes these artifacts. Provided they are distinct and structurally linked to \(S^{(1)}\), these idiosyncrasies represent valid sources of Informative Collinearity. A Local Factory absorbs these artifacts as useful proxies, leveraging the site's specific data generation quirks to triangulate the latent truth, rather than failing because of them.
\item
\textbf{Dynamic Temporal Calibration (Combating Drift):} Finally, Model Transfer typically treats the model as a static artifact frozen at the time of initial training. However, the relationship between predictors and latent drivers is often non-stationary (e.g., changes in clinical protocols, coding updates, or market regimes). Even a perfectly localized model will degrade if it is static. Because a Methodology Transfer exports the production line, not the product, the Local Factory is potentially able to continuously retrain on the live local data stream. Consequently, the model dynamically adapts to temporal drift, ensuring that the learned coefficients remain synchronized with the current reality of the local data-generating process in a way that a periodically updated universal model cannot.
\end{enumerate}

Thus, the shift to Methodology Transfer is not merely an operational preference for data privacy; it is a data-architectural necessity for overcoming Structural Uncertainty in dynamically heterogeneous environments.

\section{Modeling Methodology}
\label{Modeling Methodology}

Our framework establishes that the primary structure provides the potential for robustness by embedding a stable \(S^{(1)}\) signal within a noisy, high-D predictor-space. However, realizing this potential is contingent upon the modeling methodology employed. In the context of our analogy, simply possessing enriched heavy fuel is insufficient if one attempts to burn it in a finely-tuned engine incapable of processing the impurities.

This manuscript provides the theoretical grounding needed to systematically evaluate existing methods through this data-architectural lens -- identifying which qualify as the multi-fuel engines -- and to guide the development of novel architectures purpose-built to harness these principles.

As shown in Section~\ref{Core Theoretical Analysis}, the optimal prediction function for this architecture is inherently non-linear and requires interactions. Therefore, the choice of modeling methodology is critical. A model unable to capture these interactions cannot fully exploit the information in \(S^{\prime(2)}\) to reconstruct \(S^{(1)}\), and its performance will fall short of the theoretical ceiling. Furthermore, as formally demonstrated in Appendix~\ref{equivalence for LI viol}, this flexibility is also essential for a model to implicitly learn an expanded latent representation that can mitigate violations of critical structural assumptions, such as LI or the Baseline (Independent) Error Regime. Together, this provides a data-architectural justification for when highly flexible modeling methodologies may be superior in noisy, high-D settings that exhibit the assumed latent structure. Such methods must satisfy three related, central requirements:

\begin{enumerate}
\item
\textbf{Representation Learning:} The capacity to transform the input \(S^{\prime(2)}\) into a lower-dimensional representation that captures the \(S^{(1)}\) signal.
\item
\textbf{Interaction Modeling:} The ability to capture complex, non-linear dependencies required for inferring \(S^{(1)}\) configurations.
\item
\textbf{Regularization:} The mechanism to handle high-D and Informative Collinearity effectively.
\end{enumerate}

While no existing methods were explicitly designed with this framework's mechanisms as their guiding principles, we can categorize potential approaches based on how they engage with the latent \(S^{(1)}\) layer. The suitability of these approaches depends on trade-offs concerning performance, interpretability, and scalability.

\subsection{Explicit, Two-Stage Estimation (Representation Learning)}
\label{explicit, two-stage estimation}

This approach involves explicitly estimating the \(S^{(1)}\) states from \(S^{\prime(2)}\) in an initial, often unsupervised step, and then using these estimates as features to predict \(Y\).

\begin{itemize}
\item
\textbf{Autoencoders} \citep{goodfellow2016} are highly aligned with this goal. By forcing data through a compressed bottleneck layer, they learn a representation (analogous to \(S^{(1)}\)) that captures the essential, shared information while filtering idiosyncratic noise.
\item
\textbf{Factor Analytic Methods}, such as Exploratory Factor Analysis \citep{spearman1904}, explicitly model common variance (the signal from \(S^{(1)}\)) separately from unique variance (error), making them theoretically better suited for latent construct identification than Principal Component Analysis (PCA) \citep{pearson1901, jolliffe2002principal}.
\item
\textbf{Clustering Techniques} e.g., K-Means \citep{macqueen1967}, Gaussian Mixture Models \citep{dempster1977}: These are more appropriate when the signal from \(S^{(1)}\) is assumed to be categorical, representing distinct states, subtypes, or configurations. By partitioning the noisy, high-D space into cohesive groups, these methods effectively estimate the most probable latent state for each observation. The cluster centroid can serve as the stable signal from \(S^{(1)}\), while the distance of an individual observation from that centroid represents the idiosyncratic noise to be filtered out.
\item
\textbf{Latent Feature/Class Models}, such as the Indian Buffet Process (IBP) \citep{griffiths2011} or Mixed Membership Models \citep{airoldi2008}, are appropriate when \(S^{(1)}\) states are discrete and potentially non-exclusive, though they may face scalability challenges.
\end{itemize}

\subsection{Explicit, Simultaneous Estimation (Joint Modeling)}

This category includes unified models that simultaneously estimate the latent \(S^{(1)}\) layer and its relationship with \(Y\).

\begin{itemize}
\item
\textbf{Structural Equation Modeling (SEM)} is the archetypal method for confirmatory analysis, allowing precise specification of the hypothesized structure. However, traditional SEM faces significant computational and statistical challenges in high-D settings, limiting its practical utility for large-scale discovery.

It is worth noting, however, that this critique applies primarily to traditional SEM methodologies; the development of regularized SEM and other latent variable techniques tailored for high-D data is an active area of research \citep{jacobucci2016regularized} that may yield promising tools for our framework in the future.
\item
\textbf{Neural Networks (NNs)} \citep{goodfellow2016} are particularly promising. The architecture of a feed-forward NN mirrors the primary structure: the input layer \(S^{\prime(2)}\), hidden layers (representing \(S^{(1)}\)), and the output (\(Y\)). The hierarchical processing inherently learns latent features predictive of \(Y\).
\item
\textbf{Tabular Transformers and Attention Mechanisms:} Beyond standard feed-forward networks, modern attention-based architectures such as Tabular Transformers \citep{vaswani2017attention} offer a distinct theoretical alignment with G2G. The self-attention mechanism inherent to Transformers mathematically operationalizes \(S^{(1)}\) triangulation. By dynamically computing attention weights between predictors, these models can identify and amplify clumpy collinearity -- effectively learning to upweight informatively redundant features that reinforce the latent signal while downweighting idiosyncratic noise. Furthermore, attention maps provide a potential solution to the interpretability trade-off, offering a window into which predictor clusters the model is leveraging to resolve specific latent ambiguities \citep{arik2021tabnet}.
\end{itemize}

\subsection{Implicit Estimation}
\label{Implicit Estimation}

This category includes flexible models that do not explicitly define an \(S^{(1)}\) layer but implicitly leverage the latent structure through their capacity to model complex interactions. As discussed in Section~\ref{Core Theoretical Analysis}, an interaction term learns the effect of a combination of variables, effectively serving as an implicit proxy for a specific latent configuration.

\begin{itemize}
\item
\textbf{Ensemble Methods} such as Random Forests \citep{breiman2001} and Gradient Boosting \citep{friedman2001}, are adept at capturing high-order interactions. By aggregating diverse models, they can implicitly represent the latent structure and average out idiosyncratic errors \citep{dietterich2000}, leading to robust and scalable performance.
\item
\textbf{Flexible Direct Prediction Models} (those with many interactions, including both low- and high-order terms): This approach involves using models that, while still directly mapping the observed predictors \(S^{\prime(2)}\) to \(Y\), are explicitly designed to be flexible through the inclusion of a vast number of interaction terms. While these models lack an explicit architectural component for the \(S^{(1)}\) layer, their internal flexibility allows them to emergently learn the non-flat (hierarchical) function.
\end{itemize}

\begin{raggedright}
The distinction between Algebraic and Effective Signal Rank (Section~\ref{The Generative Origin of Low Rank: Effective Latent Complexity and Causal Consistency}) provides the theoretical mandate for these flexible approaches. While the Effective Signal Rank of the observed data is low (allowing for robust learnability), the optimal prediction function is inherently non-linear due to the Latent Complexity (\(K_{eff}\) discussed in Section~\ref{Primacy of the S(1) Layer and the Role of Interactions}). Simple linear models can only utilize the main effects, whereas high-capacity models are required to model the non-linear Softmax function necessary to distinguish between discrete configurations that may share similar linear profiles.
\end{raggedright}

\subsection{The Role of Regularization: Explicit and Implicit}

The framework advocates for using highly flexible, high-capacity models to capture the complex interactions needed to reconstruct \(S^{(1)}\). Simultaneously, it advocates for leveraging high-D, collinear, error-prone predictor-spaces. This combination -- a high-capacity model and high-D data -- is the classic recipe for overfitting. Without any constraints, the model could easily use its flexibility to memorize the idiosyncratic noise (which, as detailed in Section~\ref{Benign Overfitting Link: Towards a Unified Understanding of Robustness}, comprises both Observational Error and Structural Uncertainty) in \(S^{\prime(2)}\) rather than learn the stable, underlying \(S^{(1)}\) signal.

This risk becomes especially acute in the scenarios where predictor-space dimensionality exceeds the available sample size (\(m > N\)). Therefore, to successfully operationalize this framework and ensure the model extracts the robust signal, some form of regularization becomes practically essential. This constraint is necessary to guide the optimization process, biasing the model toward the stable, low-rank \(S^{(1)}\) structure and preventing it from fitting to noise. Regularization can be applied in two primary forms.

\begin{enumerate}
\item
\textbf{Explicit Regularization:} Leveraging high-D, informatively collinear data necessitates methodologies that can handle dependencies between predictors -- the type of regularization matters.
\begin{itemize}
    \item 
    \(L_{2}\) regularization (e.g., in neural networks) is well-suited to our framework, as it encourages the distribution of weights across correlated features \citep{hoerl1970}. This directly supports the principle of leveraging Informative Collinearity, where many predictors provide redundant signals.
    \item 
    In contrast, \(L_{1}\) regularization (which encourages parsimonious solutions) would be counterproductive as it would encourage the selection of one predictor from a correlated group while zeroing out the others \citep{tibshirani1996}, thereby destroying the redundancy this framework seeks to exploit.
    \item 
    Group Lasso \citep{yuan2006model} offers a theoretically superior alternative to standard \(L_{1}\). Although Group Lasso is a supervised technique (optimizing prediction of \(Y\)), it uniquely leverages the unsupervised structure of the data. By defining groups based on the correlational clusters identified in the Representation Learning phase, Group Lasso forces the model to perform selection at the cluster level \citep{meier2008group}. This preserves the redundant coverage required for reliability within a selected latent cluster, while still zeroing out entirely irrelevant groups. This effectively bridges the gap between unsupervised structure discovery and supervised prediction, aligning well with Outcome-Specific P-DCAI principles.
\end{itemize}
\item 
\textbf{Implicit Regularization:} Alternatively, regularization can be achieved implicitly, arising from the modeling methodology itself rather than an explicit penalty term. This can take several forms, all of which align well with this framework:
\begin{itemize}
    \item
    Ensemble-Based Regularization: As discussed in Section~\ref{Implicit Estimation}, ensemble methods like Random Forests aggregate diverse models. This process implicitly averages out idiosyncratic errors, acting as a regularizer that favors the stable, shared signal over predictor-space noise.
    \item
    Architectural Regularization: Methods described in Section~\ref{explicit, two-stage estimation}, such as Autoencoders, impose implicit regularization through their very design. By forcing data through a compressed bottleneck layer, they are architecturally biased to find a low-dimensional representation (analogous to \(S^{(1)}\)) and filter out idiosyncratic, high-D noise.
    \item
    Optimizer-Based Regularization: Beyond model-based regularization, here we also consider the interaction between the data architecture and the implicit bias of the training algorithm. In Section~\ref{Benign Overfitting Link: Towards a Unified Understanding of Robustness} we show that the primary structure generates the low-rank-plus-diagonal characteristic linked to BO. This specific data architecture may create a ``friendlier'' optimization landscape for highly flexible models like neural networks.

    It has been hypothesized that the implicit bias of algorithms like Stochastic Gradient Descent (SGD) \citep{robbins1951stochastic} inherently favors solutions that align with this low-rank signal structure. For example, in related problems like matrix factorization, gradient descent is known to implicitly regularize for low-rank solutions \citep{gunasekar2017implicit}. This suggests a powerful synthesis: the data architecture itself may activate or enhance the algorithm's implicit regularization, guiding the model toward a robust solution that successfully recovers the \(S^{(1)}\) states. This aligns with our discussion of Representation Learning in Section~\ref{The Objective-Capability Gap and Representation Learning}. Exploring this synergy -- linking the data, model, and optimization algorithm -- is a key direction for future research.    
\end{itemize}
\end{enumerate}

\subsection{Architecting for Absorption}

To realize the Absorption Hypothesis -- where a model mitigates systematic error or LI violations by isolating them as pseudo-states -- specific architectural choices are required to guide the optimization:

\begin{itemize}
    \item
    \textbf{Bottleneck Constraints:} When using Autoencoders, the width of the bottleneck layer functions as a hyperparameter for the effective rank (\(K_{eff}\)). The bottleneck must be sufficiently wide to accommodate both the signal from the \(S^{(1)}\) layer and the necessary error pseudo-states (\(Z\)), but narrow enough to force the compression of idiosyncratic noise \citep{vincent2008extracting}. Tuning this width is effectively tuning the model's capacity to absorb structural violations.
    \item
    \textbf{Metadata Injection:} To operationalize the Metadata-Adversarial strategy, models should be architected with dual input channels. The primary channel accepts the predictor set, while a secondary channel accepts measurement metadata that is expected to ideally have no effect on the outcome. By concatenating this metadata with the latent representation layer, the model is explicitly incentivized to offload variance associated with the metadata into distinct nodes, thereby forcing the separation of systematic error drivers from the true latent drivers.
\end{itemize}

\section{Simulation Study}
\label{Sim}

To empirically demonstrate core G2G and P-DCAI principles, we conducted a comprehensive simulation study utilizing R. The study was designed to stress-test the architectural theory in an environment characterized by Structural Uncertainty, Predictor Error, and non-obvious deception -- a context where variables containing signal and those containing only pure noise are statistically indistinguishable using univariate metrics. The complete, fully-annotated script is available on GitHub (\href{https://github.com/tjleestjohn/from-garbage-to-gold}{https://github.com/tjleestjohn/from-garbage-to-gold}), allowing for full reproducibility.

\subsection{Simulation Design:}

The simulation generates training and test sets consisting of \(12,000\) and \(3,000\) rows, respectively. Data from both sets adhere to the same primary structure where \(k = 4\) latent states generate a pool of \(m = 4,000\) potential predictors (the haystack), and all generative parameters are identical (within an iteration).

While \(k = 4\) is selected to ensure computational reproducibility, the theoretical mechanics of the framework are governed by the ratio of predictor dimensionality to latent complexity (\(m \gg k\) or \(m \gg K_{eff}\)); our specification of \(4,000\) predictors creates a substantial \(1000:1\) ratio of \(m\) to \(k\) (or at worst a \(250:1\) ratio of \(m\) to \(K_{rlzd}\)) that demonstrates the asymptotic properties of the high-D context. We note that results presented below are consistent when varying the number of latent states. As \(k\) increases, however, in order to achieve comparable results, \(N\) and \(m\) must also be increased. Consequently, computer RAM requirements and runtime may limit unconstrained increases in \(k\), \(N\), and \(m\). Nevertheless, we invite users to download the script and modify these parameters to examine how they interact to affect results.

For all generated predictor variables, we bound the parameters that define the observation process (\(\alpha\) and \(\beta\)) away from perfect fidelity. Specifically, Observational Error is applied such that the probability of a correct observation ranges (uniformly) between \(0.875\) and \(0.925\) across all variables. Though variation exists, this ensures that predictive signal is universally degraded by a non-trivial error rate (\(10\%\) on average), preventing perfect observational fidelity in any variable. Additionally, \(50\%\) of the generated predictors contain pure noise prior to applying Observational Error (i.e., are statistically unrelated to both other predictors and the outcome, even in the absence of Predictor Error).

To demonstrate the theoretical distinctions made in Section~\ref{Causal Consistency and Parametric Constraints}, the simulation evaluates two data generative modes for the predictors that carry signal. In both modes, the \textit{medium} latent state correlation setting from the Part 1a Latent Effective Complexity analysis is utilized.

\begin{enumerate}
\item
\textbf{Scenario A: Causal Consistency (The Smooth Manifold):} The probabilities generating \(S^{(2)}\) variables and \(Y\) are determined by a parametric design matrix of the latent states (\(S^{(1)}\)), including main effects and interactions.

Suppressed interaction scaling is used to produce relatively simple variables -- those where stable directional forces (originating from the \(k\) latent states) dominate so that variables tend to act as consistent proxies for latent states directly.

\item
\textbf{Scenario B: Chaotic Generative Variables (The Fractured Surface):} All probabilities generating \(S^{(2)}\) variables and \(Y\) are independent draws for every realized configuration of \(S^{(1)}\). This completely destroys topological structure, forcing predictors (and the outcome) to act as unique identifiers for specific latent configurations (not the latent states directly).
\end{enumerate}

\subsubsection{Generative Mechanics: Universal Gaussian-Copula Mapping and Non-Obvious Deception}

To ensure that predictive robustness arises from the architecture of the predictor portfolio rather than a few obvious signals, we designed a predictor-space with universally low to moderate Structural Strength and non-obvious deception. In short, for predictors driven by true latent signals, the design limits the strength of the bivariate statistical correlation with latent drivers. Additionally, the implementation of the copula mapping forces all variables -- whether driven by true latent signals or generated as pure noise -- to share identical marginal distributions.

\begin{itemize}
\item
\textbf{The Baseline Distribution:} All raw scores for variables (Causally Consistent, Chaotic Generative, or pure noise) are initially drawn from Normal Distributions centered at 0.
\item
\textbf{Scenario A (Causal Consistency):} Main effects are drawn from the Standard Normal Distribution. Interaction effects are drawn from a Normal Distribution with a standard deviation that strictly decays from \(1\) as the interaction order increases. For predictors, these continuous raw scores are then mapped to a probability range of \([0.25, 0.75]\) via a copula transformation. For the outcome (\(Y\)), continuous raw scores are copula-mapped to a wider range of \([0, 1]\).
\item
\textbf{Scenario B (Chaotic Generative):} Independent random draws from the Standard Normal Distribution are made for every latent configuration for both predictors and the outcome. Predictors are then copula-mapped to \([0.25, 0.75]\), and the outcome to \([0, 1]\).
\item
\textbf{Pure Noise Predictors:} To exactly match the cardinality of the \(2^k\) conditional probabilities generating the signal variables, exactly \(2^k\) random draws are made from the Standard Normal Distribution, but crucially, these values are randomly assigned to each row (breaking any structural tie to specific configurations). These are also copula-mapped to \([0.25, 0.75]\) to perfectly camouflage them among the predictors carrying signal.
\end{itemize}

\subsection{Part 1a: Latent Effective Complexity}

\begin{figure}
\centering
\includegraphics[width=.90\textwidth,keepaspectratio]{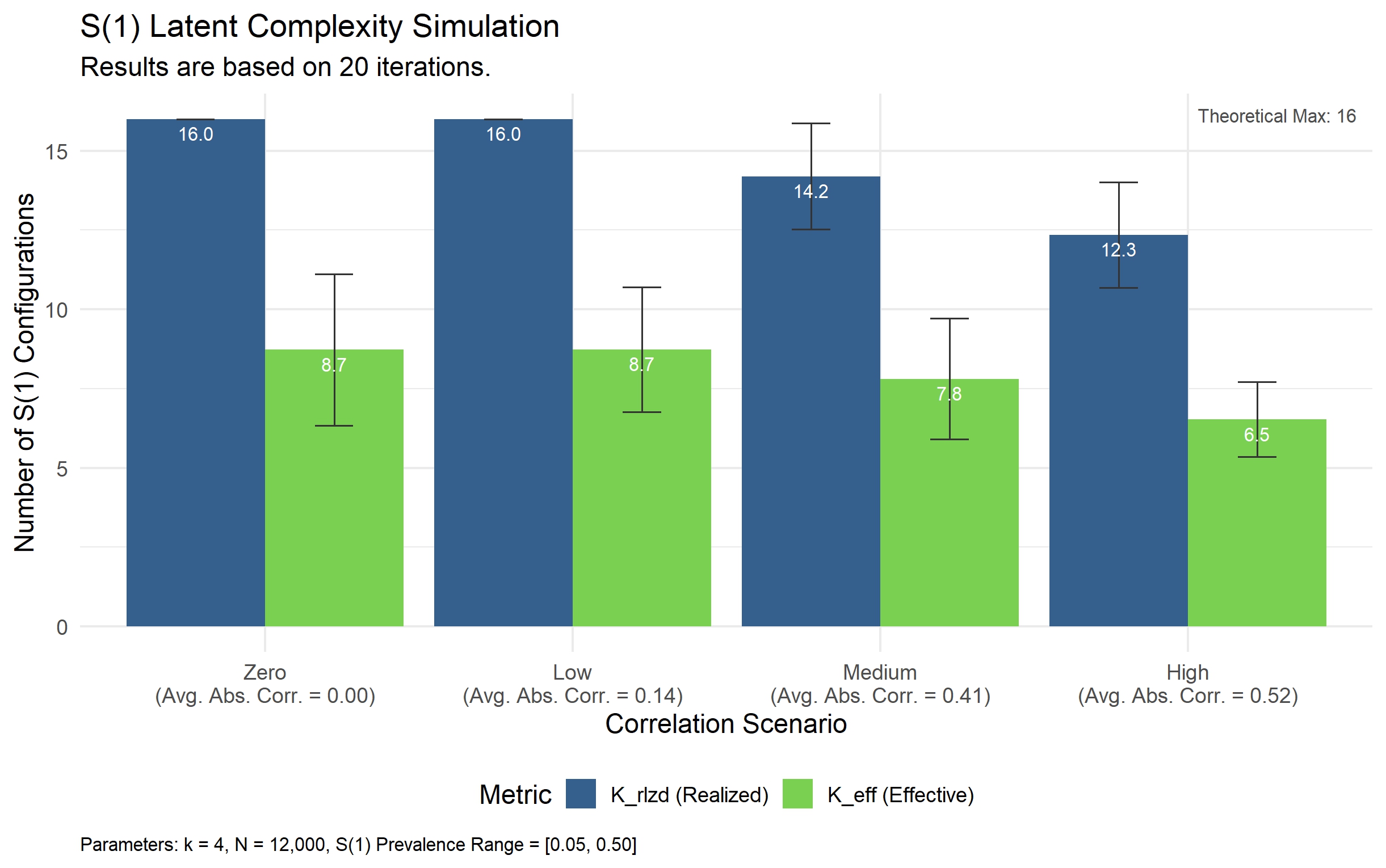}
\caption{Latent Complexity Simulation.}
\label{fig: Effective Latent Complexity Figure}
\end{figure}

Part 1a of the simulation validates the concept of Effective Latent Complexity (\(K_{eff}\)). We simulated \(20\) iterations of \(S^{(1)}\) generation (\(k = 4\)) under varying prevalence rates (ranging uniformly from \(.05\) to \(.50\)) and correlation structures (none, low, medium, high). For each, we calculated the number of realized configurations (\(K_{rlzd}\)) and the effective number of configurations (\(K_{eff} = 2^{H(S^{(1)})}\)).

\vspace{0.5\baselineskip}
\noindent
\textbf{Results and Interpretation:} The simulation confirmed a significant gap between the theoretical maximum complexity (\(2^4 = 16\)) and the observed \(K_{eff}\) (see Fig.~\ref{fig: Effective Latent Complexity Figure}). As expected given Appendix~\ref{latent sparsity}, the introduction of correlation and skewed prevalence among the latent states naturally compresses the entropy of the system, reducing the effective complexity of the latent space even before observation. Importantly, because this finding reflects the latent space, it is neither affected by the Observational Error process nor the types of observed predictors that are ultimately generated (Causally Consistent vs. Chaotic Generative).

\subsection{Part 1b: Spectral Validation and Heuristic Assessment}

\begin{figure}
\centering
\includegraphics[width=.90\textwidth,keepaspectratio]{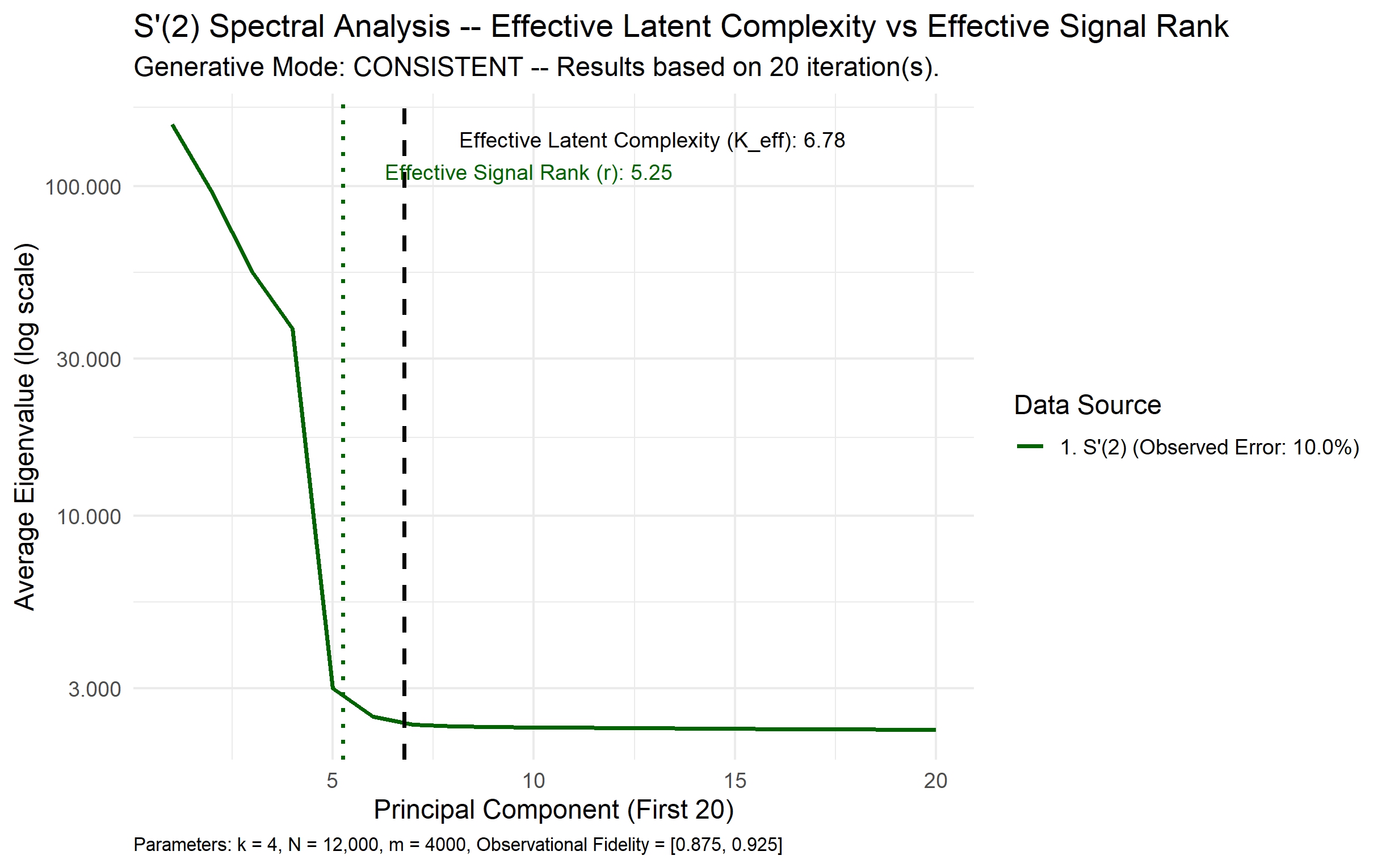}
\caption{\(S^{\prime(2)}\) Spectral Analysis: Causally Consistent Variables.}
\label{fig: Spectral Analysis: Consistent}
\end{figure}

\begin{figure}
\centering
\includegraphics[width=.90\textwidth,keepaspectratio]{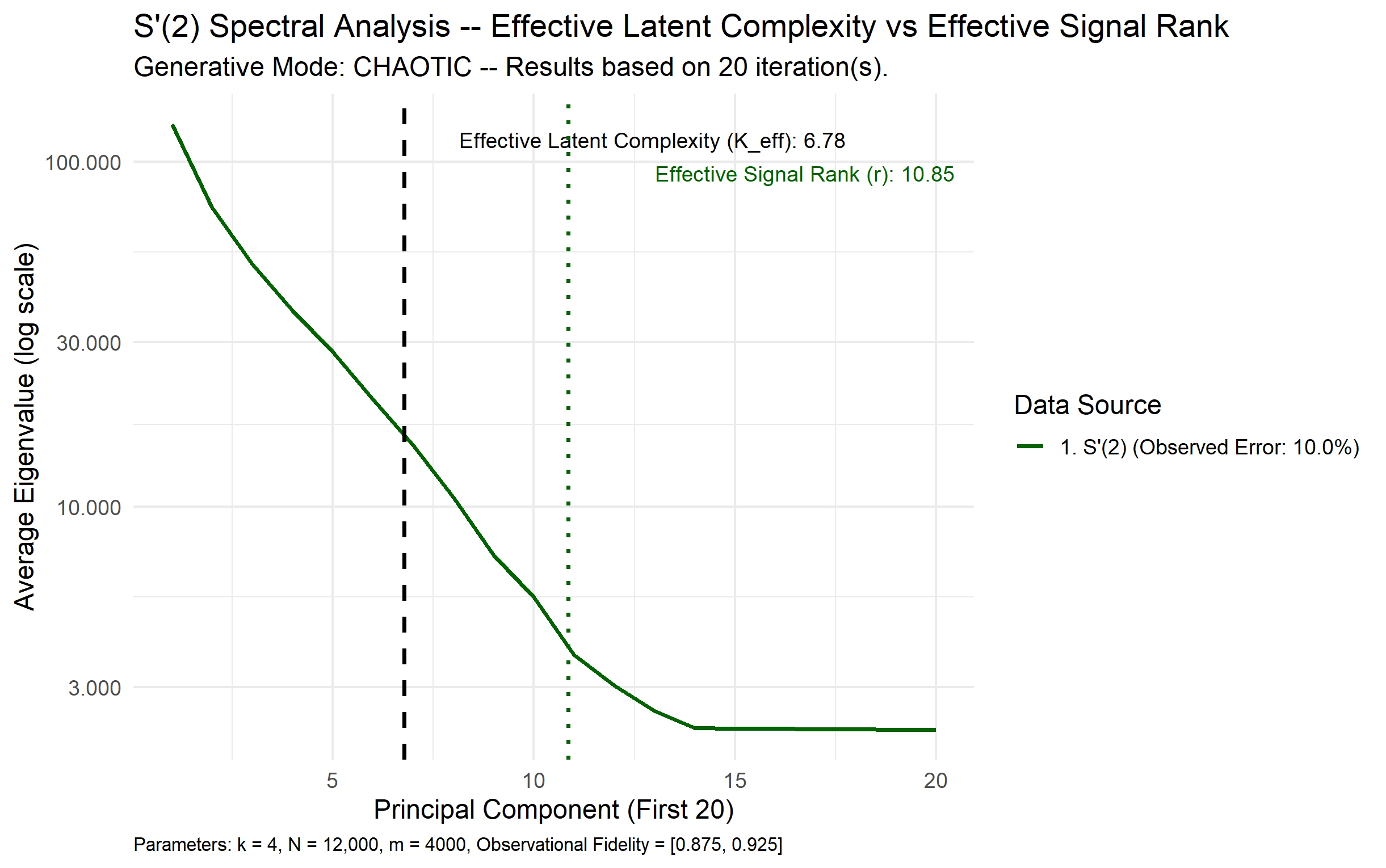}
\caption{\(S^{\prime(2)}\) Spectral Analysis: Chaotic Generative Variables.}
\label{fig: Spectral Analysis: Chaotic}
\end{figure}

This component empirically validates the utility of the spectral heuristic (Section~\ref{Assessing Framework Applicability}) and the connection to BO (Section~\ref{Benign Overfitting Link: Towards a Unified Understanding of Robustness}) by confirming the low-rank-plus-diagonal covariance structure in the presence of both universal Predictor Error and pure noise with non-obvious deception. We simulated \(20\) iterations of the generative process for the predictor-space, calculating the eigenvalues and scree plot for each simulation. Figures~\ref{fig: Spectral Analysis: Consistent} and \ref{fig: Spectral Analysis: Chaotic} show the average across the \(20\) iterations for Causally Consistent and Chaotic Generative modes respectively.

\vspace{0.5\baselineskip}
\noindent
\textbf{Results and Interpretation:} Despite the fact that all variables are corrupted by Predictor Error, Structural Strength is bounded to low to moderate levels, and the signal and pure noise predictors share comparable univariate distributions, both scree plots reveal distinct elbows. The location of the elbows tracks with the underlying drivers -- between \(k\) and \(K_{eff}\) in Causally Consistent mode and between \(K_{eff}\) and \(K_{rlzd}\) in Chaotic mode (see \(E(K_{rlzd}) = 14.2\) from Fig.~\ref{fig: Effective Latent Complexity Figure}).

This demonstrates that the spectral geometry of the observed data is driven by the covariance of the shared drivers in either mode, confirming that the latent drivers can create a detectable signal subspace, even when observed variables are universally error-prone, have low to moderate Structural Strength, and are also camouflaged (due to non-obvious deception). However, we note that what precisely constitutes a latent ``driver'' varies across generative modes (latent states vs. latent state configurations) as discussed in Section~\ref{Distinguishing Algebraic vs. Effective Signal Rank}.

\subsection{Part 2: Breadth vs. Benchmark Performance}

The core of the study evaluates the ability of high-D Breadth to reduce predictive uncertainty. Within each iteration (\(20\) are performed in total), a new, full universe of variables is generated. The training set is then used to implement the feature selection strategies outlined below. Final models are then fit at increasing levels of dimensionality, and performance is calculated using the test set.

The primary performance metric is Conditional Entropy (\(H(Y|S^{\prime(2)})\)), quantifying the raw information gain: specifically, how close the model's posterior probability distribution comes to the theoretical limit of certainty imposed by the latent structure itself. For broad familiarity, the AUROC and AUPRC graphs are provided in Appendix~\ref{Additional Simulation Graphs: Breadth vs. Depth as AUROC and AUPRC}.

\vspace{0.5\baselineskip}
\begin{raggedright}
\textbf{Modeling Engine:} For the final models, to estimate posterior probabilities in a manner that did not require specialized hardware (e.g., GPUs), we utilized a Random Forest classifier (via the \texttt{ranger} package in R). The implementation used the ``gini'' splitrule, an \texttt{mtry} of \(\sqrt{m}\) for a given set size, a \texttt{min.node.size} of \(10\), a \texttt{max.depth} setting of \(0\) (unlimited depth allowable), and estimated \(1,250\) trees per model.
\end{raggedright}

As theoretically derived in Section~\ref{Primacy of the S(1) Layer and the Role of Interactions}, optimally exploiting a latent hierarchical structure requires a model capable of capturing high-order non-linear interactions. The Random Forest serves as a light-weight, high-capacity learner, working towards this ideal. However, we note that this ideal would be better served by an even higher capacity model (e.g., a deep neural network). The Random Forest was primarily chosen for its minimum sufficiency in demonstrating the core G2G mechanism using the generated data without requiring specialized hardware and/or exceedingly long compute times.

The same Random Forest implementation was also leveraged for the proxy model used in the Targeted Residual Expansion feature selection strategy. However, for this lighter-weight model, only \(125\) trees were estimated per model.

\subsubsection{The Traditional Benchmark}

\begin{itemize}
\item
\textbf{Idealized Depth:} This represents the theoretical limit of the conventional clean parsimony strategy. It utilizes a low-D set of predictors with zero Observational Error. Features were selected using MRMR run on perfectly clean data to align with the traditional aim of selecting features that are highly predictive of the outcome but minimally related to one another. The budget consists of roughly \(2 \times K_{eff}\) features (calculated from the rounded Effective Latent Complexity), aligning with the context where a domain expert selects two features per driver to attempt to ensure reliable, complete latent coverage while remaining in a low-D predictor-space. It is the only strategy in the simulation that is traditional in all senses: it relies on perfect knowledge of the latent structure (idealized domain expertise),  perfectly clean data, and a low-D predictor-space obtained by penalizing collinearity.
\end{itemize}

\subsubsection{Breadth Strategies (Error-Prone)}

The following strategies select features from the full haystack, including predictors carrying true signal and those consisting of pure noise (a \(50/50\) split of the haystack). Dimensionality of the feature set is expanded using the various selection strategies until the pre-determined budget (\(37.5\%\) of the haystack) is met.

\begin{enumerate}
\item
\textbf{Random (Baseline):} This implementation selects features from the haystack at random. This serves as a worst-case benchmark for Breadth strategies, indicating no knowledge of either \(Y\) or the internal structure of \(S^{(2)}\).
\item
\textbf{MRMR (Penalized Redundancy):} This implementation of MRMR leverages high-D Breadth -- the predictor set expands to the same high-D levels as the other Breadth strategies (a nontraditional stance). However, this supervised strategy does adhere to the traditional philosophy of penalizing collinearity -- for a given \(m\), it actively seeks to create a set of predictors that maximize information about \(Y\) and minimize internal redundancy. Crucially, while MRMR actively penalizes internal redundancy, forcing it to fulfill a massive, high-D budget from a finite haystack fundamentally alters its behavior. Once highly orthogonal signal variables are exhausted, MRMR is mathematically forced into a trap where it must choose between pure noise variables (yielding near-zero relevance and zero redundancy) and informatively redundant variables (yielding both positive relevance and redundancy penalties).
\item
\textbf{Supervised G2G (Targeted Residual Expansion):} An outcome-specific P-DCAI approach that explicitly targets Predictive Efficiency as described in Section~\ref{Automated Feature Selection}. It iteratively expands the set by selecting features that maximize correlation with the residuals of the current ensemble proxy model. By targeting the predictive weakest link, it actively seeks features that resolve specific latent ambiguities, implicitly valuing collinearity when doing so clarifies the predictive signal.
\item
\textbf{Unsupervised G2G (Spectral-Anchored):}  A general-purpose P-DCAI approach operating without access to \(Y\), this strategy implements a two-stage logic: first, it identifies variables to serve as ``Elite Core'' prototypes of latent drivers based on spectral embedding (Promax-rotated loadings) to ensure complete coverage of the latent space. Second, it employs an \textbf{Adaptive Weakest Link} selection mechanism. This mechanism iteratively calculates the centroid of selected features for each latent factor, then probabilistically samples new features that reinforce the factor with the highest reconstruction error, explicitly operationalizing the goal of maximizing reliability via redundancy targeting the weakest link.
\end{enumerate}

\begin{figure}
\centering
\includegraphics[width=.90\textwidth,keepaspectratio]{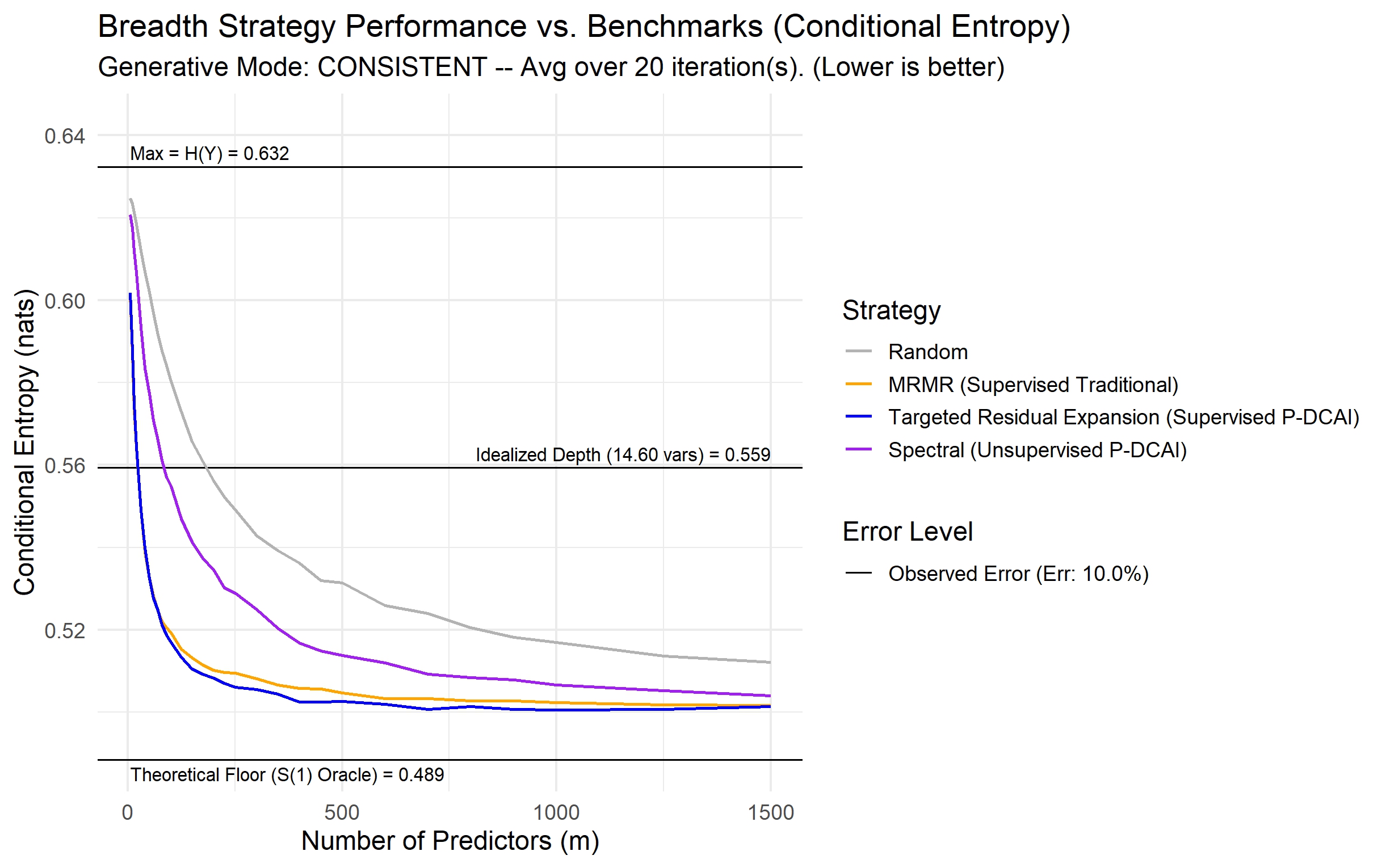}
\caption{Breadth vs. Depth: Causally Consistent Variables.}
\label{fig: Breadth vs. Depth: Consistent}
\end{figure}

\begin{figure}
\centering
\includegraphics[width=.90\textwidth,keepaspectratio]{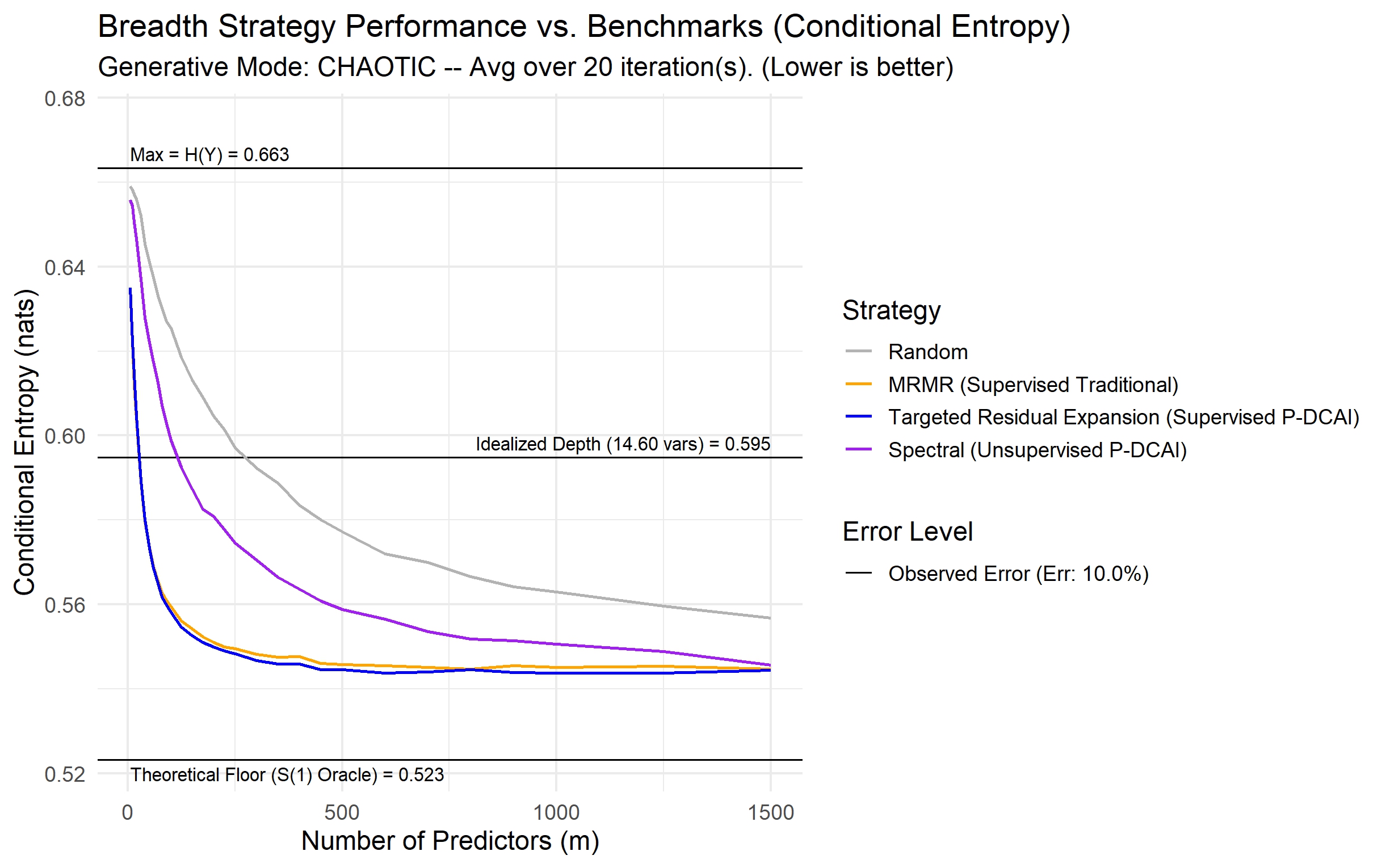}
\caption{Breadth vs. Depth: Chaotic Generative Variables.}
\label{fig: Breadth vs. Depth: Chaotic}
\end{figure}

\subsubsection{Results and Interpretation}

\textbf{Primary Comparison:}

\begin{itemize}
\item
\textbf{Dirty Breadth vs. Clean Parsimony:} In both modes, the Idealized Depth strategy hits a substantive Performance Floor -- it cannot overcome the Structural Uncertainty inherent to the small predictor set, demonstrating that the targeted cleaning of a fixed set of predictors can be insufficient in complex, hierarchical generative contexts. 

In contrast, in both modes, all Breadth strategies -- including Random selection -- achieve predictive performance (as \(m\) increases) that surpasses the Idealized Depth benchmark, demonstrating that high-D Breadth can overcome both Predictor Error and the Structural Uncertainty inherent in the smaller set.
\end{itemize}

\vspace{0.5\baselineskip}
\begin{raggedright}
\textbf{Secondary Comparison -- Efficiency and Strategy:} The simulation revealed distinct efficiency profiles as the predictor count increased, with remarkably consistent overarching trends across both generative modes.
\end{raggedright}

\begin{itemize}
\item
\textbf{Startup Efficiency:} In both modes, the Targeted Residual Expansion and MRMR strategies exhibit high initial efficiency. By utilizing their access to the outcome, they rapidly reduce entropy in the early stages of feature set expansion. The Spectral-Anchored and Random strategies also improve, but their initial rates are less steep, demonstrating the expected ``startup advantage'' of supervised methods in this generative context.
\item
\textbf{The Pure Noise Trap (Targeted Residual Expansion vs. MRMR):} Crucially, in both modes, a slight performance gap favoring Targeted Residual Expansion over MRMR begins to emerge at about an eighth of the budget. This small but persistent gap (the relative flattening of the MRMR curve) is more pronounced when viewing the AUROC and AUPRC trajectories in Appendix~\ref{Additional Simulation Graphs: Breadth vs. Depth as AUROC and AUPRC}. This divergence provides a direct empirical demonstration of the mathematical trap inherent to MRMR's scoring mechanism in high-D spaces. As the selected feature set expands and highly orthogonal signals are exhausted, the redundancy penalty for the remaining informatively collinear features becomes relatively large. When this penalty outweighs a feature's relevance, MRMR is mathematically forced to begin selecting pure noise features (which carry a net score of zero due to near-zero correlation with both the outcome and the selected set). This explains why MRMR's efficiency trajectory flattens earlier on. In contrast, Targeted Residual Expansion does not penalize collinearity, allowing for Informative Redundancy to be selected when doing so helps resolve the weakest predictive link. The small gap due to the redundancy penalty MRMR imposes is also a direct consequence of the specific generation parameters of the simulated data. In other generative contexts where high degrees of Informative Collinearity are more or less prevalent than in this simulation, we can expect the efficiency profiles of Targeted Residual Expansion and MRMR strategies to diverge or converge, respectively.
\item
\textbf{Asymptotic Convergence (Spectral-Anchored vs. Supervised):} Finally, in both modes, a convergence occurs as \(m\) continues to increase toward the budget limit. While Spectral-Anchored has a slower start, its performance eventually approaches the supervised strategies. By focusing solely on reconstructing the \(S^{(1)}\) layer via spectral prototypes and reinforcing weakest links, the Spectral-Anchored strategy built a dataset (within the budget) nearly as capable of robust prediction as the supervised methods that ``saw'' the answer. Though limited to this simulation's generative context, this provides strong empirical support for P-DCAI's core premise: structurally reconstructing \(S^{(1)}\) is a highly viable, robust proxy for predicting \(Y\) when labels are unavailable, yet to be determined, or error-prone (rendering supervised approaches problematic).
\end{itemize}

\vspace{0.5\baselineskip}
\begin{raggedright}
\textbf{Asymptotic Performance Gap}
\end{raggedright}

\begin{itemize}
\item We note a persistent gap between the final performance of the Breadth strategies and the theoretical minimum conditional entropy. This gap is an expected consequence of the simulation's design and practical computational limits, driven by two interacting factors:

First, the haystack dimensionality is capped at \(m = 4,000\). While large relative to the latent complexity (\(k\) and \(K_{rlzd}\)), the deliberately low to moderate Structural Strength and stochastic generative mechanisms ensure that this finite haystack still contains a substantial degree of Structural Uncertainty. Consequently, the theoretical ceiling of infinite breadth cannot be perfectly reached with this constrained set. Effectively, because the haystack is finite (\(m = 4,000\)) and Structural Strength is deliberately bounded between \([0.25, 0.75]\), the available predictor set retains a non-trivial degree of unresolvable Structural Uncertainty.

Second, calculating the true conditional entropy for a high-D predictor set is computationally impractical; therefore, we must estimate it using the modeled score. These entropy estimates are bounded by the capacity of the specific Random Forest implementation to uncover the optimal prediction function -- a task that becomes increasingly difficult as \(m\) increases. We invite users to modify this parameter in the simulation script (e.g., by reducing or increasing the number of estimated trees) to examine how changes to model capacity are associated with the magnitude of the asymptotic gap.

Ultimately, this result reinforces our argument that fully exploiting the data's architectural potential is strictly contingent on the model's capacity to do so. This finding also underscores the practical advantage of P-DCAI: by achieving target robustness levels with significantly fewer predictors than MRMR or Random selection strategies, P-DCAI reduces the dimensional burden on the model. This efficiency limits the required model capacity, making it a critical strategy for constrained real-world applications.
\end{itemize}

\subsection{Part 3: Comparative Strategic Spectral Analysis}

We calculated the Spectral SNR (\(SNR_{spectral}\)) at the elbow (determined using the full haystack) for the selected subsets to examine how different feature selection strategies affect the identifiability of the signal subspace. The comparative spectral analysis provides the mechanistic explanation for the performance results. 

\vspace{0.5\baselineskip}
\begin{raggedright}
\textbf{Results and Interpretation}
\end{raggedright}

\begin{itemize}
\item
In both modes, all deliberate feature selection strategies outperformed Random selection on \(SNR_{spectral}\). Additionally, all deliberate strategies resulted in larger signal eigenvalues than their Random selection counterparts.

Importantly, the Spectral-Anchored strategy produced subsets with the highest \(SNR_{spectral}\) and highest signal eigenvalues generally, reflecting that this strategy tended to select features that allow for comprehensive reconstruction of the latent space.

The large \(SNR_{spectral}\) advantage of the Spectral-Anchored strategy in Causally Consistent mode is expected -- our operationalization of near-linear Causal Consistency is highly aligned with traditional factor analytic treatment of latent factors, which also aligns with the simulation's Spectral-Anchored strategy. Thus, the Spectral-Anchored strategy is uniquely positioned in this simulation to extract and recover the latent states when variables are generated in the Causally Consistent mode. The high performance of Spectral-Anchored in Chaotic mode, however, is somewhat less expected -- its performance indicates that despite latent drivers not being aligned with traditional factor methods in this generative mode, the traditional spectral tools still were able to uncover and exploit the latent structure. This positive result, despite the methodological mismatch, suggests that the development of more general latent factor representation methods may achieve even higher spectral performance in contexts where the underlying factor structure may not be well-aligned with traditional factor analytic assumptions. Such development (discussed in Section~\ref{Limitations and a Future Research Agenda}) is a clear direction for future research.
\item 
Synthesizing Efficiency and Robustness: Ultimately, this spectral analysis concretizes the distinction between general-purpose and outcome-specific P-DCAI strategies. Unsupervised approaches (like Spectral-Anchored) optimize for a large, balanced Spectral SNR, effectively building a robust, comprehensive representation of all drivers of the predictor-space originating from the \(S^{(1)}\) layer. This is geometrically ideal for mitigating Outcome Error (via Benign Overfitting). Conversely, supervised approaches slightly sacrifice global spectral balance to maximize Predictive Efficiency, demonstrating that a sub-optimal global covariance structure is perfectly acceptable -- and indeed highly efficient -- when the sole objective is predicting a high-fidelity outcome.
\end{itemize}

\begin{figure}
\centering
\includegraphics[width=.90\textwidth,keepaspectratio]{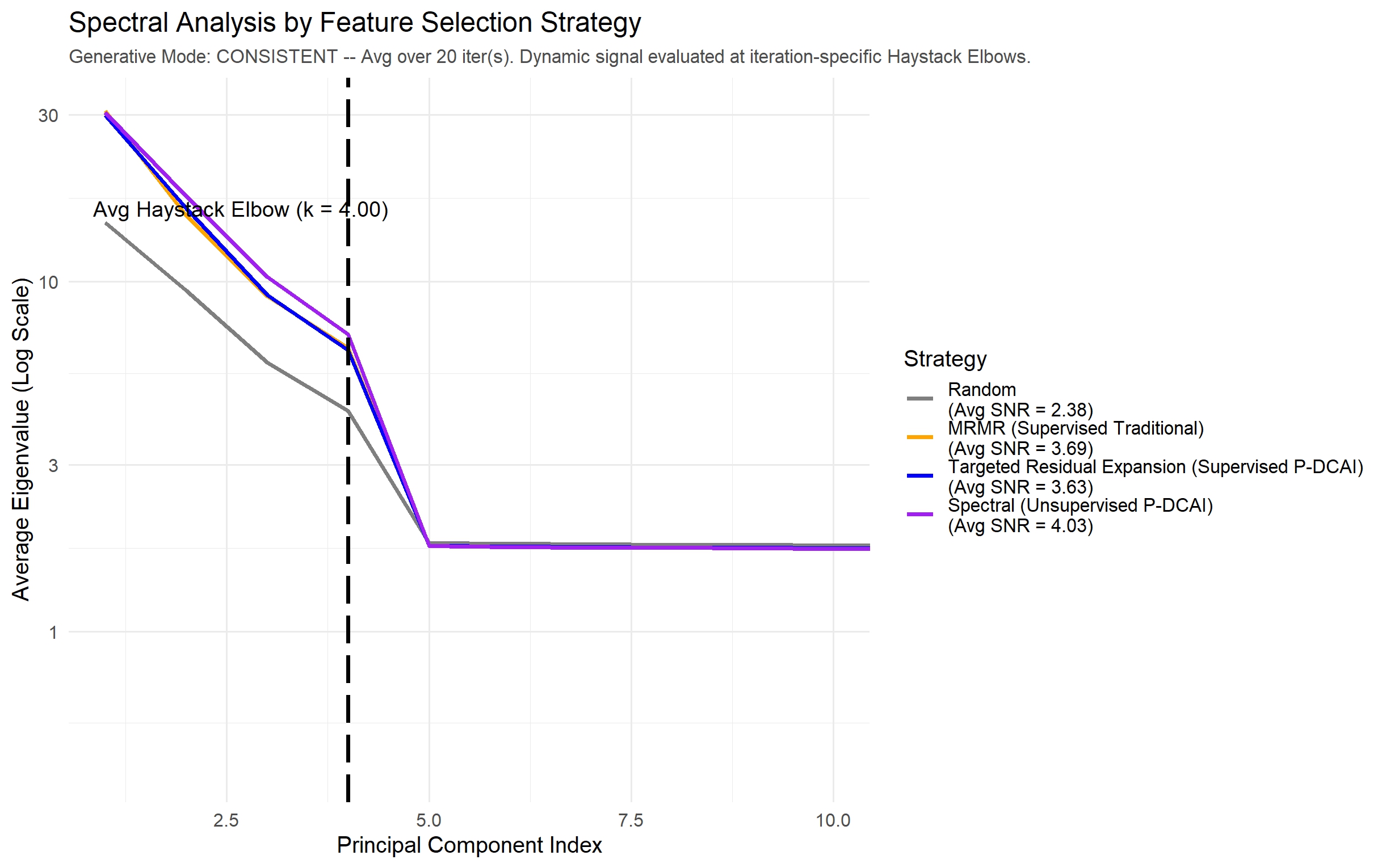}
\caption{Comparative Spectral Analysis: Causally Consistent Variables.}
\label{fig: Comp Spec: Consistent}
\end{figure}

\begin{figure}
\centering
\includegraphics[width=.90\textwidth,keepaspectratio]{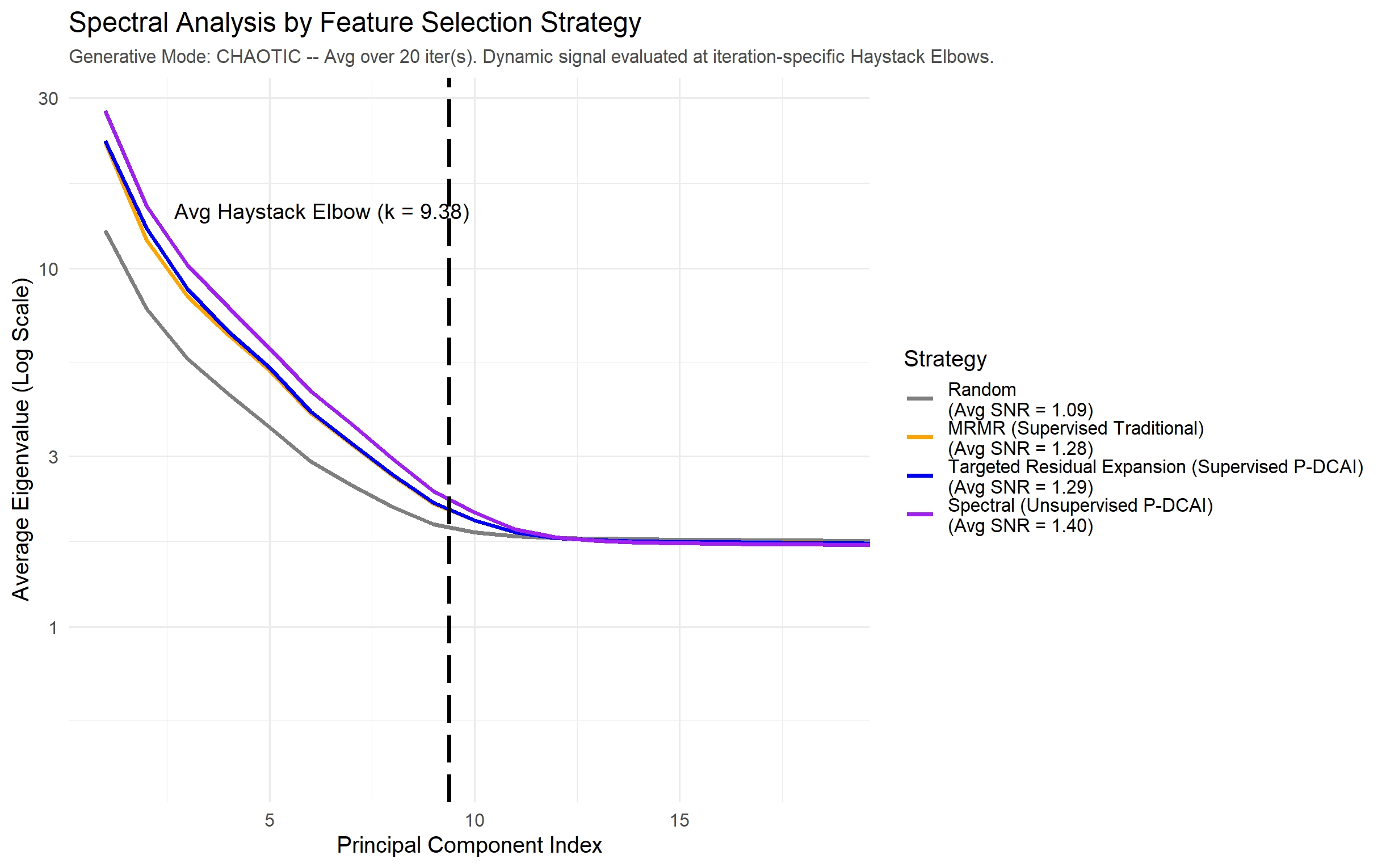}
\caption{Comparative Spectral Analysis: Chaotic Generative Variables.}
\label{fig: Comp Spec: Chaotic}
\end{figure}

\section{Motivating Case Study: Cleveland Clinic Abu Dhabi}
\label{CCAD Motivating Example}

As noted in the Section~\ref{Intro}, the anomaly of the Cleveland Clinic Abu Dhabi (CCAD) case study result \citep{leestjohn2024} was a core motivator for this theoretical development. We can now re-examine that study through the lens of this framework to understand exactly why it succeeded where traditional methods failed.

\subsection{Mapping Clinical Data to G2G}
G2G is highly applicable to prediction tasks using complex electronic health record (EHR) data, such as forecasting the onset of a complex disease like heart failure within a specified period (the binary outcome \(Y\): Develops Heart Failure Yes/No). The true, underlying health status of a patient, which dictates this risk, can be conceptualized as a set of latent, binary first-stage states (\(S^{(1)}\)). For instance, \(X_{1}^{(1)}\) might represent the True Presence of Clinically Significant Underlying Cardiac Dysfunction (Yes/No), while \(X_{2}^{(1)}\) could signify a True Systemic Pro-Inflammatory State (Yes/No).

The unobserved pathophysiological states (\(S^{(1)}\)) then manifest as various true, error-free binary clinical conditions or biomarker threshold states (\(S^{(2)}\)). Examples include \(X_{1}^{(2)}\): True Ejection Fraction is critically low (e.g., below 35\%) (Yes/No), \(X_{2}^{(2)}\): True NT-proBNP level is severely elevated (e.g., above 2000 pg/mL) (Yes/No), or \(X_{3}^{(2)}\): True high-sensitivity C-Reactive Protein (hsCRP) is pathologically high (e.g., above 3 mg/L) (Yes/No).

In practice, clinicians and researchers work with observed EHR data (\(S^{\prime(2)}\)), which are imperfect, often binarized (e.g., against a threshold or reference range) representations of these second-stage states. For example, \(X_{1}^{\prime(2)}\) might be: Echocardiogram report states EF \(< 35\%\) (Yes/No), subject to errors from measurement variability or reporting delays. Similarly, \(X_{2}^{\prime(2)}\) could be: Lab result shows NT-proBNP \(> 2000\) pg/mL (Yes/No), with potential errors from assay variability or transient conditions. Physician notes on dyspnea, specific medication prescriptions, or the presence of certain diagnostic codes can also be binarized to form other \(S^{\prime(2)}\) variables, each carrying its own Predictor Error.

\subsection{The Motivating Case Study: Cleveland Clinic Abu Dhabi}

The mapping above aligns with the assumed data architecture leveraged in the CCAD case study introduced in Section~\ref{Intro}. That endeavor was explicitly motivated by the failure of traditional risk calculators to apply to the local population in Abu Dhabi \citep{alshamsi2020performance, hajat2012weqaya, oulhaj2020agreement, radaideh2017cardiovascular}, a common generalizability problem. In the language of our framework, these traditional risk calculators represented failed Model Transfer -- the importation of a static product optimized for a Western population that proved brittle in a new environment. To address this, the CCAD case study shifted to Methodology Transfer. They did not import a better model; they built a ``Local Factory'' capable of refining their own raw, live stream of operational data (their local data swamp) into a bespoke ML instrument with the intention of reflecting ``the contemporary health dynamics relevant to the local populations.'' While the study's result was initially positioned as an empirical curiosity, G2G now allows us to begin to identify the mechanisms responsible for its performance.

In retrospect, the vast pool of \(32,000+\) error-prone predictors -- encompassing demographics, diagnoses, procedures, labs, medications, flowsheets, vitals, and hospital encounter descriptors -- functioned as a high-entropy \(S^{\prime(2)}\) layer that had sufficient Breadth to lower the Prevalence Floor to the point where local sample sizes were sufficient for robust inference. 

The contrarian success of the model was not merely due to the volume of data, but due to the architecture of the selected subsets. The researchers employed an automated feature selection process that did not penalize collinearity when reducing the pool to a smaller high-D set (\(m \approx 900\) to \(4,000\)). Specifically, their algorithm aimed to discard saturated noise while implicitly preserving Informative Redundancy. The resulting, refined predictor portfolio remained broad enough to triangulate the latent \(S^{(1)}\) states (underlying latent health states, risk and protective factors, and clinical care pathways) amid real-world noise, effectively lowering the Structural Uncertainty floor that had limited the performance of standard-of-care Clean Parsimony models, while also remaining within the model's and system's capacities. This performance was enabled because the Local Factory was not bound by the Intersectional Constraint; it was free to utilize the thousands of site-specific data artifacts that a universal model would have been forced to discard.

Through this retrospective lens, the CCAD case study provides a demonstration that it is possible for Dirty Breadth -- when managed through high-D redundancy-preserving selection -- to outperform Clean Parsimony. It suggests that automated, self-adaptive systems trained on uncurated live streams are not just theoretically possible, but may also be empirically viable. Ultimately, the CCAD case study suggests that the solution to the generalizability crisis is not to force the world's data to fit a universal model, but to \textit{democratize the blueprint}. By deploying a Local Factory, the hospital successfully converted its unique data swamp -- including site-specific coding artifacts -- from a liability into a high-performance predictive asset for its local population.

\section{Limitations and Future Research Agenda}
\label{Limitations and a Future Research Agenda}

The framework's current presentation has several limitations that pave the way for important future research. We first discuss these limitations, then suggest future lines of related inquiry or development.

\subsection{Framework Limitations}

The theoretical underpinnings and scope of the current work are subject to certain constraints.

\begin{itemize}
\item
Although extensions to continuous data (Appendix~\ref{continuous vars}) and a model for a specific type of systematic error (Section~\ref{A Model of Systematic Predictor Errors and Its Impact}) are discussed, the analysis primarily details dynamics using binary variables and assumes independent Predictor Errors for clarity. The framework does not include an explicit analytical model for Correlated Error Realizations.
\item
Although extensions to alternative causal structures are introduced in Appendix~\ref{Robustness of the Framework Across Alt Causal Structures}, the current framework is developed based on a specific hierarchical model.
\item
The current framework does not encapsulate systems where processes are modeled longitudinally.
\item
The theoretical analysis primarily focuses on the asymptotic behavior as \(m\) increases, generally assuming sufficient \(N\). While Section~\ref{Interplay of N and m} provides a foundational analysis of the interplay between \(N\), \(m\), and latent complexity -- explaining the statistical feasibility of the Breadth strategy -- we have not yet provided formal sample complexity bounds that define the minimum \(N\) required for reliable \(S^{(1)}\) recovery as a function of latent dimensionality (\(k\)), predictor dimensionality (\(m\)), and the system's signal-to-noise ratios.
\item
A critical practical difficulty involves assessing the framework's applicability. Formally validating the assumed latent structure (e.g., via conditional independence tests or standard SEM fit indices) is challenging when the observed data are error-prone, as these errors can confound such diagnostic tests.
\item
While we discuss the spectral heuristic for applicability, we do not provide formal sample complexity bounds defining the phase transition point at which the latent signal eigenvalues successfully separate from the noise spectrum as a function of \(N\), \(m\), and the signal-to-noise ratio.
\item
A critical practical difficulty lies in disentangling Informative Collinearity from other sources of correlation that do not offer the same benefits.
\item
A practical difficulty lies in the choice or development of modeling methodologies capable of extracting relevant information from latent hierarchies.
\item
The strategy of increasing dimensionality to improve robustness introduces a trade-off with computational and statistical efficiency (i.e., the resource cost of filtering sludge from the crude during training). Determining the point of diminishing returns -- where the cost of filtration outweighs the value of the extracted signal -- is a key practical challenge.
\item
By focusing on abstract latent states (\(S^{(1)}\)) to achieve robustness, the framework may reduce the direct interpretability of individual observed predictors (\(S^{\prime(2)}\)), which can be a significant drawback in applications where model transparency in terms of the specific observed predictors is paramount.
\item
The proposed mechanisms for mitigating systematic error (Section~\ref{Benign Overfitting Link: Towards a Unified Understanding of Robustness}) and violations of LI (Section~\ref{Violating LI}) rely on the hypothesis that highly flexible, unsupervised models have the capacity to successfully absorb these dependencies as pseudo-states without excessively increasing the Effective Signal Rank. The conditions under which this absorption is successful remain largely uncharacterized, though we hypothesize that Causal Consistency plays a critical role in preventing the Effective Signal Rank from becoming prohibitive.
\item
While the framework addresses robustness to Outcome Error via the connection to BO (Section~\ref{Benign Overfitting Link: Towards a Unified Understanding of Robustness}), the core information-theoretic analysis (Section~\ref{Core Theoretical Analysis}) does not explicitly model the impact of Observational Error in the outcome variable (\(Y^{\prime}\)). Additionally, we do not currently provide a formal information-theoretic derivation for the limits of the Metadata-Adversarial strategy in scenarios where metadata is noisy, incomplete, or confounded with the outcome itself.
\item
While the framework discusses the theoretical potential for identifying systematic errors, it relies on the semantic gap: the mathematical indistinguishability of low-rank latent signals and low-rank systematic errors. Consequently, the mitigation of malignant overfitting due to Common Method Variance is practically contingent upon the availability and accuracy of external metadata.
\item
The framework's benefits, particularly the connection to BO and the mitigation of the Curse of Dimensionality, rely on the assumption that the latent structure (\(S^{(1)}\)) is effectively low-rank relative to dimensionality (\(m\)). While the analysis demonstrates how correlation and prevalence induce sparsity (Section~\ref{Benign Overfitting Link: Towards a Unified Understanding of Robustness}), the applicability of the framework to domains with inherently high-D and decentralized latent drivers (i.e., intrinsically high \(K_{eff}\)) remains uncharacterized.
\item 
While the spectral heuristic is robust for binary data (where covariance rank tracks configurations), a methodological dissonance arises in mixed data regimes (e.g., continuous latent states generating continuous and binary observations). In such cases, linear diagnostics may identify artifactual difficulty factors due to non-linear thresholding functions (e.g., sigmoids), inflating the Effective Signal Rank estimate well beyond both \(k\) and \(K_{eff}\) \citep{McDonaldAhlawat1974}.
\item
The framework currently relies on the premise that high-capacity models will ``emergently'' discover the latent structure despite optimizing a topologically flat objective (e.g., Cross-Entropy). We do not currently provide a formal bound quantifying the efficiency loss -- or optimization drag -- incurred by this Topological Dissonance compared to methods that explicitly operationalize the latent structure in their loss functions.
\item
The current simulation study is limited to random Predictor Error and does not empirically test the mitigation of an SERg.
\item
While the simulation study empirically validates the core mechanics of the framework, the presented results utilize a low-dimensional latent space (\(k=4\)) to ensure the simulations can be reproduced on standard local hardware. Consequently, the empirical demonstration does not fully stress-test the framework against the extreme latent complexity (\(K_{eff}\)) expected in real-world data swamps, though the provided codebase allows users to scale \(k\) to test these limits locally.
\end{itemize}

\subsection{Future Research}

Addressing these limitations points to several key directions for future work. By pursuing these avenues, the theoretical robustness and practical applicability of the framework can be significantly enhanced.

\begin{flushleft}
\textbf{Theoretical Advancements:}
\end{flushleft}

\begin{itemize}
\item
Extend the complete mathematical analysis to encompass mixed data types, building upon the groundwork laid in Appendix~\ref{continuous vars}.
\item
Extend the mathematical analysis to encompass different common causal structures and directions, building upon the groundwork laid in Appendix~\ref{Robustness of the Framework Across Alt Causal Structures}.
\item
Extend the mathematical analysis to analyze violations of LI, building upon the discussion in Section~\ref{Violating LI}.
\item
Conduct investigations into alternative systematic error models (beyond that discussed in Section~\ref{A Model of Systematic Predictor Errors and Its Impact}), focusing on their impact on efficiency and asymptotic convergence.
\item
Extend the mathematical analysis to explicitly model error in the outcome variable \(Y^{\prime}\).
\item
Extend the mathematical analysis to explicitly model different sources of collinearity.
\item
To address the challenge of interpretability, focus on creating novel post-hoc explanation methods designed for this latent variable framework.
\item
Extend the mathematical analysis to handle dynamic systems or longitudinal data, where latent states \(S^{(1)}\) and \(S^{(2)}\) may evolve over time, and observations \(S^{\prime(2)}\) are collected sequentially.
\item
Extend the mathematical analysis linking the data-generating structure to BO to formally describe boundaries imposed by unmitigated systematic error.
\item
Develop formal sample complexity bounds to quantify the relationship between \(N\), \(m\), \(k\), and related signal-to-noise and latent complexities (e.g., \(K_{rlzd}\) and \(K_{eff}\)) establishing the conditions under which G2G robustness benefits can be practically realized.
\item
Develop an ``Entropic Scree'' estimator utilizing Total Correlation (TC) to generalize spectral diagnostics to mixed-data systems. This research would aim to mathematically define ``Entropic Eigenvalues'' and derive the associated sample complexity bounds required to rigorously distinguish the signal subspace from the noise floor in mixed-data systems where linear metrics may overestimate signal complexity.
\item
Formalize the mechanics of Topological Alignment. Future work should mathematically derive the efficiency gains (in terms of convergence rate \(R\) and sample complexity) achieved by resolving the objective-capability gap -- specifically comparing the performance of standard flat objective functions against those that explicitly encode the latent structure in the loss function (e.g., Variational Autoencoders or structured prediction).
\end{itemize}

\begin{flushleft}
\textbf{Empirical Validation and Guideline Development:}
\end{flushleft}

\begin{itemize}
\item
Further systematic simulation studies to map the framework's performance boundaries. These should investigate the rate of improvement in \(S^{(1)}\) estimation as dimensionality increases, the influence of varying \(S^{(1)} - S^{(2)}\) correlation strengths, the impact of different types and degrees of systematic errors, and the interplay between \(N\) and \(m\).
\item
Conduct systematic simulation studies to examine the relative ability of different modeling methodologies to successfully extract relevant information from data generated from the primary structure and its highly aligned variants.
\item
Conduct systematic simulation studies to examine the relative effectiveness of different automated feature selection strategies to efficiently retain relevant information from data generated from the primary structure and its highly aligned variants.
\item
Conduct systematic sensitivity analyses to evaluate how robust the framework's benefits are to unmitigated deviations from its core assumptions, such as violations of LI, departures from the independent error model, and misspecification of the number of latent states in \(S^{(1)}\).
\item
Conduct systematic simulation studies to empirically demonstrate the proof showing that the ``favorable covariance matrix'' prerequisites for BO are footprints of the \(S^{(1)} \rightarrow S^{(2)} \rightarrow S^{\prime(2)}\) structure.
\item
Conduct systematic sensitivity analyses to empirically demonstrate that BO is more likely to occur in data aligned with the primary structure.
\item
Conduct systematic studies to validate the Absorption Hypothesis, empirically investigating the conditions (e.g., model capacity, sample size, complexity of violations) under which highly flexible, unsupervised models successfully mitigate systematic errors and LI violations by learning an expanded latent space.
\item
Perform empirical validation on diverse real-world datasets characterized by complex Predictor Error structures to substantiate the framework's practical utility.
\item 
Conduct systematic simulation studies to validate the Metadata-Adversarial strategy, specifically evaluating the efficacy of the coefficient zeroing technique in preventing malignant overfitting under varying degrees of correlation between systematic error (\(Z\)) and the observed outcome (\(Y^{\prime}\)).
\item
Scale the simulation framework using high-performance compute clusters to formally map the boundaries of the efficiency rate (\(R\)) as latent dimensionality (\(k\)) increases, thoroughly stress-testing the framework against the extreme latent complexity (\(K_{eff}\)) expected in real-world data swamps.
\end{itemize}

\begin{flushleft}
\textbf{Methodological Innovation for Applicability Assessment:}
\end{flushleft}

\begin{itemize}
\item
Develop robust methodologies and diagnostic tools for assessing the appropriateness of the proposed framework's structure when working with error-prone data, building upon the ideas discussed in Section~\ref{Framework Model and Notation}.
\item
Develop models explicitly designed to leverage the framework's proposed core mechanisms, as well as those related to mitigating systematic error and violations of LI.
\item
Develop Relative Redundancy-based feature selection algorithms to efficiently operationalize the theory of maximizing redundancy for reliability.
\end{itemize}

\section{Conclusion}
\label{Conclusion}

This paper introduces G2G as a theoretical framework that helps resolve a central paradox in modern ML: the success of highly flexible models on high-D, collinear, error-prone data. By analyzing a hierarchical data-generating structure, the framework recasts high-D Breadth and Informative Collinearity as assets for predictive robustness, as they enable the stable recovery of the latent \(S^{(1)}\) layer that fundamentally govern the outcome (\(Y\)).

Unlike prior work focused on predictive ML that treats noise as a monolithic nuisance, we show that Predictor Error and Structural Uncertainty obey different information-theoretic limits. This allows us to prove that while Breadth can asymptotically eliminate both forms of noise, Depth strategies -- analogous to targeted data cleaning -- remain fundamentally bounded by Structural Uncertainty regardless of measurement precision. Significantly, we prove that accessing these benefits necessitates interactions. This justifies the superiority of highly flexible models in noisy, high-D settings, explaining how they implicitly bridge the gap between the Flat Topological Operationalization (via the standard loss functions) and the hierarchical reality of the data by using their capacity to hack the loss function and reconstruct the latent layer. Yet, this reliance on model capacity incurs a tangible cost; the robustness offered by Breadth is inextricably linked to increased computational burden. Consequently, the asymptotically optimal \textit{use everything} strategy must be balanced against the practical constraints of finite processing power -- the engine's capacity. This tension underscores the critical role of P-DCAI strategies to achieve efficiency -- deliberately enriching the fuel's energy density to achieve the maximum rate of latent structure recovery with the minimum necessary predictor set -- as a governing principle for real-world application.

Our analysis extends to complex scenarios, establishing the theoretical boundaries imposed by pervasive Systematic Error Regimes while demonstrating resilience to localized ones. We also show why highly flexible, unsupervised models can theoretically mitigate violations of core structural assumptions (LI and independent error). We demonstrate that this capability is principled from a data-architectural perspective: the model is not ``fixing'' the data but rather discovering a more complex, underlying latent structure (proven to be distributionally equivalent) that naturally absorbs these dependencies. However, this capability creates a critical vulnerability: if the same systematic error affects both the predictors and the outcome (Common Method Variance), the model risks malignant overfitting by treating measurement artifacts as predictive signals. To mitigate this, we advocate for Metadata-Adversarial strategies -- explicitly modeling and then neutralizing error sources -- to ensure that the absorbed structures represent genuine latent drivers rather than shared biases. Alternatively, an exclusive focus on cleaning the outcome data (a strategy that aligns with traditional DCAI) can also resolve this threat -- severing the link between the systematic error source and the outcome handles label noise while absorption mitigates Predictor Error and Structural Uncertainty without requiring the exhaustive curation of the predictor-space. Thus, G2G provides one principled, data-architectural justification for the emphasis traditional DCAI places on ensuring high label fidelity, particularly in contexts characterized by Common Method Variance.

Crucially, we address the statistical feasibility of the Breadth strategy, resolving the apparent tension with the Curse of Dimensionality. We explain why the latent architecture creates a Prevalence Floor that averts data sparsity, clarifying the distinct roles of \(m\) in solving predictor ambiguity and \(N\) in solving outcome ambiguity. Furthermore, we highlight that while these bounds hold for the worst-case chaotic generative system, the presence of Causal Consistency in real-world data fundamentally shifts the learning task from rote memorization to parametric interpolation, significantly relaxing the sample complexity required for convergence. We acknowledge, however, that applying this framework presents a practical paradox: confirming the presence of the necessary latent architecture is inherently difficult because the Predictor Error the framework mitigates also confounds standard diagnostic tests. Successful application therefore relies on the use of robust heuristics, such as the Spectral and Model-Based heuristics proposed herein, rather than traditional validation metrics.

Perhaps most significantly, we demonstrate that this same architecture provides a generative origin for the low-rank-plus-diagonal covariance characteristics linked to BO. We identify two distinct mechanisms responsible for this phenomenon: first, latent sparsity naturally constrains the Effective Latent Complexity (\(K_{eff}\)) significantly below the theoretical maximum, providing the plausibility of low-rank structure required for BO even in chaotic generative systems; second, Causal Consistency acts as an accelerant, further collapsing the rank toward the number of latent drivers (\(k\)) and widening the spectral gap necessary for robust generalization. Crucially, this analysis redefines the noise component of that covariance structure. We establish that this component is not merely a nuisance artifact of Predictor Error, but a formal limit imposed by the system’s probabilistic architecture. This connection provides a first step towards a unified data-architectural understanding of robustness -- one that encompasses both Outcome Error and predictor-space noise.

Overall, this manuscript argues that robustness is not solely the product of models and algorithms but also arises from structures in the data when certain conditions are met, suggesting a reconceptualization of gold not solely as the absence of data errors, but more comprehensively, as the presence of an exploitable architecture. Ultimately, this framework explains when and why big data can be viewed through the lens of Measurement Theory. We clarify that Structural Uncertainty is a function of the predictor set, implying that Clean Parsimony is fundamentally limited by the information it fails to capture. Just as a psychometrician does not rely on a few perfect questions to measure intelligence, but rather aggregates hundreds of imperfect questions to triangulate a reliable score, ML practitioners should not seek a handful of pristine predictors. Instead, we can treat high-D data as a vast, multi-item measurement instrument. While G2G was motivated by the complexities of clinical EHR data, this logic holds regardless of the specific domain, so long as the data are generated by the latent architecture. In financial forecasting, for example, latent economic states (e.g., investor sentiment) are inferred through thousands of noisy trading signals. In remote sensing and petroleum exploration, latent biophysical or geological conditions are triangulated via sensor arrays subject to atmospheric and processing artifacts. Similarly, in the biological sciences -- ranging from drug discovery to single-cell multi-omics -- latent molecular activities and on-target engagements are observed through highly variable, artifact-prone assays. In all these cases, applying this framework allows practitioners to leverage the high-D Breadth of these imperfect measurements to robustly recover the underlying latent signals, shifting the engineering focus from the curation of individual variables to the architectural design of the predictor portfolio. Ultimately, gold is not the lack of error in the individual variables, but is the reliability of the aggregate signal.

By providing a theoretical explanation for how robustness might be engineered from uncurated, error-prone data under specific structural conditions, this framework offers initial justification for exploring the Local Factory approach to ML deployment. This approach solves the generalizability problem not by exporting weights -- which are brittle due to the Intersectional Constraint -- but by exporting the methodology, allowing institutions to continuously learn from their own local, real-world data streams. In this more democratized, Methodology Transfer paradigm, the messiness of local data is no longer automatically viewed as an insurmountable liability, but instead as a proprietary resource -- a rich crude oil -- that can be uniquely leveraged for local predictive benefit.

We hope this work spurs future theoretical development and empirical testing to expand and develop G2G further, leading to a shift in how the ML community views and approaches predictive robustness in the era of big data.

\section*{Disclosures}
\label{Disclosures}

This research did not receive any grants from funding agencies in the public, commercial, or not-for-profit sectors.

T.J. Lee-St. John and J.L. Lawson have a pending patent application on methods and systems for implementing aspects of the framework presented in this manuscript. These inventions were developed independently of the authors' institutional affiliations.

\appendix

\section{Formulation of \(\phi(X_{i}^{(1)},X_{j}^{(2)})\)}
\label{phi corr}

The general formula for the phi correlation coefficient for two binary variables \(A\) and \(B\) is:

\[\phi(A,B) = \frac{P(A = 1,B = 1) - P(A = 1)P(B = 1)}{\sqrt{P(A = 1)P(A = 0)P(B = 1)P(B = 0)}}\]

\begin{raggedright}
\vspace{0.5\baselineskip}
Let \(A = X_{i}^{(1)}\) and \(B = X_{j}^{(2)}\).

\vspace{0.5\baselineskip}
\textbf{Definitions:}
\end{raggedright}

\begin{enumerate}
\def\labelenumi{\arabic{enumi}.}
\item
Marginal Probability of \(X_{i}^{(1)}:\)
\end{enumerate}

\[p_{i}^{(1)} = P(X_{i}^{(1)} = 1),\]
\[P(X_{i}^{(1)} = 0) = 1 - p_{i}^{(1)}.\]

\begin{enumerate}
\def\labelenumi{\arabic{enumi}.}
\setcounter{enumi}{1}
\item
Conditional Probabilities for \(X_{j}^{(2)}\) given \(X_{i}^{(1)}\): These parameters define the relationship or link strength between \(X_{i}^{(1)}\) and \(X_{j}^{(2)}\). They are based on the set of conditional probabilities \(\varphi_{S^{(1)} \rightarrow X_{j}^{(2)}}\) described in Section~\ref{Framework Model and Notation}:
\end{enumerate}

\[\gamma_{1 \rightarrow j} = P(X_{j}^{(2)} = 1|X_{i}^{(1)} = 1),\]
\[\gamma_{0 \rightarrow j} = P(X_{j}^{(2)} = 1|X_{i}^{(1)} = 0).\]

\vspace{0.5\baselineskip}
If \(X_{j}^{(2)}\) is determined solely by \(X_{i}^{(1)}\), then \(\gamma_{1 \rightarrow j}\) and \(\gamma_{0 \rightarrow j}\) are specific values from the set \(\varphi_{S^{(1)} \rightarrow X_{j}^{(2)}}\). More generally, they represent these conditional probabilities after marginalizing out any other states in \(S^{(1)}\) that might also influence \(X_{j}^{(2)}\).

\begin{enumerate}
\def\labelenumi{\arabic{enumi}.}
\setcounter{enumi}{2}
\item
Marginal Probability of \(X_{j}^{(2)}\) (denoted \(p_{j}^{(2)}\)): This is derived from the above parameters.
\end{enumerate}

\[p_{j}^{(2)} = P(X_{j}^{(2)} = 1) = P(X_{j}^{(2)} = 1 | X_{i}^{(1)} = 1)P(X_{i}^{(1)} = 1) + P(X_{j}^{(2)} = 1 | X_{i}^{(1)} = 0)P(X_{i}^{(1)} = 0)\]
\[p_{j}^{(2)} = \gamma_{1 \rightarrow j}p_{i}^{(1)} + \gamma_{0 \rightarrow j}(1 - p_{i}^{(1)})\]

\vspace{0.5\baselineskip}
Then \(P(X_{j}^{(2)} = 0) = 1 - p_{j}^{(2)}\).

\begin{enumerate}
\def\labelenumi{\arabic{enumi}.}
\setcounter{enumi}{3}
\item
Joint Probability
\(P(X_{i}^{(1)} = 1, X_{j}^{(2)} = 1)\):
\end{enumerate}

\[P(X_{i}^{(1)} = 1,\ X_{j}^{(2)} = 1) = P(X_{j}^{(2)} = 1|\ X_{i}^{(1)} = 1)P(X_{i}^{(1)} = 1) = \gamma_{1 \rightarrow j}p_{i}^{(1)}\]

\begin{raggedright}
\vspace{0.5\baselineskip}
\textbf{Substitution} into the general formula yields:

\[\phi(X_{i}^{(1)},X_{j}^{(2)}) = \frac{p_{i}^{(1)}(1 - p_{i}^{(1)})(\gamma_{1 \rightarrow j} - \gamma_{0 \rightarrow j})}{\sqrt{p_{i}^{(1)}(1 - p_{i}^{(1)})p_{j}^{(2)}(1 - p_{j}^{(2)})}}.\]

\vspace{0.5\baselineskip}
Which simplifies to:

\[\phi(X_{i}^{(1)},X_{j}^{(2)}) = (\gamma_{1 \rightarrow j} - \gamma_{0 \rightarrow j})\sqrt{\frac{p_{i}^{(1)}(1 - p_{i}^{(1)})}{p_{j}^{(2)}(1 - p_{j}^{(2)})}}.\]

\vspace{0.5\baselineskip}
The formula reveals that the observed correlation is not solely a function of the raw signal strength but is modulated by the base rates of the latent and predictor states. Understanding this interplay is critical for interpreting predictor quality.
\end{raggedright}

\subsection{Impact of Latent State Prevalence: The ``Maximum Ambiguity'' Principle}

The term \(\sqrt{p_{i}^{(1)}(1 - p_{i}^{(1)})}\) in the numerator is proportional to the standard deviation of the binary latent state \(X_{i}^{(1)}\). This term is maximized when the base rate \(p_{i}^{(1)} = 0.5\) and approaches zero as the base rate nears \(0\) or \(1\).

Interpretation: For a fixed signal strength, the observable correlation will be strongest when the latent state is most ambiguous (a \(50/50\) chance of being present). A latent state that is extremely rare or extremely common has very little variance, making it difficult for its effect to manifest as a strong correlation, even if its causal influence is large. For example, if a light switch is almost always ``on'' (base rate near \(1\)), its position has a very weak correlation with the light's status, even though it is the direct cause.

\subsection{Impact of Predictor Prevalence: The ``Rare but Reliable Signal'' Effect}

The term \(\sqrt{p_{j}^{(2)}(1 - p_{j}^{(2)})}\) is in the denominator. This means the phi correlation is maximized when the predictor's own variance is low (i.e., when \(p_{j}^{(2)}\) is close to \(0\) or \(1\)).

Interpretation: This reveals a powerful effect: a predictor that is very rare (or very common) can be extremely informative. Because it rarely changes its state, any change it does make is a very reliable signal. A fire alarm is a perfect example. Its base rate of being ``on'' is extremely low (\(p_{j}^{(2)} \approx 0\)). This makes it an incredibly powerful predictor; the rare event of it turning on has an exceptionally high correlation with the presence of a fire.

\subsection{Synthesis: The Interplay of Signal Strength and Base Rates}

The overall phi correlation combines these effects. A predictor's ultimate value is not determined by its signal strength alone. A predictor with a moderate signal strength but optimal base rates (a latent state near \(50\%\) prevalence and a rare true predictor state) could have a higher phi correlation than a predictor with a very strong signal but suboptimal base rates.

This has direct implications for the feature selection strategies discussed in Section~\ref{Towards Proactive Data-Centric AI}, suggesting that practitioners should not just hunt for the strongest causal links but also consider the underlying prevalence of the phenomena they are measuring.

\section{Generalization to Continuous Variables}
\label{continuous vars}

Broadly interpreted, the core principles of our framework (Statements 1-3) -- that Predictor Error and Structural Uncertainty can be mitigated by high-D Breadth and Informative Collinearity, and that this benefits prediction of \(Y\) -- readily generalize to scenarios involving continuous variables, as the logic describing these mechanisms depends on the universal properties of information flow. These statements, built upon mutual information, conditional entropy, and the DPI have direct analogues in the continuous domain and the relational properties are universal across variable types \citep{cover2006}. Mutual Information remains non-negative and measures the dependency between variables, regardless of whether they are discrete or continuous. The principle that conditioning (adding information) cannot increase uncertainty holds for both types. The DPI, which states that information cannot be gained by processing data through a chain, is universally applicable. Therefore, Statements 1-3 (broadly interpreted) hold whether the variables are discrete or continuous. This appendix provides related detail.

We first note, however, that other conclusions discussed do not automatically translate to continuous variables (or more generally, to mixed variable types) simply due to the universal properties of information flow, but instead require explicit extension of the primary structure (and subsequent analysis). Specific arguments requiring such extension include the precise asymptotic behavior (efficiency, reliability, completeness, and the effect of Systematic Error Regimes) and the inherent non-linearity of the optimal prediction function. Such theoretical exposition generalizing the full range of results to mixed variable types is a clear direction for future investigation.

\subsection{Core Generalizations}

The discrete entropy \(H(X)\) for a discrete random variable \(X\) is given by \(H(X) = -\sum_{x}{p(x)\log(p(x))}.\) For continuous variables, the analogous concept is differential entropy, denoted \(h(X)\) for a continuous random variable \(X\) with probability density function (\(pdf\)) \(f(x)\), defined as: 

\[h(X) = -\int_{- \infty}^{\infty}{f(x)\log(f(x))dx}.\]

\vspace{0.5\baselineskip}
Importantly, differential entropy does not share all the same properties as discrete entropy (e.g., it can be negative). However, key relational concepts like conditional entropy and mutual information have direct and meaningful analogues in the continuous domain.

The conditional differential entropy \(h(X|Y)\) for continuous variables \(X\) and \(Y\) with joint \(pdf\), \(f(x,y)\), and conditional \(pdf\), \(f(x|y)\), is defined as:

\[h(X|Y) = -\int_{-\infty}^{\infty}{\int_{-\infty}^{\infty}{f(x,y)\log(f(x|y))dxdy}}.\]

\vspace{0.5\baselineskip}
Similar to the discrete case, \(h(X|Y) \leq h(X)\), with equality if and only if \(X\) and \(Y\) are independent. This property is crucial for our arguments regarding uncertainty reduction.

Mutual information \(I(X;Y)\) for continuous variables retains its definition in terms of entropies and remains non-negative (\(I(X;Y) \geq 0\)) and symmetric:

\[I(X;Y) = h(X) - h(X|Y) = h(Y) - h(Y|X) = \int_{-\infty}^{\infty}{\int_{-\infty}^{\infty}{f(x,y)\log(\frac{f(x,y)}{f(x)f(y)})dxdy}}.\]

\begin{raggedright}
\vspace{0.5\baselineskip}
Revisiting Statements 1-3 with Differential Entropy:
\end{raggedright}

\begin{itemize}
\item
Statement 1 Analogue: If the Markov chain \(Y \leftarrow S^{(1)} \rightarrow S^{(2)}\) holds for continuous variables, then \(Y\bot S^{(2)}|S^{(1)}\), implying \(I(Y;S^{(2)} | S^{(1)}) = 0\). Consequently, \(h(Y | S^{(1)}, S^{(2)}) = h(Y|S^{(1)})\), meaning true second-stage continuous states are redundant for predicting \(Y\) given true first-stage continuous states.
\item
Statement 2a Analogue: \(h(X_{i}^{(1)}|S^{\prime(2)})\) would be non-increasing as more continuous observed variables \(X_{j}^{\prime(2)}\) are added to the set \(S^{\prime(2)}\). The logic \(I(X_{i}^{(1)};X_{(m + 1)}^{\prime(2)}|S^{\prime(2)}) \geq 0\) applies.
\item
Statement 2b Analogue: \(h(X_{i}^{(1)}|X_{j}^{\prime(2)})\) would be non-increasing as the statistical dependence (e.g., Pearson correlation if linearity is assumed, or more general measures of dependence) between \(X_{i}^{(1)}\) and \(X_{j}^{(2)}\) strengthens. If the dependence between \(X_{i}^{(1)}\) and \(X_{j}^{(2)}\) increases, \(I(X_{i}^{(1)};X_{j}^{(2)})\) generally increases. Given the Markov chain \(S^{(1)} \rightarrow X_{j}^{(2)} \rightarrow X_{j}^{\prime(2)}\) and a fixed error mechanism \(X_{j}^{(2)} \rightarrow X_{j}^{\prime(2)}\) that does not completely mask \(X_{j}^{(2)}\), then \(I(X_{i}^{(1)};X_{j}^{\prime(2)})\) will also be non-decreasing. Since \(h(X_{i}^{(1)} | X_{j}^{\prime(2)}) = h(X_{i}^{(1)}) - I(X_{i}^{(1)};X_{j}^{\prime(2)})\), this implies \(h(X_{i}^{(1)} | X_{j}^{\prime(2)})\) is non-increasing.
\item
Statement 2c Analogue: \(h(X_{i}^{(1)} | X_{j}^{\prime(2)})\) is non-increasing as the fidelity of the observational process \(X_{j}^{(2)} \rightarrow X_{j}^{\prime(2)}\) strengthens. In the continuous domain, fidelity corresponds to the strength of the statistical dependence between the true underlying variable \(X_{j}^{(2)}\) and its observed counterpart \(X_{j}^{\prime(2)}\). As this dependence strengthens (e.g., the variance of measurement noise decreases), the mutual information \(I(X_{j}^{(2)};X_{j}^{\prime(2)})\) increases. The variables form a Markov chain \(S^{(1)} \rightarrow X_{j}^{(2)} \rightarrow X_{j}^{\prime(2)}\). Given a fixed relationship between \(S^{(1)}\) and \(X_{j}^{(2)}\) (provided that \(I(X_{i}^{(1)};X_{j}^{(2)}) > 0\)), an increase in \(I(X_{j}^{(2)};X_{j}^{\prime(2)})\) generally leads to an increase in the end-to-end mutual information \(I(X_{i}^{(1)};X_{j}^{\prime(2)})\). Since \(h(X_{i}^{(1)} | X_{j}^{\prime(2)}) = h(X_{i}^{(1)}) - I(X_{i}^{(1)};X_{j}^{\prime(2)})\), this implies that \(h(X_{i}^{(1)} | X_{j}^{\prime(2)})\) is non-increasing as observational fidelity strengthens.
\item
Statement 2d Analogue: Reduced joint differential entropy \(h(S^{\prime(2)})\) (indicative of increased collinearity or statistical dependence within the continuous predictor-space \(S^{\prime(2)}\)) can reflect the strengthening of dependence between \(X_{j}^{(2)}\) states and a common \(X_{i}^{(1)}\) state across multiple indices \(j\). As \(X_{j}^{(2)}\) states become stronger proxies for a common \(X_{i}^{(1)}\), they become more inter-correlated (collinear). If this correlational structure is not entirely masked by Predictor Error, the observed continuous variables \(X_{j}^{\prime(2)}\) will also exhibit increased collinearity. For continuous variables, especially in cases like multivariate Gaussian distributions, increased correlation among components generally leads to a more compact joint distribution, which is associated with a lower determinant of the covariance matrix and thus a lower differential entropy \(h(S^{\prime(2)})\).
\item
Statement 2e Analogue: \(h(Y|S^{\prime(2)})\) would be non-increasing with an increased number of informative continuous \(S^{\prime(2)}\) variables or with increased Informative Collinearity (arising from shared continuous \(S^{(1)}\) influences) among them. This relies on the improved estimation (reduced conditional differential entropy) of continuous \(S^{(1)}\) states from \(S^{\prime(2)}\).
\item
Statement 3 Analogue: \(h(Y|S^{(1)}) \leq h(Y|S^{\prime(2)})\) 

The primary structure implies the Markov property (\(Y \bot S^{\prime(2)} | S^{(1)}\)) and the DPI (\(h(Y | S^{(1)}) \leq h(Y|S^{\prime(2)})\)). These results follow directly from the factorization of the joint distribution implied by the structure and the general properties of mutual information. They are independent of the variable types.
\end{itemize}

The core idea of leveraging redundancy through multiple, imperfect, correlated continuous observations \(S^{\prime(2)}\) to achieve a more precise estimate of underlying continuous latent states \(S^{(1)}\) (reducing \(h(S^{(1)}|S^{\prime(2)})\)) remains central. Assuming these \(S^{(1)}\) states are informatively linked to a continuous outcome \(Y\), then a more precise estimation of \(S^{(1)}\) will lead to a more robust prediction of \(Y\) (reducing \({h(Y|S}^{\prime(2)})\)).

\section{Asymptotic Analysis: Comparing Infinite Breadth and Infinite Depth}
\label{asymp Breadth v Depth}

The following assumptions, consistent with the framework's definitions, are necessary for the asymptotic analysis:

\begin{enumerate}
\def\labelenumi{\arabic{enumi}.}
\item
\textbf{Identifiability of \(S^{(1)}\):} Different configurations of \(S^{(1)}\) produce distinct distributions over the infinite sequence of \(S^{(2)}\) states, such that the average KL divergence between these distributions is bounded away from zero.
\item
\textbf{Unsaturated Error:} The error mechanisms in both regimes (\(L = 0\) and \(L = 1\)) do not completely mask the \(S^{(2)}\) signal, ensuring positive average information gain per variable.
\item
\textbf{Distinct Conditional Error Regimes:} The error mechanisms corresponding to \(L = 0\) and \(L = 1\) are statistically distinct for the affected variables (set \(A\)) when conditioned on \(S^{(1)}\), such that the average KL divergence between the induced distributions over \(A\) is bounded away from zero.
\end{enumerate}

Let \(S_{\infty}^{\prime(2)}\) denote the infinite sequence of observed variables. Then:

\[H_{L}(\infty) = H(S^{(1)} | S_{\infty}^{\prime(2)},L) + H(L | S_{\infty}^{\prime(2)}) - H(L|S^{(1)},S_{\infty}^{\prime(2)}).\]

\vspace{0.5\baselineskip}
Consider a fixed error regime \(L = l\). We examine the distinguishability of two distinct configurations of first-stage latent states, \(s_{1}\) and \(s_{2}\). Due to LI, the likelihood of the observations factorizes:

\[P(S^{\prime(2)}|S^{(1)} = s,L = l) = \prod_{j = 1}^{m}{P(X_{j}^{\prime(2)}|S^{(1)} = s,L = l)}.\]

\vspace{0.5\baselineskip}
The distinguishability of the joint distributions is measured by the Kullback-Leibler (KL) divergence \citep{kullback1951information}. Due to the factorization, the joint KL divergence is the sum of the individual KL divergences:

\[D_{KL}(P(S^{\prime(2)} | s_{1},l) \parallel P(S^{\prime(2)} | s_{2},l)) = \sum_{j = 1}^{m}{D_{KL}(P(X_{j}^{\prime(2)} | s_{1},l) \parallel P(X_{j}^{\prime(2)} | s_{2},l))}.\]

\vspace{0.5\baselineskip}
Under the assumptions of Identifiability of \(S^{(1)}\) and Unsaturated Error, the distributions induced on the observations by \(s_{1}\) and \(s_{2}\) are distinct. This implies that the average KL divergence across the infinite sequence of variables is bounded away from zero. Therefore, as \(m \rightarrow \infty\), the joint KL divergence goes to infinity, implying that the distributions generated by \(s_{1}\) and \(s_{2}\) are asymptotically orthogonal (perfectly distinguishable). By the properties of KL divergence and the Law of Large Numbers, the posterior probability \(P(S^{(1)} | L = l,S_{\infty}^{\prime(2)})\) concentrates entirely on the true underlying \(S^{(1)}\) state. Thus, the conditional entropy is zero (\(H_{0}(\infty) = 0\) and \(H_{1}(\infty) = 0\)). Consequently, \(H(S^{(1)} | S_{\infty}^{\prime(2)},L) = 0\). Substituting this into \(H_{L}(m)\) yields:

\[H_{L}(\infty) = I(S^{(1)};L|S_{\infty}^{\prime(2)}).\]

{\raggedright
\vspace{0.5\baselineskip}
This result now depends on how the number of affected predictors, \(m_{A}\), scales with \(m\).}

\subsection{Localized Systematic Error}

In the scenario where systematic error affects only a fixed, finite number of predictors, as \(m \rightarrow \infty\), the number of unaffected variables \(m_{A^{C}} \rightarrow \infty\).

The variables in \(S_{A^{C}}^{\prime(2)}\) follow the baseline regime regardless of the state of \(L\). Thus,

\[P(S_{A^{C}}^{\prime(2)} | S^{(1)},L) = P(S_{A^{C}}^{\prime(2)}|S^{(1)}).\]

\vspace{0.5\baselineskip}
By the logic of Infinite Breadth applied to this subset, an infinite sequence of such variables allows for the perfect recovery of \(S^{(1)}\):

\[H(S^{(1)} | S_{\infty A^{C}}^{\prime(2)}) = 0.\]

\vspace{0.5\baselineskip}
Since \(S^{(2)}\) includes \(S_{A^{C}}^{\prime(2)}\), conditioning cannot increase entropy:

\[H_{L}(\infty) = H(S^{(1)} | S_{\infty}^{\prime(2)}) \leq H(S^{(1)} | S_{{\infty A}^{C}}^{\prime(2)}) = 0.\]

\vspace{0.5\baselineskip}
Thus, if the systematic error is localized to a finite subset of predictors, the Infinite Breadth strategy completely mitigates its impact asymptotically. The infinite number of predictors with independent errors overwhelm the localized correlation.

\subsection{Pervasive Systematic Error}

In this scenario, the number of affected predictors grows indefinitely (e.g., a fixed fraction of variables are affected, or the universal case where \(m_{A} = m\)).

Compare the distributions generated under the two regimes for a fixed configuration \(S^{(1)} = s\): \(P(S_{\infty}^{\prime(2)}|s,L = 0)\) and \(P(S_{\infty}^{\prime(2)}|s,L = 1)\). Under the assumption of Distinct Error Regimes, the mechanisms defining \(L = 0\) and \(L = 1\) are different. This implies that the KL divergence between the distributions they induce on observed variables (given \(s\)) is bounded away from zero.

\[D_{KL}(P(S_{\infty}^{\prime(2)} | s,L = 0) \parallel P(S_{\infty}^{\prime(2)} | s,L = 1)) =\] 
\[\sum_{j = 1}^{\infty}{D_{KL}(P(X_{j}^{\prime(2)} | s,L = 0) \parallel P(X_{j}^{\prime(2)} | s,L = 1))} = \infty\]

\vspace{0.5\baselineskip}
Since the joint distributions are asymptotically orthogonal, the active error regime can be identified with perfect certainty from the infinite sequence of observed variables when \(S^{(1)}\) is known. Consequently, \(H(L | S^{(1)},S_{\infty}^{\prime(2)}) = 0\). Substituting this into \(H_{L}(\infty)\), we obtain:

\[H_{L}(\infty) = H(L | S_{\infty}^{\prime(2)}).\]

{\raggedright
\vspace{0.5\baselineskip}
This result is crucial to our framework. In the limit of Infinite Breadth when a pervasive number of variables are affected by the SERg, the residual uncertainty about \(S^{(1)}\) is exactly equal to the uncertainty remaining about the error regime \(L\) after observing the data \(S_{\infty}^{\prime(2)}\). Both Predictor Error and Structural Uncertainty are overcome in this limit.}

Thus, asymptotic performance depends on whether the error regimes are statistically distinguishable from the observations alone.

\[H(L | S_{\infty}^{\prime(2)}) = H(L) - I(L;S_{\infty}^{\prime(2)})\]

{\raggedright
\vspace{0.5\baselineskip}
, where \(H(L)\) is the binary entropy of the prior, \(H_{b}(\pi) = - \pi\log(\pi) - (1 - \pi)\log(1 - \pi)\).}

Distinguishable Regimes: If the statistical patterns in \(S_{\infty}^{\prime(2)}\) generated by \(L = 0\) are distinct from those generated by \(L = 1\) such that the regime can be partially identified from the data alone (\(I(L;S_{\infty}^{\prime(2)}) > 0\)), then \(H(L | S_{\infty}^{\prime(2)}) < H_{b}(\pi)\). If identification is perfect, \(H(L | S_{\infty}^{\prime(2)}) = 0\), and \(H_{L}(\infty) = 0\). Thus, perfect recovery of \(S^{(1)}\) is possible even with systematic errors, provided the regimes are perfectly identifiable from the data.

Indistinguishable Regimes (A Conservative Bound): If the observations \(S_{\infty}^{\prime(2)}\) provide no information about the error regime (\(I(L;S_{\infty}^{\prime(2)}) = 0\)), the uncertainty about \(L\) remains at its maximum, \(H_{b}(\pi)\).

Interpretation: The asymptotic conditional entropy \(H_{L}(\infty)\) fundamentally depends on the scaling behavior of the number of affected predictors, \(m_{A}\):

\begin{enumerate}
\def\labelenumi{\arabic{enumi}.}
\item
If \(m_{A}\) is finite (Localized Correlation), \(H_{L}(\infty) = 0\). The systematic error is asymptotically negligible.
\item
If \(m_{A}\) is infinite (Pervasive Correlation), \(H_{L}(\infty) = H(L | S_{\infty}^{\prime(2)})\), and the residual entropy is bounded by the inability to identify the error regime from the observed data.
\end{enumerate}

\subsection{Analysis of the Infinite Depth Strategy: The Fundamental Limitation of Structural Uncertainty}

\begin{itemize}
\item
Let \(N\) denote the number of repeated measurements (the Depth).
\item
The \(n^{\text{th}}\) measurement of \(X_{j}^{(2)}\) is denoted \(X_{j,n}^{\prime(2)}\).
\item
The total observed dataset is \(S_{N}^{\prime(2)}\).
\item
Let \(S_{N}^{\prime(2)}\) denote the sequence of \(N\) variables and \(S_{N \rightarrow \infty}^{\prime(2)}\) denote the infinite sequence as \(N \rightarrow \infty\).
\item
We assume the baseline error model where observations are conditionally independent across predictors and across repetitions, given \(S^{(2)}\). We also assume unsaturated error.
\end{itemize}

Proposition: In the limit of Infinite Depth, the asymptotic conditional entropy of \(S^{(1)}\) given \(S_{N}^{\prime(2)}\) is equal to the Structural Uncertainty inherent in the \(S^{(1)} \rightarrow S^{(2)}\) step:

\[H(S^{(1)} | S_{N \rightarrow \infty}^{\prime(2)}) = H(S^{(1)} | S^{(2)}).\]

\vspace{0.5\baselineskip}
Use the information-theoretic identity, setting \(A = S^{(1)}\), \(B = S_{N}^{\prime(2)}\), and \(C = S^{(2)}\) to obtain:

\[H(S^{(1)} | S_{N}^{\prime(2)}) = H(S^{(1)} | S^{(2)},S_{N}^{\prime(2)}) + I(S^{(1)};S^{(2)}|S_{N}^{\prime(2)}).\]

\vspace{0.5\baselineskip}
Due to the Markov property, \(S^{(1)}\) is conditionally independent of \(S_{N}^{\prime(2)}\) given \(S^{(2)}\). Therefore, \(H(S^{(1)} | S^{(2)},S_{N}^{\prime(2)}) = H(S^{(1)}|S^{(2)})\), and \(H(S^{(1)} | S_{N}^{\prime(2)}) = H(S^{(1)}|S^{(2)}) + I(S^{(1)};S^{(2)}|S_{N}^{\prime(2)})\). The first term is the Structural Uncertainty. The second term is the residual mutual information.

Expand the residual term using the definition of conditional mutual information:

\[I(S^{(1)};S^{(2)} | S_{N}^{\prime(2)}) = H(S^{(2)} | S_{N}^{\prime(2)}) - H(S^{(2)} | S^{(1)},S_{N}^{\prime(2)}).\]

\vspace{0.5\baselineskip}
We now analyze the behavior of \(H(S^{(2)} | S_{N}^{\prime(2)})\) as \(N \rightarrow \infty\). Consider a single state \(X_{j}^{(2)}\). Its repeated observations are conditionally independent given \(X_{j}^{(2)}\). Under the unsaturated error assumption, the distributions \(P(X_{j,n}^{\prime(2)} | X_{j}^{(2)} = 1)\) and \(P(X_{j,n}^{\prime(2)} | X_{j}^{(2)} = 0)\) are distinct. The KL divergence between the joint distributions induced by the two states of \(X_{j}^{(2)}\) grows linearly with \(N\). As \(N \rightarrow \infty\), the KL divergence approaches infinity, implying the distributions become asymptotically orthogonal (perfectly distinguishable). Consequently, the uncertainty about \(X_{j}^{(2)}\) vanishes. Since this holds for all \(j's\): \(H(S^{(2)} | S_{N \rightarrow \infty}^{\prime(2)}) = 0\). Thus,

\[I(S^{(1)};S^{(2)} | S_{N \rightarrow \infty}^{\prime(2)}) = H(S^{(2)} | S_{N \rightarrow \infty}^{\prime(2)}) - H(S^{(2)} | S^{(1)},S_{N \rightarrow \infty}^{\prime(2)}) = 0 - H(S^{(2)} | S^{(1)},S_{N \rightarrow \infty}^{\prime(2)}).\]

\vspace{0.5\baselineskip}
Because conditioning cannot increase entropy, \(H(S^{(2)} | S^{(1)},S_{N \rightarrow \infty}^{\prime(2)})\) must also converge to zero. Substituting these results into \(H(S^{(1)} | S_{N}^{\prime(2)})\) we obtain:

\[H(S^{(1)} | S_{N \rightarrow \infty}^{\prime(2)}) = H(S^{(1)} | S^{(2)}).\]

{\raggedright
\vspace{0.5\baselineskip}
This result proves that while Infinite Depth can eliminate uncertainty arising from Predictor Error, it cannot reduce uncertainty about \(S^{(1)}\) beyond what is attainable from a fixed \(S^{(2)}\) set. This represents a fundamental limitation compared to the Infinite Breadth strategy.}

\subsection{Proof of Degeneracy Elimination and the Prevalence Floor}
\label{proof:degeneracy}

We formally demonstrate that increasing Breadth (\(m \rightarrow \infty\)) eliminates configuration-level degeneracy, ensuring that the probability of observing evidence correctly identifying a latent state \(s_a\) converges to the state's prior prevalence.

\vspace{0.5\baselineskip}
\noindent \textbf{Definitions:}
\begin{itemize}
    \item \textbf{Degeneracy:} The condition where a specific observed configuration \(g\) has non-zero likelihood under multiple latent configurations (i.e., \(P(g|s_a) > 0\) and \(P(g|s_b) > 0\)), making the inverse inference ambiguous.
    \item \textbf{Equivalence Class (\(\mathcal{G}_a\)):} The set of all observable predictor configurations \(g\) for which \(s_a\) is the Maximum A Posteriori (MAP) estimate: 
    \[ \mathcal{G}_a = \{ g : \operatorname*{argmax}_{s_k} P(s_k | g) = s_a \}. \]
\end{itemize}

\vspace{0.5\baselineskip}
\noindent \textbf{Theorem:} As \(m \rightarrow \infty\), degeneracy is eliminated such that the probability mass of any latent configuration \(s_k\) (\(k \neq a\)) falling into \(\mathcal{G}_a\) vanishes, and the total probability of \(\mathcal{G}_a\) converges to \(P(s_a)\).

\vspace{0.5\baselineskip}
\noindent \textbf{Proof:}
The probability of observing a configuration belonging to the class \(\mathcal{G}_a\) is:
\[ P(\mathcal{G}_a) = \sum_{g \in \mathcal{G}_a} P(g). \]

Expanding via the Law of Total Probability over all \(K_{rlzd}\) latent configurations:
\[ P(\mathcal{G}_a) = \sum_{k=1}^{K_{rlzd}} P(s_k) \left[ \sum_{g \in \mathcal{G}_a} P(g|s_k) \right]. \]

The term in the brackets, \(\sum_{g \in \mathcal{G}_a} P(g|s_k)\), represents the probability that a vector generated by \(s_k\) falls into the classification region \(\mathcal{G}_a\).

Under the assumption of \textit{Asymptotic Identifiability} (Appendix~\ref{asymp Breadth v Depth}, Assumption 1), the Kullback-Leibler (KL) divergence between the distributions induced by distinct latent states grows linearly with \(m\). This implies the distributions become asymptotically orthogonal. Therefore, the overlap between the generating distribution of \(s_k\) and the classification region of \(s_a\) behaves as follows:

\begin{enumerate}
    \item \textbf{Resolution of Degeneracy (\(k \neq a\)):} The probability that a distinct state \(s_k\) generates a vector falling into \(\mathcal{G}_a\) vanishes.
    \[ \lim_{m\to\infty} \sum_{g \in \mathcal{G}_a} P(g|s_k) = 0 \]
    
    \item \textbf{Completeness of Identification (\(k = a\)):} The probability mass of \(s_a\) concentrates entirely within its own equivalence class.
    \[ \lim_{m\to\infty} \sum_{g \in \mathcal{G}_a} P(g|s_a) = 1 \]
\end{enumerate}

Substituting these limits back into the total probability expansion:
\[ \lim_{m\to\infty} P(\mathcal{G}_a) = P(s_a) \cdot (1) + \sum_{k \neq a} P(s_k) \cdot (0) \]
\[ \lim_{m\to\infty} P(\mathcal{G}_a) = P(s_a). \]

\noindent \textbf{Conclusion:}
In the limit of infinite Breadth, degeneracy is fully resolved. The probability of observing an ambiguous indicator vanishes, and the total probability of observing the unique signature for \(s_a\) is limited only by the prevalence of the latent state itself.

\section{Analysis of Latent Sparsity and Effective Latent Complexity}
\label{latent sparsity}

This appendix demonstrates how correlation among the latent states \(S^{(1)}\) reduces the effective rank, \(K_{eff}\), of the signal covariance matrix, \(C_{signal}\). This analysis addresses the concern that the maximum rank bound of \(2^{k} - 1\) might be too high to satisfy the low-rank link to BO.

Two properties of the latent space contribute to sparsity. First, the joint entropy \(H(S^{(1)})\) is fundamentally constrained by the individual entropies of its component states. The entropy of a single binary state \(H(X_{i}^{(1)})\) is maximized when its prevalence is \(0.5\) and approaches zero as its prevalence nears \(0\) or \(1\). Therefore, a system built from very rare or very common latent states will have a low joint entropy and a small \(K_{eff}\) even before considering dependencies.

Second, correlations among the states further reduce the joint entropy beyond the constraints imposed by prevalence. Here we utilize the Information-Theoretic identity to show that correlations among \(S^{(1)}\) states naturally induce sparsity in the latent space. The argument parallels the logic used in Statement 2d.

\subsection{Assumptions}
We maintain the core assumptions of the framework and add one assumption for analytical clarity, mirroring the approach used in Statement 2:

\begin{raggedright}
\vspace{0.5\baselineskip}
\textbf{Fixed Marginal Entropies:} To isolate the effect of the correlation structure on the joint distribution, we assume the marginal probabilities \(P(X_{i}^{(1)})\), and thus the individual entropies \(H(X_{i}^{(1)})\), are fixed.    
\end{raggedright}

\subsection{The Argument}

\begin{enumerate}
\item
\textbf{Defining the Effective Latent Complexity}

The Algebraic Rank of \(C_{signal} = Cov(E[S^{\prime(2)} | S^{(1)}])\) is bounded by the effective number of configurations of \(S^{(1)}\), \(K_{eff}\). We aim to show that \(K_{eff} \ll 2^{k}\) when the latent \(S^{(1)}\) states are correlated.

\item
\textbf{The Total Correlation Identity (Parallel to Statement 2d)}

We utilize the Information-Theoretic identity that decomposes joint entropy using Total Correlation (TC) \citep{watanabe1960information}. TC measures the total dependency or redundancy within a system.

In Statement 2d, this identity was applied to the observed predictors \(S^{\prime(2)}\) to demonstrate how Informative Collinearity structures the observed space. The same identity governs the structure of the latent space \(S^{(1)}\):

\[H(S^{(1)}) = \sum_{i=1}^{k} H(X_{i}^{(1)}) - TC(S^{(1)}).\]
\item
\textbf{Quantifying Effective Configurations via Entropy}

The joint entropy \(H(S^{(1)})\) measures the uncertainty or ``spread'' of the joint distribution across the \(K_{rlzd}\) realized configurations.

We quantify \(K_{eff}\) using the concept of perplexity: \(K_{eff} = 2^{H(S^{(1)})}\). A distribution concentrated on a few configurations has low entropy and a low \(K_{eff}\).

\item
\textbf{The Mechanism: Correlation Reduces Joint Entropy}

We analyze how correlation among the latent states affects the joint entropy. In the decomposition, since the sum of marginal entropies is fixed (Fixed Marginal Entropies assumption), \(H(S^{(1)})\) is strictly determined by the \(TC(S^{(1)})\).

\(TC(S^{(1)})\) captures all dependencies (both pairwise and higher-order) among the latent states. For binary variables, the mutual information between any pair increases monotonically with the magnitude of their phi correlation. An increase in the magnitude of dependencies (e.g., stronger average pairwise phi correlations) increases the overall dependency captured by \(TC(S^{(1)})\).

\item
\textbf{Conclusion: Correlation Lowers Effective Latent Complexity}

As the dependency among latent states increases:

\begin{itemize}
\item
total correlation (\(TC(S^{(1)})\)) increases,
\item
joint entropy (\(H(S^{(1)})\)) strictly decreases,
\item
the effective number of configurations (\(K_{eff}\)) decreases exponentially.
\end{itemize}

This proves that correlations among the latent states work towards concentrating the probability mass onto a sparse subset of the configuration space. Just as Statement 2d shows how shared causes structure the observed space, this analysis shows how correlation structures the latent space itself -- that even moderate correlation can drastically reduce the Effective Latent Complexity from the \(2^{k}\) theoretical maximum. Consequently, the Effective Signal Rank of covariance matrix remains low, resolving the rank complexity issue required for BO.
\end{enumerate}

\section{Alignment with Alternative Causal Structures}
\label{Robustness of the Framework Across Alt Causal Structures}

Here we examine how our core analysis aligns with different common causal models governing the relationships among states. This list of causal models is not meant to be exhaustive, however -- other causal structures exist, and their alignment/misalignment with G2G is an open question for future theoretical exposition.

\subsection{Uni-Directional Structures: (\(Y \leftarrow S^{(1)} \leftarrow S^{(2)})\) or (\(Y \rightarrow S^{(1)} \rightarrow S^{(2)})\)}

\begin{itemize}
\item
The primary structure explicitly models \(S^{(1)}\) as an antecedent to both \(Y\) and \(S^{(2)}\). However, the core benefit highlighted by our analysis -- the ability of \(S^{\prime(2)}\) observations to facilitate reduction in uncertainty about the \(S^{(1)}\) states -- is of broader relevance than this specific causal ordering. The crucial insight is that imperfect knowledge of \(S^{(1)}\) can obscure its true relationship with \(Y\). If uncertainty about \(S^{(1)}\) states is reduced, then the characterization of the \(Y - S^{(1)}\) link becomes more reliable regardless of the causal direction between \(Y\) and \(S^{(1)}\) or between \(S^{(1)}\) and \(S^{(2)}\). Thus, uni-directional structures are highly aligned with the proposed mechanism.
\end{itemize}

\subsection{Direct Causation \((A \rightarrow B)\)}

\begin{itemize}
\item
The primary structure describing how \(S^{(2)}\) relates to \(S^{(1)}\) is itself a direct causal link (\(S^{(1)} \rightarrow S^{(2)}\)). This is the foundational scenario upon which Statements 2a and 2b are built.
\end{itemize}

\subsection{Common Cause \((A \leftarrow C \rightarrow B)\)}

\begin{itemize}
\item
\(Y\) and \(S^{(2)}\) states commonly caused by \(S^{(1)}\) (i.e., \(Y \leftarrow S^{(1)} \rightarrow S^{(2)}\)): This structure defines the primary structure. Alignment is perfect.
\item
Multiple \(S^{(2)}\) states commonly caused by a single \(S^{(1)}\) state (i.e., \(X_{i}^{(1)} \rightarrow S^{(2)}\)): This structure also aligns perfectly with the primary structure.
\end{itemize}

\subsection{Indirect Causation/Mediation \((A\rightarrow M \rightarrow B)\)}

\begin{itemize}
\item
If \(S^{(1)}\) states (\(A\)) indirectly cause \(S^{(2)}\) states (\(B\)) via an unobserved set of intermediate states \(M\) (i.e., \(S^{(1)} \rightarrow S^{(M)} \rightarrow S^{(2)}\)), then \(S^{(2)}\) still carries information about \(S^{(1)}\). The causal chain ensures that variations in \(S^{(1)}\) propagate to \(S^{(2)}\). Thus, \(S^{(2)}\) can contribute to an estimation of \(S^{(1)}\), and the principles of our proposal still apply, though the strength of the \(S^{(1)} - S^{(2)}\) relationship might be weaker because it is now more distal, potentially requiring more or higher-quality \(S^{\prime(2)}\) variables. The fundamental idea of leveraging downstream effects to clarify upstream latent states remains highly aligned.
\end{itemize}

\subsection{Common Effect/Collider Structures \((A \rightarrow C \leftarrow B)\)}

Given the discussion of directionality in Section 6.1, collider structures also align well with our framework in all scenarios except where \(S^{(1)}\) are colliders caused by both \(S^{(2)}\) and \(Y\).

\begin{itemize}
\item
\(Y\) as the Collider Caused by multiple \(S^{(1)}\) states (i.e., \(S^{(1)} \rightarrow Y\)): Defines the primary structure.
\item
\(S^{(1)}\) as Colliders Caused by multiple \(S^{(2)}\) states (i.e., \(S^{(2)} \rightarrow S^{(1)}\)): Highly aligned (Section 6.1) with the primary structure.
\item
\(S^{(2)}\) as Colliders Caused by multiple \(S^{(1)}\) states (i.e., \(S^{(1)} \rightarrow S^{(2)}\)): Defines the primary structure.
\item
\(S^{(1)}\) as Colliders Caused by \(S^{(2)}\) and \(Y\) (i.e., \(S^{(2)} \rightarrow S^{(1)} \leftarrow Y\)): In this specific topology, the latent state \(S^{(1)}\) is a common effect of \(S^{(2)}\) and \(Y\). Attempting to use the framework's logic here would involve estimating \(S^{(1)}\) from \(S^{(2)}\) in order to predict \(Y\). This is a classic statistical error. Conditioning on a collider (or its descendants) is known to induce spurious, non-causal associations between its causes \citep{pearl2009causality}. In this case, it could create a misleading statistical relationship between \(S^{(2)}\) and \(Y\) that does not exist in reality.

The correct identification of this specific topological vulnerability underscores that the framework is not a universal panacea, but a tool that must be applied with careful reasoning.

We note, however, that when the aim is purely predictive (not causal), utilizing the induced association might be acceptable if the goal is maximizing accuracy and the association is stable.
\end{itemize}

\subsection{Conclusion: Robustness to Different Causal Models}

The framework is robust and beneficial when predictors exhibit direct causation, common cause, or indirect causation. Its principles also remain applicable in certain collider structure scenarios, specifically when the causes of a collider are either always upstream or downstream in the causal chain. However, if a collider variable is midstream (i.e., \(S^{(2)} \rightarrow S^{(1)} \leftarrow Y\)), the framework's logic is misapplied, and this can lead to a confounded estimation of \(S^{(1)}\), potentially degrading the prediction of \(Y\). Table~\ref{Table: Proposal's Alignment with Alternative Causal Structures.} provides a summary of our framework's robustness to the causal structures discussed.

\begin{longtable}[]{@{}
>{\centering\arraybackslash}p{\dimexpr(\linewidth - 2\tabcolsep)*7847/10000\relax}
>{\centering\arraybackslash}p{\dimexpr(\linewidth - 2\tabcolsep)*2153/10000\relax}@{}}
\caption{Proposal's Alignment with Alternative Causal Structures.} \label{Table: Proposal's Alignment with Alternative Causal Structures.}\\
\toprule\noalign{}
\begin{minipage}[b]{\linewidth}\centering
\textbf{Alternative Causal Structures}
\end{minipage} & \begin{minipage}[b]{\linewidth}\centering
\textbf{Alignment with Proposal}
\end{minipage} \\
\midrule\noalign{}
\endhead
\bottomrule\noalign{}
\endlastfoot
Reversed Direction of the Entire Framework
(\(Y \leftarrow S^{(1)} \leftarrow S^{(2)}\)) & High
\\
\hline
Direct Causation (\(A \rightarrow B\)) & Perfect
\\
\hline
Multiple \(S^{(2)}\) States Commonly Caused by a Single \(S^{(1)}\) State (\(X_{i}^{(1)} \rightarrow S^{(2)}\)) & Perfect
\\
\hline
\(Y\) and \(S^{(2)}\) States Commonly Caused by \(S^{(1)}\) (\(Y \leftarrow S^{(1)} \rightarrow S^{(2)}\)) & Perfect
\\
\hline
Indirect Causation/Mediation (\(A \rightarrow M \rightarrow B\)) & High
\\
\hline
\(Y\) as the Collider Caused by Multiple \(S^{(1)}\) States (\(S^{(1)} \rightarrow Y\)) & Perfect
\\
\hline
\(S^{(1)}\) as Colliders Caused by Multiple \(S^{(2)}\) States (\(S^{(2)} \rightarrow S^{(1)}\)) & High
\\
\hline
\(S^{(2)}\) as Colliders Caused by Multiple \(S^{(1)}\) states (\(S^{(1)} \rightarrow S^{(2)}\)) & Perfect
\\
\hline
\(S^{(1)}\) as Colliders Caused by \(S^{(2)}\) and \(Y\) (\(S^{(2)} \rightarrow S^{(1)} \leftarrow Y\)) & Low\\
\end{longtable}

\section{Derivation of the Finite-\(m\) Convergence Rate (Efficiency)}
\label{finite convergence}

This appendix provides a derivation of the convergence rate for the perfect recovery of \(S^{(1)}\) as the number of predictors \(m\) increases, formalizing the concept of ``efficiency'' discussed in Section~\ref{Core Theoretical Analysis}. We analyze the rate the probability of error decays using the Chernoff bound.

\begin{raggedright}
\vspace{0.5\baselineskip}
\textbf{Setup:} We aim to perfectly recover the configurations of \(S^{(1)}\) from the observed data \(S^{\prime(2)}\). Let \(s^{*}\) be the true configuration and \({\widehat{S}}_{m}^{(1)}\) be an optimal estimate based on the \(m\) predictors (e.g., the Maximum A Posteriori estimate). We are interested in the probability of error,
\end{raggedright}

\[P_{e}(m) = P({\widehat{S}}_{m}^{(1)} \neq s^{*}).\]

\subsection{Pairwise Error Probability and Chernoff Information}

Consider two distinct configurations of the \(S^{(1)}\) states, \(s_{p}\) and \(s_{\omega}\). Let \(P_{j^{p}} = P(X_{j}^{\prime(2)}|s_{p})\) and \(P_{j^{\omega}} = P(X_{j}^{\prime(2)}|s_{\omega})\) be the distributions of the \(j^{\text{th}}\) observation under these configurations, respectively. The Chernoff Information (CI) between two distributions \(P\) and \(Q\) is:

\[C(P,Q) = - \min_{\lambda \in [0,1]}\log(\sum_{x}{{P(x)}^{\lambda}{Q(x)}^{1 - \lambda}})\]

{\raggedright
\vspace{0.5\baselineskip}
, where \(P(x)\) and \(Q(x)\) are the probabilities of observing a specific outcome, \(x\), under two different distributions, \(P\) and \(Q\) -- in the context of our framework, this represents the distributions of a predictor under the two different configurations, \(s_{p}\) and \(s_{\omega}\). The summation term measures the amount of overlap between the two distributions, where a value near \(1\) means the distributions are very similar, while a value near \(0\) means they are very distinct. Taking the negative log of this minimized overlap value converts it into an information measure, where a smaller overlap (more distinguishability) results in a larger CI value.}

Due to LI, the observations are independent given \(S^{(1)}\) (though generally not identically distributed). Thus, factorization means the Chernoff bound for the pairwise probability of error (\(P(s_{p} \rightarrow s_{\omega})\)) can be expressed as the exponentiated sum of the individual CIs \citep{cover2006}:

\[P(s_{p} \rightarrow s_{\omega}) \leq \exp(- \sum_{j = 1}^{m}{C(P_{j^{p}},P_{j^{\omega}})}).\]

{\raggedright
\vspace{0.5\baselineskip}
The \(C(P_{j^{p}},P_{j^{\omega}})\) term quantifies the \(j^{\text{th}}\) predictor's contribution to distinguishing this specific pair of configurations, \(s_{p}\) and \(s_{\omega}\).}

\textbf{Informative Redundancy for Reliability:} The summation (which relies on the conditional independence assumption) is the precise mathematical formalization of achieving ``reliability'' from informative redundancy. Imagine predictor A provides a large CI term to the summation. We then incorporate a new predictor, B, which is redundant with A. This means the \(\gamma\) vector (\(\varphi\)) of B is strongly correlated (positively or negatively) with the \(\gamma\) vector of A, resulting in a high magnitude of unconditional correlation between the predictors.

If B were trivially redundant (e.g., a duplicate of A), the realizations of its uncertainty would not be independent of A. This would violate the conditional independence required for this factorization, and B would provide zero conditional information gain.

However, if B is distinct (meaning the realizations of its uncertainty are conditionally independent of A), its redundancy is informative. Because B shares a strongly correlated \(\gamma\) pattern with A, it is similar in signal (though potentially inverted) and thus contributes a large CI term to the summation, just as A does. By repeating this process, each new distinct, informatively redundant predictor adds another large CI term to the sum. This aggregation of large distinguishability scores causes the error bound for distinguishing between \(s_{p}\) and \(s_{\omega}\) to shrink exponentially.

\subsection{The Efficiency Rate as the Average Chernoff Information}

Let \({\overline{C}}_{p,\omega}(m)\) be the average CI across the \(m\) predictors for the pair \((s_{p},s_{\omega})\):

\[{\overline{C}}_{p,\omega}(m) = (\frac{1}{m})\ \sum_{j = 1}^{m}{C(P_{j^{p}},P_{j^{\omega}})}.\]

{\raggedright
\vspace{0.5\baselineskip}
The bound can be rewritten as \(P(s_{p} \rightarrow s_{\omega}) \leq exp(- m \cdot {\overline{C}}_{p,\omega}(m)\)).}

The overall probability of error across all configuration pairs, \(P_{e}(m)\), can be bounded using the union bound over all realized configurations, \(K_{rlzd}\). Assuming non-degenerate priors \(P(s_{p}) > 0\):

\[P_{e}(m) \leq \sum_{p}{P(s_{p})}\sum_{p \neq \omega}{P(s_{p} \rightarrow s_{\omega})} \leq K_{rlzd}^{2} \cdot \max_{p \neq \omega}\exp(- m \cdot {\overline{C}}_{p,\omega}(m)).\]

\vspace{0.5\baselineskip}
Because we are interested in perfect recovery, the convergence rate is limited by the ``least distinguishable'' pair of configurations (i.e., the ``weakest link''). We define the ``Asymptotic Efficiency Rate,'' \(R\), as the average Chernoff information for the least distinguishable pair of configurations:

\[R = \min_{p \neq \omega}{\liminf}_{m \rightarrow \infty}{{\overline{C}}_{p,\omega}(m)}.\]

\vspace{0.5\baselineskip}
This definition highlights the interplay between novelty and informative redundancy as they relate to efficiency. Informative redundancy increases the average CI for specific configuration pairs by accumulating evidence through distinct pathways, thereby enhancing reliability for those configurations. When such redundancy targets the weakest link, this improves efficiency.

\textbf{Novelty for Completeness}: Novelty ensures that the minimum average CI across all pairs of configurations is maximized. \(R\) is the precise mathematical formalization of achieving ``completeness'' from novelty. Imagine a set of predictors all very good at distinguishing between configurations 1 and 2, but all very poor at distinguishing between 1 and 3. The overall efficiency rate will be low because it is limited by this difficult case (configuration 1 vs. configuration 3). Now, we add a new predictor with novel information capable of distinguishing between 1 and 3 -- for this new predictor, its individual CI term will be high. While this does not affect the CI values for the other predictors, its contribution raises the CI for the weakest link, thereby increasing the overall efficiency rate.

\subsection{Conclusion on Convergence}

Under the Asymptotic Identifiability assumption, the average KL divergence is bounded away from zero. Since KL divergence upper bounds the Chernoff information, this assumption also guarantees that \(R > 0\).

Therefore, the probability of error in recovering \(S^{(1)}\) decays exponentially with the number of predictors \(m\) and with \(R\):

\[P_{e}(m) \leq \exp(- m \cdot R).\]

\subsection{Link to Conditional Entropy}

We use Fano's inequality to relate \(P_{e}(m)\) to the conditional entropy \(H(S^{(1)}|S^{\prime(2)})\):

\[H(S^{(1)} | S^{\prime(2)}) \leq H_{b}(P_{e}(m)) + P_{e}(m)\log_{2}(K_{rlzd} - 1)\]

{\raggedright
\vspace{0.5\baselineskip}
, where \(H_{b}()\) is the binary entropy function. Since \(P_{e}(m)\) decays exponentially to zero, \(H(S^{(1)}|S^{\prime(2)})\) also converges rapidly to zero.}

The rate \(R\) governs this convergence and formalizes ``efficiency'': higher quality predictors (those contributing novel or informatively redundant information about the weakest link) increase \(R\) leading to faster convergence.

\section{Proof of Equivalence for LI Violations}
\label{equivalence for LI viol}

This proof demonstrates that a data-generating process with a direct causal link between second-stage states (a violation of LI) is distributionally equivalent to a process with an expanded latent layer where LI holds. This supports the claim that highly flexible, unsupervised models can mitigate such violations by implicitly learning this expanded latent structure.

\begin{raggedright}
\vspace{0.5\baselineskip}
\textbf{Proposition 1:} A system where LI is violated by a direct causal link between second-stage states is distributionally equivalent to an alternative system that incorporates the parent node of the dependency as a ``pseudo-state'' into an expanded latent layer, \(S_{exp}^{(1)}\), with respect to which LI holds.

\begin{enumerate}
\item
\textbf{The Original System (LI Violation):} Define a system that violates the LI assumption. For clarity and without loss of generality, consider a direct causal link from \(X_{1}^{(2)}\) to \(X_{2}^{(2)}\). Assume all other second-stage states (\(X_{j}^{(2)}\) for \(j \geq 3\)) are conditionally independent of each other given the original latent layer, \(S^{(1)}\).

The true joint probability of the second-stage states \(S^{(2)}\) given the first-stage states \(S^{(1)}\) is described by the factorization of the underlying directed acyclic graph:

\[P_{Sys1}(S^{(2)} | S^{(1)}) = P(X_{1}^{(2)} | S^{(1)}) P(X_{2}^{(2)} | S^{(1)},X_{1}^{(2)}) \prod_{j=3}^{m}P(X_{j}^{(2)} | S^{(1)}).\]

\vspace{0.5\baselineskip}
This expression explicitly shows the violation of LI, as the probability of \(X_{2}^{(2)}\) depends not only on \(S^{(1)}\) but also on another second-stage state, \(X_{1}^{(2)}\).

\item
\textbf{The Equivalent System (Restored LI):} Now, we construct an alternative system that is based on an expanded latent layer and adheres to the LI assumption. Let the expanded latent layer be \(S_{exp}^{(1)} = S^{(1)} \cup Z\), where \(Z\) is a new binary latent variable, or ``pseudo-state.''

In this equivalent system, the generative process for \(S^{(2)}\) assumes LI holds with respect to \(S_{exp}^{(1)}\). Therefore, the conditional joint probability is:

\[P(S^{(2)} | S_{exp}^{(1)}) = P(S^{(2)} | S^{(1)},Z) = \prod_{j=1}^{m}P(X_{j}^{(2)} | S^{(1)},Z).\]

\vspace{0.5\baselineskip}
To show equivalence with the original system, we must demonstrate that the marginal probability \(P_{Sys2}(S^{(2)} | S^{(1)})\), obtained by integrating out the pseudo-state \(Z\), is identical to \(P_{Sys1}(S^{(2)} | S^{(1)})\). The marginal probability for System 2 is:

\[P_{Sys2}(S^{(2)} | S^{(1)}) = \sum_{z \in \{0,1\}}P(S^{(2)} | S^{(1)},Z=z)P(Z=z | S^{(1)})\]

\item
\textbf{Constructive Proof of Equivalence:} We now show that by defining the conditional probability distributions for System 2, we can achieve \(P_{Sys2}(S^{(2)} | S^{(1)}) = P_{Sys1}(S^{(2)} | S^{(1)})\). The construction is as follows:

\vspace{0.5\baselineskip}
\textbf{Step a -- Define the generation of the pseudo-state \(Z\):} We define the distribution of the pseudo-state \(Z\) to be identical to the distribution of the ``parent'' node in the dependency chain, \(X_{1}^{(2)}\):

\[P(Z=z | S^{(1)}) \equiv P(X_{1}^{(2)} = z | S^{(1)}).\]

\vspace{0.5\baselineskip}
\textbf{Step b -- Define the generation of the \(S^{(2)}\) states from the expanded latent layer \(S_{exp}^{(1)}\):}

\begin{itemize}
\item
\textbf{For \(X_{1}^{(2)}\) (the parent node):} We make \(X_{1}^{(2)}\) a deterministic copy of the pseudo-state \(Z\):

\[P(X_{1}^{(2)} = x | S^{(1)},Z=z) \equiv \delta(x,z)\]

\vspace{0.5\baselineskip}
, where \(\delta(x,z)\) is the Kronecker delta, which is \(1\) if \(x=z\) and \(0\) otherwise.

\item
\textbf{For \(X_{2}^{(2)}\) (the child node):} We define its generation based on \(S^{(1)}\) and the pseudo-state \(Z\) to match its dependency in the original system.

\[P(X_{2}^{(2)} = x | S^{(1)},Z=z) \equiv P(X_{2}^{(2)} = x | S^{(1)},X_{1}^{(2)}=z)\]

\item
\textbf{For all other states \(X_{j}^{(2)}\) (\(j \geq 3\)):} We define their generation to be independent of the pseudo-state \(Z\), as they were originally independent of the (\(X_{1}^{(2)} \rightarrow X_{2}^{(2)}\)) chain given \(S^{(1)}\):

\[P(X_{j}^{(2)} = x | S^{(1)},Z=z) \equiv P(X_{j}^{(2)} = x | S^{(1)}).\]
\end{itemize}

\textbf{Step c -- Substitute the definitions into the marginal probability for System 2:}

\[P_{Sys2}(S^{(2)} | S^{(1)}) = \sum_{z \in \{0,1\}}P(Z=z | S^{(1)}) [\prod_{j=1}^{m}P(X_{j}^{(2)} | S^{(1)},Z=z)]\]

\vspace{0.5\baselineskip}
Using our definitions:

\[= \sum_{z \in \{0,1\}}P(X_{1}^{(2)}=z | S^{(1)}) \delta(X_{1}^{(2)},z) P(X_{2}^{(2)} | S^{(1)},X_{1}^{(2)}=z) \prod_{j=3}^{m}P(X_{j}^{(2)} | S^{(1)}).\]

\vspace{0.5\baselineskip}
\textbf{Step d -- Collapse the summation:}
The Kronecker delta term, \(\delta(X_{1}^{(2)},z)\), is the key. This term is non-zero (equal to \(1\)) only when the summation index \(z\) takes on the specific value that the random variable \(X_{1}^{(2)}\) has realized. For the other value of \(z\), the entire term inside the summation is zero. Therefore, the summation collapses to a single term where \(z\) is replaced by the value of \(X_{1}^{(2)}\):

\[P(X_{1}^{(2)}=X_{1}^{(2)} | S^{(1)}) (1) P(X_{2}^{(2)} | S^{(1)},X_{1}^{(2)}=X_{1}^{(2)}) \prod_{j=3}^{m}P(X_{j}^{(2)} | S^{(1)}),\]

, which simplifies to:

\[P_{Sys2}(S^{(2)} | S^{(1)}) = P(X_{1}^{(2)} | S^{(1)}) P(X_{2}^{(2)} | S^{(1)},X_{1}^{(2)}) \prod_{j=3}^{m}P(X_{j}^{(2)} | S^{(1)})\]

This final expression is identical to the joint probability \(P_{Sys1}(S^{(2)} | S^{(1)})\).
\end{enumerate}

\textbf{Conclusion and Implications for the Framework:} This proof establishes that any data-generating process with direct causal links among \(S^{(2)}\) states is distributionally indistinguishable from a process with a larger latent space where LI holds. This has profound implications for our framework:

\begin{itemize}
\item
\textbf{Justification for Flexible Models:} The proof of distributional equivalence provides a data-architectural justification for the hypothesis that using highly flexible, unsupervised modeling methodologies (like autoencoders) can mitigate violations of LI. Such models, designed to identify all significant patterns of co-variation, are not constrained by a pre-specified latent structure. Therefore, they have the theoretical capacity to discover the true, expanded latent representation (including pseudo-states to account for residual dependencies), thereby restoring the LI structure required for the framework's robustness mechanisms to function.
\item
\textbf{Generality:}
The logic of this proof can be extended from a single causal link to any directed acyclic graph (DAG) of dependencies within the \(S^{(2)}\) layer. Each parent node in the dependency graph can be treated as a pseudo-state in an expanded latent layer, restoring LI for the entire system.
\item
\textbf{Scope of Equivalence:}
This distributional equivalence applies only to systematic dependencies that can be represented as stable pseudo-states. Random Outcome Error lacks this structural consistency and cannot be absorbed in this manner; fitting to such error remains malignant overfitting (as detailed in Section~\ref{Limitations and a Future Research Agenda}).
\item
\textbf{Connection to BO:}
This result also strengthens the connection to BO. By showing that LI violations can be absorbed into an expanded latent layer, it suggests that the favorable low-rank-plus-diagonal covariance structure may still hold, provided the total complexity of the true latent states plus the pseudo-states remains low relative to the dimensionality of the predictor-space (\(m\)).
\end{itemize}
\end{raggedright}

\section{Additional Simulation Graphs: Breadth vs. Depth as AUROC and AUPRC}
\label{Additional Simulation Graphs: Breadth vs. Depth as AUROC and AUPRC}

\begin{figure}[H]
\centering
\includegraphics[width=.90\textwidth,keepaspectratio]{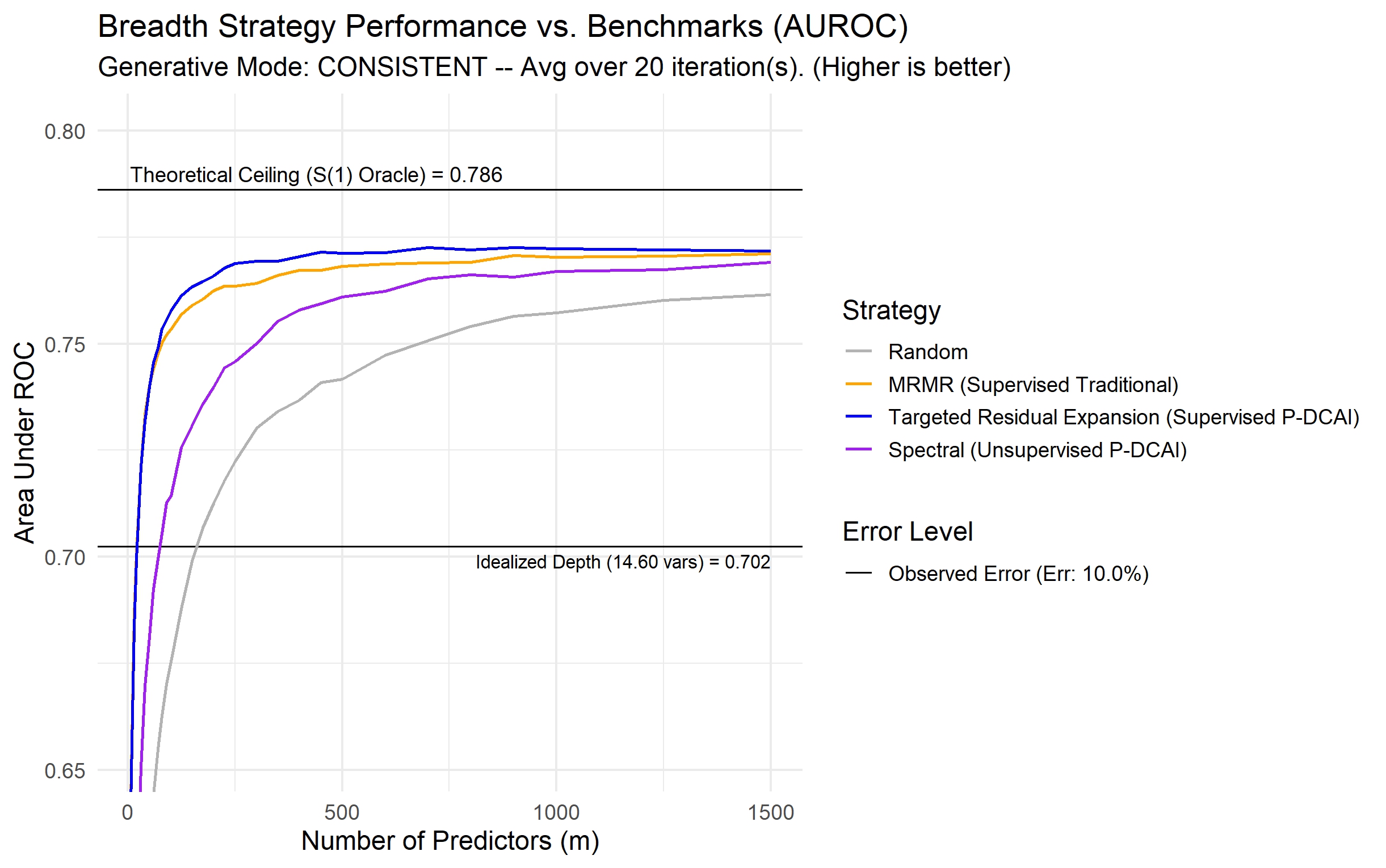}
\caption{Breadth vs. Depth (AUROC): Causally Consistent Variables.}
\label{fig: Breadth vs. Depth (AUROC): Causally Consistent Variables}
\end{figure}

\begin{figure}[H]
\centering
\includegraphics[width=.90\textwidth,keepaspectratio]{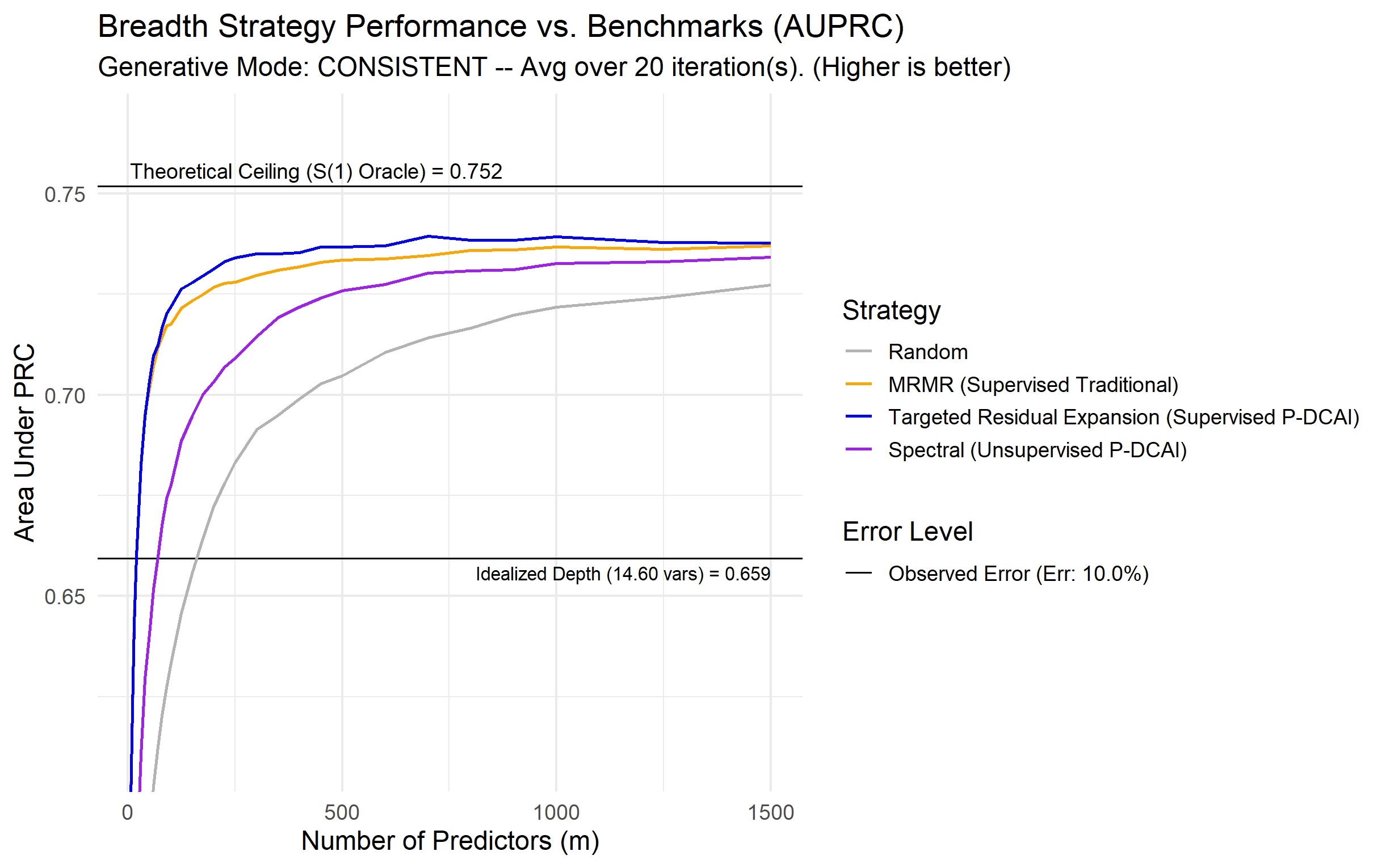}
\caption{Breadth vs. Depth (AUPRC): Causally Consistent Variables.}
\label{fig: Breadth vs. Depth (AUPRC): Causally Consistent Variables}
\end{figure}

\begin{figure}[H]
\centering
\includegraphics[width=.90\textwidth,keepaspectratio]{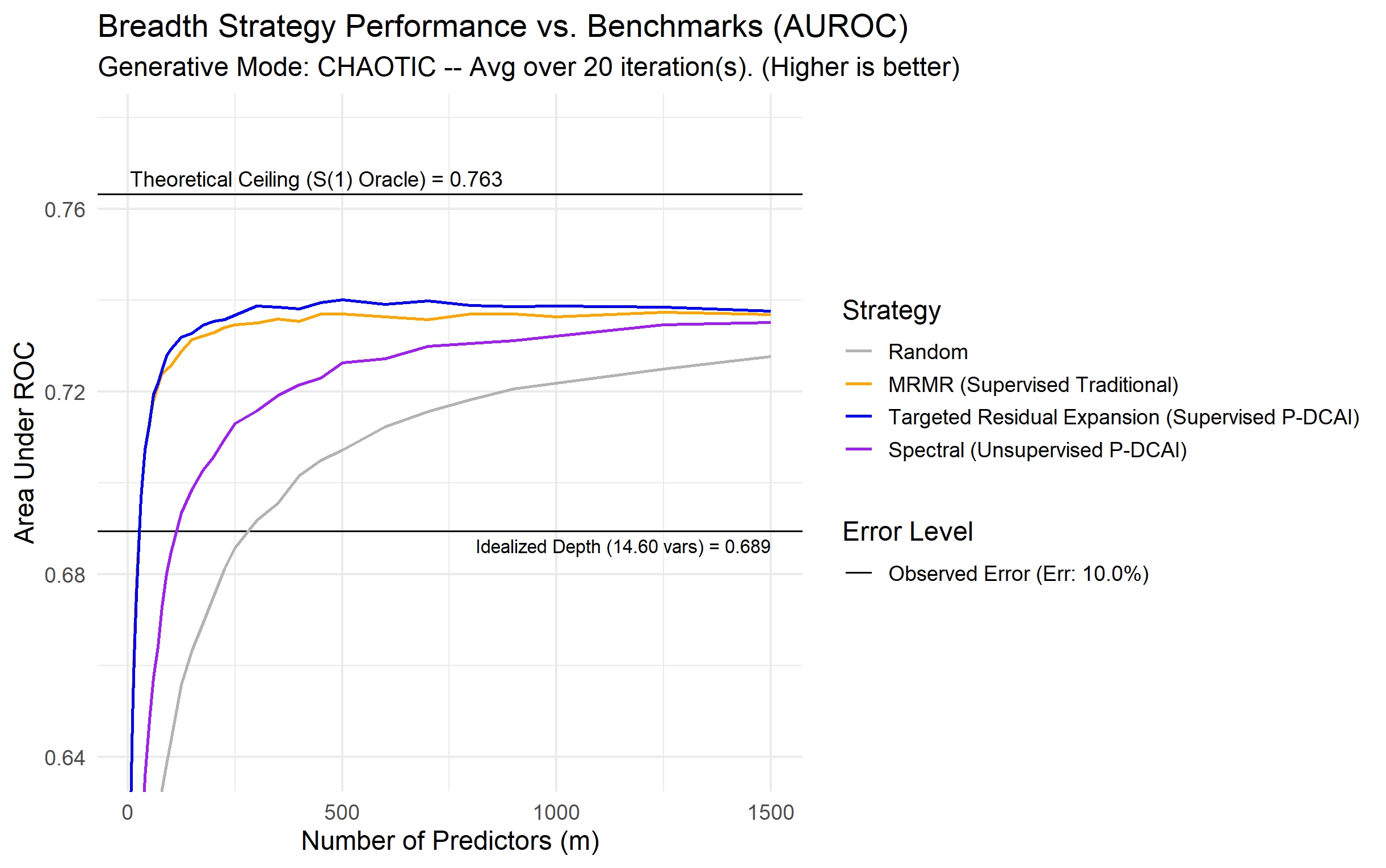}
\caption{Breadth vs. Depth (AUROC): Chaotic Generative Variables.}
\label{fig: Breadth vs. Depth (AUROC): Chaotic Generative Variables}
\end{figure}

\begin{figure}[H]
\centering
\includegraphics[width=.90\textwidth,keepaspectratio]{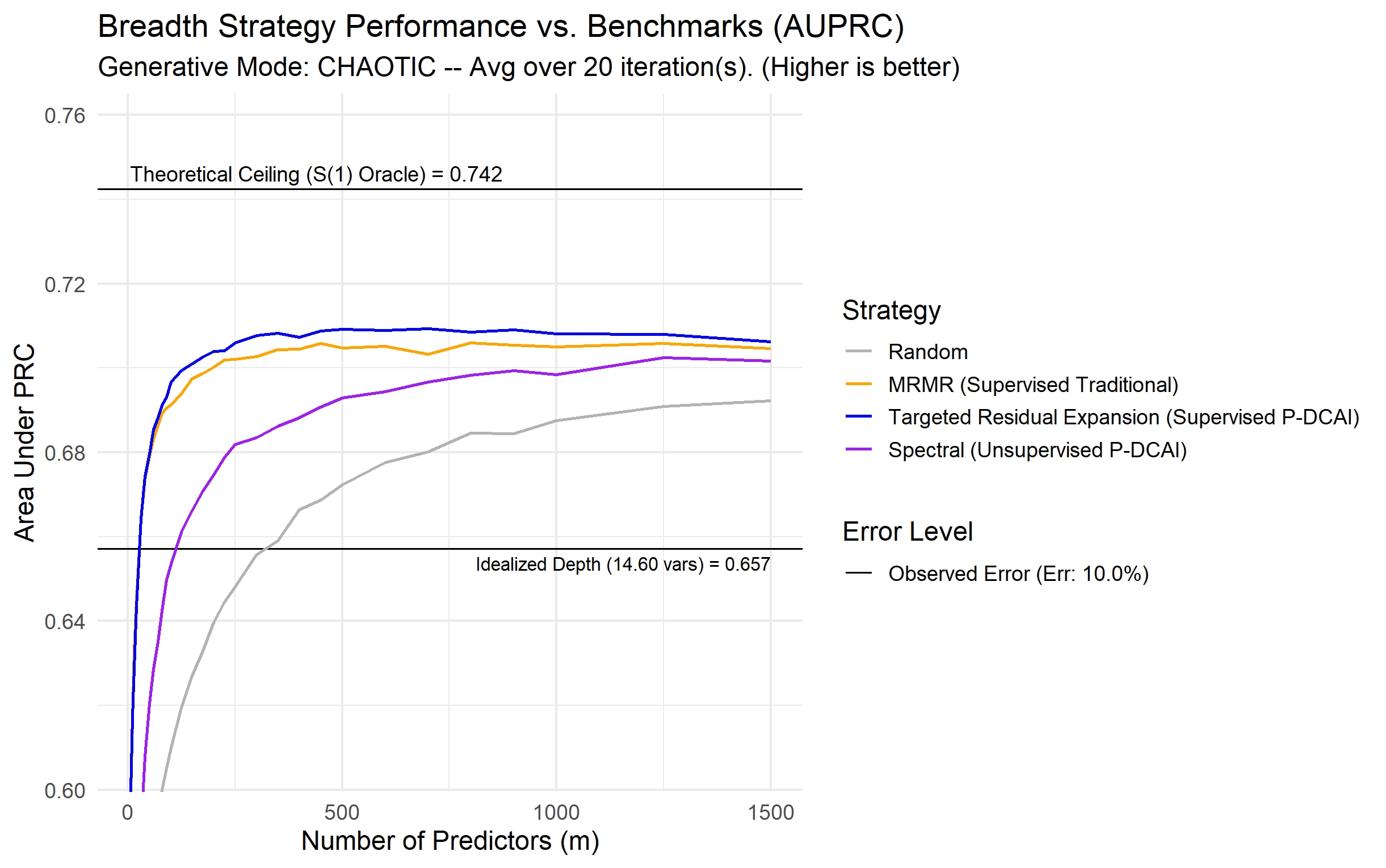}
\caption{Breadth vs. Depth (AUPRC): Chaotic Generative Variables.}
\label{fig: Breadth vs. Depth (AUPRC): Chaotic Generative Variables}
\end{figure}

% Bibliography: jmlr2e loads natbib and sets plainnat
\bibliography{ref}

\end{document}